\documentclass[10pt,journal,comsoc]{IEEEtran}
\usepackage{amsmath,amssymb,amsfonts}
\usepackage{algorithmic}
\usepackage{algorithm}
\usepackage{array}
\usepackage[caption=false,font=footnotesize,labelfont=rm,textfont=rm]{subfig}
\usepackage{textcomp}
\usepackage{stfloats}
\usepackage{url}
\usepackage{verbatim}
\usepackage{graphicx}
\usepackage{cite}
\usepackage{multirow}   
\usepackage{longtable}  
\usepackage{booktabs}  
\usepackage{subfig}  
\usepackage[switch]{lineno}
\usepackage{bm}         
\usepackage{amsthm}%
\usepackage{mathrsfs}%
\usepackage{boondox-cal}
\usepackage{hyperref}
\hypersetup{
	colorlinks=true,
	linkcolor=blue,
	anchorcolor=blue,      
	citecolor=blue,
}

\usepackage {xr}
\makeatletter
\newcommand*{\addFileDependency}[1]{
	\typeout{(#1)}
	\@addtofilelist{#1}
	\IfFileExists{#1}{}{\typeout{No file #1.}}
}
\makeatother

\newcommand*{\myexternaldocument}[1]{
	\externaldocument{#1}
	\addFileDependency{#1.tex}
	\addFileDependency{#1.aux}
}

\myexternaldocument{supplementary}

\begin{document}
	
	\title{How to characterize imprecision \\in multi-view clustering?}
	
	\author{Jinyi Xu, Zuowei Zhang, Ze Lin, Yixiang Chen, Zhe Liu, Weiping Ding	
	%
	%
	\thanks{Jinyi Xu is with Department of Embedded Software and Systems, Software Engineering Institute, East China Normal University, Shanghai, 200062, China, and also with Institute de Recherche en Informatique et Syst\`{e}mes Al\'{e}atoires, Rennes, 35042, France (e-mail: xujinyi96@gmail.com).}
	\thanks{Zuowei Zhang is with Institute de Recherche en Informatique et Syst\`{e}mes Al\'{e}atoires, University of Rennes 1, Lannion, 22300, France (e-mail: zhangzuowei0720@gmail.com).}
	\thanks{Ze Lin is with Department of Embedded Software and Systems, Software Engineering Institute, East China Normal University, Shanghai, 200062, China (e-mail: zee\_lin@foxmail.com).}
	\thanks{Yixiang Chen is with Department of Embedded Software and Systems, Software Engineering Institute, East China Normal University, Shanghai, 200062, China (e-mail: yxchen@sei.ecnu.edu.cn).}
	\thanks{Zhe Liu is now with School of Computer Science and Technology, Hainan University, Haikou 570228, China (e-mail: liuzhe921@gmail.com).}
	\thanks{Weiping Ding is now with Nantong University, Nantong 226019, China (e-mail: dwp9988@1163.com).}}
	\maketitle

	
	\begin{abstract}
		It is still challenging to cluster multi-view data since existing methods can only assign an object to a specific (singleton) cluster when combining different view information. As a result, it fails to characterize imprecision of objects in overlapping regions of different clusters, thus leading to a high risk of errors. 
		In this paper, we thereby want to answer the question: how to characterize imprecision in multi-view clustering? Correspondingly, we propose a multi-view low-rank evidential $c$-means based on entropy constraint (MvLRECM). The proposed MvLRECM can be considered as a multi-view version of evidential $c$-means based on the theory of belief functions. In MvLRECM, each object is allowed to belong to different clusters with various degrees of support (masses of belief) to characterize uncertainty when decision-making. Moreover, if an object is in the overlapping region of several singleton clusters, it can be assigned to a meta-cluster, defined as the union of these singleton clusters, to characterize the local imprecision in the result. In addition, entropy-weighting and low-rank constraints are employed to reduce imprecision and improve accuracy. Compared to state-of-the-art methods, the effectiveness of MvLRECM is demonstrated based on several toy and UCI real datasets.
	\end{abstract}
	

\section{Introduction}
\label{sec:intro}
CLUSTERING has been widely used in various fields such as financial analysis, medical diagnosis, pattern recognition, image processing, and big data~\cite{zhang2020reconstructing,chen2022game,liu2022adcoc}. It is known as an unsupervised classification technique aiming to assign objects to different clusters without any prior information. Many clustering methods based on different ideas have emerged~\cite{chen2022k,luo2020sca2,favati2023two}. However, the information from a single view is not sufficient to obtain a great solution and is also in contrast to human learning.
The research of multi-view clustering is rising since it is closer to the real-world situation. That is, it describes the same object from different aspects and considers the related information from different views.

There are various ideas used in multi-view clustering~\cite{fang2023comprehensive}. One of the most representatives is graph-based. They first generate the similarity graph which can characterize the data structure, and then transfer the clustering problem into a graph partitioning problem, finding the minimum cut using spectral clustering~\cite{xia2022tensorized}. 
However, sometimes spectral clustering cannot directly obtain the specified number of clusters. In this case, they usually carry out \emph{k}-means method for secondary clustering~\cite{shi2000normalized}. 
We divide graph-based clustering into two categories based on how to integrate multi-view graph information. The first category focuses on learning a low dimensional embedding of original data based on the graphs~\cite{cao2015diversity,luo2018consistent}. 
The second category is late fusion based, which generates the unified graph by integrating individual graphs of different views~\cite{huang2021measuring,xia2022tensorized}.
Graph-based clustering also often combines with subspace techniques, such as~\cite{li2022multi,zhang2022kernelized}.
However, it is considered as a hard clustering technique. As a result, it cannot characterize uncertainty when decision-making. That is, each object is assigned to one singleton (specific) cluster certainly and cannot belong to different singleton clusters with various degrees of support at the same time.

\begin{figure}
	\centering
	\includegraphics[scale=0.15]{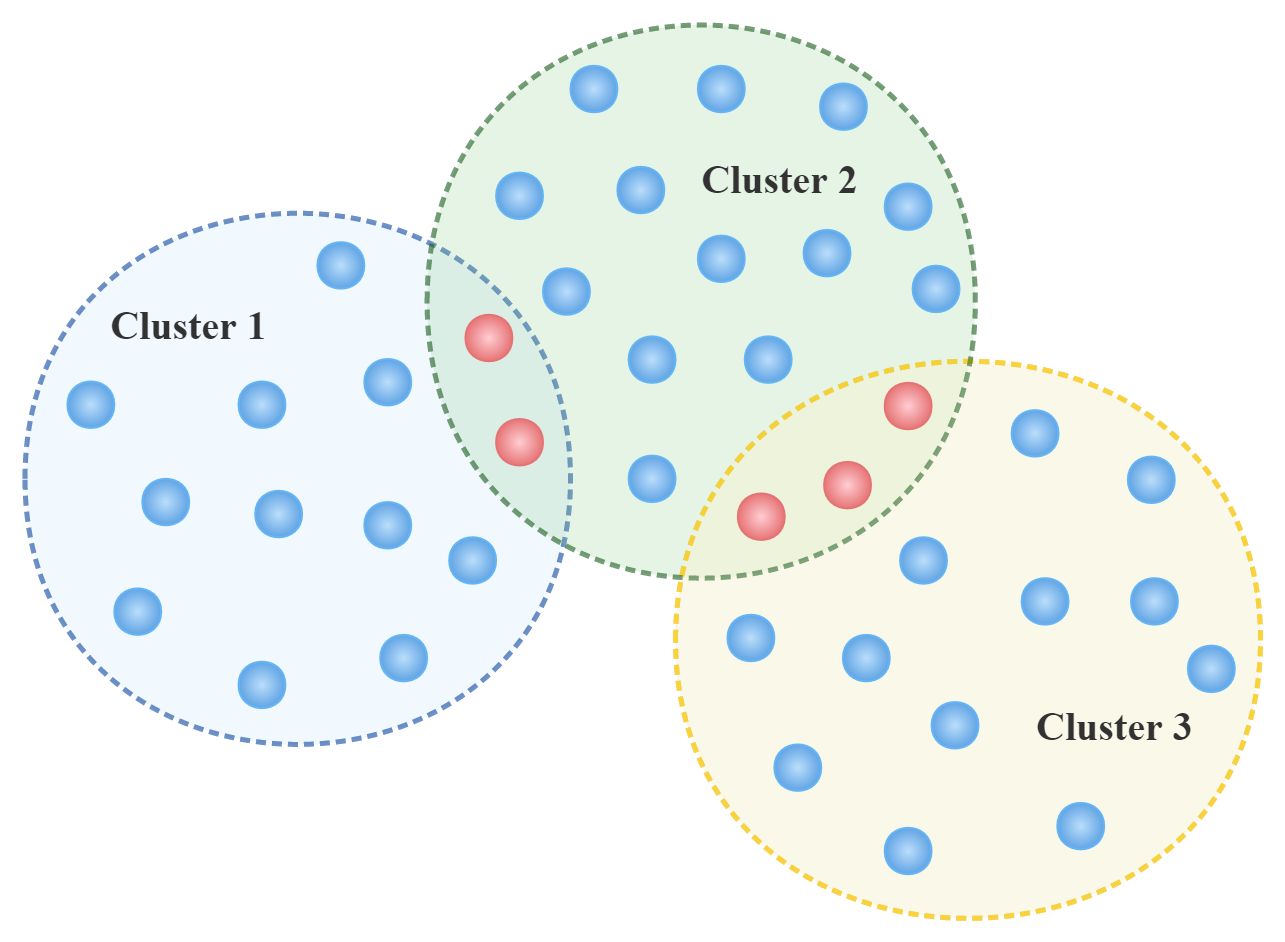}
	\caption{Illustration of imprecision in clustering. The red objects in the overlapping regions are indistinguishable.} 
	\label{fig:Eg}
\end{figure}

Unlike hard clustering, fuzzy clustering describes uncertainty by allowing an object belongs to different singleton clusters with various memberships. These methods first design a suitable objective function and then optimize it to obtain a membership matrix describing the different clusters to which each object belongs. The most popular is the fuzzy \emph{c}-means (FCM) clustering method~\cite{bezdek2013pattern}. Many multi-view methods are proposed based on FCM, such as Co-FKM~\cite{cleuziou2009cofkm}, WV-Co-FCM~\cite{jiang2014collaborative}, Co-FW-MVFCM~\cite{yang2021collaborative}, A$^{\rm SR}$MF~\cite{guo2023pixel}.
Although fuzzy clustering uses memberships to characterize uncertainty in the process, it cannot describe imprecision in the result. That is, although we use multi-view data, the cluster information of some objects is still indistinguishable in a few cases. For example, in Fig.~\ref{fig:Eg}, there are two overlapping regions between different singleton clusters. The objects in both regions are indistinguishable. Forcing them into a singleton cluster will lead to a high risk of errors.

To characterize uncertainty and imprecision, another concept based on the theory of belief functions~\cite{shafer1976mathematical}, credal partition~\cite{denoeux2004evclus}, is introduced. Credal partition extends the concepts of hard and fuzzy partitions\footnote{The clustering method based on credal partition is called evidential clustering. Similarly, hard clustering is based on hard partition and fuzzy clustering is based on fuzzy partition.} by assigning the objects, not only to singleton clusters but also to the union of different singleton clusters, called meta-cluster, with different masses of belief (degrees of support). If the mass of an object belonging to a cluster is not equal to 1, the masses can characterize uncertainty. In this case, the object can belong to different clusters with various masses. When an object is assigned to a meta-cluster with the biggest mass of belief, we can say the cluster information of this object is imprecise, and this object may belong to any singleton clusters in the meta-cluster; Otherwise, the cluster information of this object is precise if it is assigned to a singleton cluster with the biggest mass of belief. 
Based on the credal partition, a single-view clustering method, called evidential \emph{c}-means (ECM)~\cite{masson2008ecm}, is proposed. Although it can characterize uncertainty and imprecision, it cannot handle multi-view data. This is the question we want to address in this paper, or more precisely:

\emph{Can we extend credal partition in multi-view data to characterize uncertainty and imprecision?}

The answer to this question could have significant practical implications. 
Thus, we propose an alternative in this paper, called entropy-weighting multi-view evidential \emph{c}-means clustering based on low-rank constraint (MvLRECM).
The main contributions of MvLRECM can be summarized as follows.
\begin{enumerate}
	\item \emph{Motivation:} We study how to characterize uncertainty and imprecision in multi-view clustering.
	\item \emph{Methods:} We propose a multi-view evidential clustering method to characterize uncertainty and imprecision. First, an objective function is proposed to find the best credal partition based on the theory of belief functions. Second, an entropy-weighting fusion strategy based on low-rank constraint is used to further reduce the risk of errors while improving accuracy. 
	\item \emph{Evaluation:} We extend some definitions of existing clustering evaluation metrics so that they can evaluate hard, fuzzy and credal partitions simultaneously.
\end{enumerate}

The rest of this paper is organized as follows. Sec.~\ref{sec:back} briefly recalls the theory of belief functions. Sec.~\ref{sec:MvLRECM} introduces the proposed method, MvLRECM. Sec.~\ref{sec:exp} analyzes the experiments of MvLRECM compared with other methods. The conclusion is given in Sec.~\ref{sec:con}.

\section{Background}
\subsection{Basics of belief functions}
\label{sec:back}
The theory of belief functions (TBF) is first proposed by Dempster and formed by Shafer, which is also called Dempster–Shafer theory or evidence theory~\cite{shafer1976mathematical}. It is a theoretical framework famous for dealing with uncertain and imprecise information~\cite{smets1998transferable}. The TBF is widely used in various fields, including classification and data clustering~\cite{masson2008ecm,liu2015credal,zhang2021learning,zhou2021evidential,zhang2021dynamic,zhang2021new}.

For a given framework of discernment, $\varOmega=\{a_1,\dots,a_C\}$, it is extended to the power-set $2^\varOmega$ under the theory of belief functions (TBF). $2^\varOmega$ includes all subsets of $\varOmega$. For example, if $\varOmega=\{a_1,a_2,a_3\}$, then $2^\varOmega=\{\emptyset,\{a_1\},\{a_2\},\{a_3\},\{a_1,a_2\},$ $\{a_2,a_3\},\{a_1,a_3\},\varOmega\}$.
The mass of belief is introduced to express the degrees of support for different elements in $2^\varOmega$~\cite{shafer1976mathematical}, defined as a function $m(.)$ from $2^\varOmega$ to $[0, 1]$, verifying
\begin{equation}
\sum_{A \in 2^\Omega}m(A)=1,
\end{equation}
where $A$ is called the focal element of $m(A)$ if $A \subseteq 2^\Omega$ and $m(A)\geq0$, and the set of all its focal elements is called the core of $m(A)$. In clustering problem, these focal elements are regarded as clusters and in clustering results, it can generate three types of clusters: 1) singleton cluster (\emph{e.g.}, $\{a_1\}$); 2) meta-cluster (\emph{e.g.}, $\{a_1,a_2\}$); and 3) noise cluster ($\emptyset$). 
Singleton cluster is independent of others. If the information is sufficient to classify an object exactly, it is assigned to a singleton cluster.
In contrast, meta-cluster is the union of several singleton clusters. If an object is assigned to a meta-cluster, it may belong to any singleton clusters in this meta-cluster.
Noise cluster is a separate cluster. It includes the objects which are far from all singleton clusters. Once an object is assigned to the noise cluster, it is regarded as noise or outliers.

Here, we only introduce the main concepts of this theory. A more detailed description can be found in~\cite{shafer1976mathematical}.

\subsection{Low-rank technique}
\label{sec:Sl}
Low-rank matrix approximation (LRMA) technique, which aims to
recover the underlying low-rank matrix from its degraded observation. LRMA has a wide range of applications in computer
vision and machine learning, such as feature extraction, data compression, image reconstruction, and multi-view clustering. 

There are two categories of methods about how to apply low-rank technique to clustering. The first category studies the multi-view subspace clustering methods by using low-rank kernel mapping to present low-dimensional subspace structure in the high-dimensional feature space~\cite{liu2022adaptive,wang2022hyper}.
The second one introduces low-rank constraints to the multi-view membership or mass matrix to integrate multi-view information~\cite{hu2020squeeze}. However, none of these works tries to use the low-rank constraint in a clustering method in a power-set framework, thus achieving imprecision characterization based on the fusion of information from multiple views.

\section{MvLRECM}
\label{sec:MvLRECM}
In this section, we present the proposed method, MvLRECM, in two parts. Specifically, we construct the objective function of MvLRECM in Subsec.~\ref{sec:alg} and present the optimization method for MvLRECM in Subsec.~\ref{sec:opt}. 

Before going into the details, we introduce some notational conventions: Throughout the paper, matrices are written in boldface capital letters (\emph{e.g.}, $\boldsymbol{X}$); Vectors are written in boldface lowercase letters (\emph{e.g.}, $\boldsymbol{x}$); Constants or sets are written in capital letters (\emph{e.g.}, $X$); Scalars are written in lowercase letters (\emph{e.g.}, $x$). 
For a matrix $\boldsymbol{X} \in \mathbb{R}^{N \times D}$, the $j$th column vector and the $ij$th entry are denoted by
$\boldsymbol{x}_j$ and $x_{ij}$, respectively.
The nuclear norm and the Frobenius norm of $\boldsymbol{X}$ are denoted by$\parallel\boldsymbol{X}\parallel_*$, and $\parallel\boldsymbol{X}\parallel_F^2$, respectively.

For a multi-view dataset with $Q$ views, let $\boldsymbol{X}^1,\dots\boldsymbol{X}^Q$ be the data matrices and $\boldsymbol{X}^q=\{\boldsymbol{x}_1^q,\dots\boldsymbol{x}_N^q\} \in \mathbb{R}^{N \times D_q}$
be the $q$th view of data, where $N$ is the number of objects, and $D_q$ is the dimensionality of the $q$th view. 

We assume that the center of each subset (cluster) $A_j \in 2^\varOmega$ in the $q$th view is represented as $\boldsymbol{\overline{v}}^q_j$, defined by
\begin{equation}
\boldsymbol{\overline{v}}_j^q = \frac{1}{c_j}\sum_{k=1}^C{s_{kj}\boldsymbol{v}_k^q},
\end{equation}
\begin{align}
s.t. ~~~~~~~~~~~~~~~~~~~~~~~~~~~s_{kj} = 
\begin{cases}
1,  & \text{if $a_k \in A_j$}, \\ 
0,  & \text{else},~~~~~~~~~~~~~~~~~~~~~~~~~~~
\end{cases}
\end{align}
where $C$ denotes the number of singleton clusters, $c_j=|A_j|$ denotes the cardinality of $A_j$ (\emph{i.e.} $|\{a_1,a_2\}|=2$), and $\boldsymbol{v}_j^q \in \mathbb{R}^{D_q}$
represents the center of singleton cluster $a_j \in \varOmega$.

For each object $\boldsymbol{x}_i^q$, the masses are represented as 
\begin{equation}
m_{ij}^q=m_i^q(A_j),
\end{equation} 
where $A_j \neq \emptyset, A_j \in 2^\varOmega$. In such a way, $m_{ij}^q$ is low (high) when the distance $d^q_{ij}$ between $\boldsymbol{x}_i^q$ and the center of cluster $A_j$ is high (low). We use the Euclidean distance $|| \cdot ||$ to describe the distance between objects and centers, defined by

\begin{equation}
d^q_{ij} \triangleq \parallel \boldsymbol{x}_i^q - \boldsymbol{\overline{v}}_j^q \parallel.
\end{equation}

\subsection{Model}
\label{sec:alg}
The objective function of the evidential single-view clustering is given by~\cite{masson2008ecm} as follows 
\begin{equation}
\label{eq:ECM_ob}
\min_{\boldsymbol{M},\boldsymbol{V}}\sum_{i=1}^N{\sum_{\{j|A_j \neq \phi, A_j \subseteq \Omega\}}}c_j^\alpha m_{ij}^\beta d_{ij}^2 + \sum_{i=1}^N\delta^2m_{i\phi}^\beta,
\end{equation}
\begin{align}
s.t.~~~ \sum_{\{j|A_j \neq \phi, A_j \subseteq \Omega\}} m_{ij}+m_{i\phi} = 1,  \quad i = 1,2,\dots N,
\end{align}
where $c_j^\alpha$ is a weighting coefficient aiming at penalizing the subsets $A_j \in 2^\varOmega$ of high cardinality. The exponent $\alpha$ controls the degree of penalization.
Parameter $\beta >1$ is a weighting exponent that controls the fuzziness of partition. Parameter $\delta$ controls the number of objects considered as outliers. We take $\beta = 2$, as suggested in~\cite{masson2008ecm}.
%

Unlike single-view data, multi-view data reflects different aspects of each object. The distinguishability of different views is different. In some views, there are highly overlapping regions between different singleton clusters which can only provide less complementary information for the calculations. In this case, the weights of these views should be reduced in the iteration. Here we employ Shannon's entropy-weighting term to 
control the weights of different views. Therefore, the objective function of multi-view evidential $c$-mean clustering is defined initially as follows
\begin{align}
\label{eq:En_MvLRECM}
\min_{\boldsymbol{M},\boldsymbol{V},\boldsymbol{w}}&\sum_{q=1}^Qw_q\left[\sum_{i=1}^N\sum_{\{j|A_j \neq \phi, A_j \subseteq \Omega\}}c_j^\alpha(m_{ij}^q)^2(d_{ij}^q)^2 \right. \nonumber \\
&\left. + \sum_{i=1}^N\delta^2(m_{i\phi}^q)^2\right] + \eta\sum_{q=1}^Qw_q\ln{w_q},
\end{align}
\begin{align}
\label{eq:st}
s.t.~~~
\begin{cases}
\sum_{\{j|A_j \neq \phi, A_j \subseteq \Omega\}} m_{ij}^q+m_{i\phi}^q = 1,~i = 1,2,\dots N,\\
\\
\sum_{q=1}^Qw_q=1,~w_q\in[0,1],~q=1,2,\dots Q, 
\end{cases}
\end{align}
where the weights $\boldsymbol{w}=\{w_1,\dots w_Q\}$ are constrained by the entropy term in the process of clustering and determined automatically. Parameter $\eta $ is a weighting exponent that balances multi-view weights.

Moreover, in the iteration, the complementary information provided by the different views should ideally converge or be highly similar. In other words, the mass matrices of different views should be highly consistent. Of course, the difference (diversity) of mass matrices for different views is also crucial since this diverse information can be complementarily fused to improve accuracy. Whereas most methods complement based on pairs of views, which is inadequate and complex. Low-rank constraints are increasingly being applied to multi-view clustering due to the advantages of globally and efficiently achieving multi-view complementarity while improving the consistency of the mass matrices across different views. Therefore, we also add a low-rank term for the mass matrix of each object from different views as follows
\begin{align}
\label{eq:LR_MvLRECM}
\min_{\boldsymbol{M},\boldsymbol{V},\boldsymbol{w}}&\sum_{q=1}^Qw_q\left[\sum_{i=1}^N\sum_{\{j|A_j \neq \phi, A_j \subseteq \Omega\}}c_j^\alpha(m_{ij}^q)^2(d_{ij}^q)^2 + \right. \nonumber \\&\left. \sum_{i=1}^N\delta^2(m_{i\phi}^q)^2\right]  	 + \theta\sum_{i=1}^N r(\boldsymbol{M}_i) + \eta\sum_{q=1}^Qw_q\ln{w_q}, 
\end{align}
where Parameter $\theta $ is a weighting exponent that controls the degree of mass complementarity in the fusion strategy. 

Since rank minimization is NP-hard and difficult to solve, we relax the low-rank term in Eq.~(\ref{eq:LR_MvLRECM}) by substitutively minimizing the nuclear norm of the estimated matrix, which is a convex relaxation of rank minimizing~\cite{fazel2002matrix}. Finally, we construct the objective function of MvLRECM as follows
\begin{align}
\label{eq:MvLRECM}
\min_{\boldsymbol{M},\boldsymbol{V},\boldsymbol{w}}&\sum_{q=1}^Qw_q\left[\sum_{i=1}^N\sum_{\{j|A_j \neq \phi, A_j \subseteq \Omega\}}c_j^\alpha(m_{ij}^q)^2(d_{ij}^q)^2 + \right.  \nonumber \\
&\left. \sum_{i=1}^N\delta^2(m_{i\phi}^q)^2\right]+ \theta\sum_{i=1}^N\left( \rho\left| \left|  \boldsymbol{Z}_i \right| \right|_* + \parallel\boldsymbol{M}_i - \boldsymbol{Z}_i\parallel_F^2\right)  \nonumber \\
&+\eta\sum_{q=1}^Qw_q\ln{w_q}, 
\end{align}
where $\boldsymbol{Z}_i$ denotes a low-rank approximation of $\boldsymbol{M}_i \in \mathbb{R}^{2\times Q}$. 
We thereby use $J_{MvLRECM}$ to represent the designed objective function, constrained by Eq.~(\ref{eq:st}). 

In the next subsection, we will show how to optimize the designed $J_{MvLRECM}$.

\subsection{Optimization}
\label{sec:opt}
For $J_{MvLRECM}$, we use the Lagrange multiplier method to optimize the variables $\boldsymbol{M},\boldsymbol{V},\boldsymbol{w}$, 
and we use nuclear norm minimization (NNM)~\cite{fazel2002matrix} to obtain the optimized $\boldsymbol{Z}$. 
The specific updated rules are shown as follows

$\bullet$ \emph{Fix $\boldsymbol{M}(t)$ and $\boldsymbol{w}(t)$, update $\boldsymbol{V}(t+1)$,}

$\bullet$ \emph{Fix $\boldsymbol{M}(t)$ and $\boldsymbol{V}(t+1)$, update $\boldsymbol{w}(t+1)$,}

$\bullet$ \emph{Fix $\boldsymbol{V}(t+1)$, $\boldsymbol{w}(t+1)$, and $\boldsymbol{Z}^*(t)$, update $\boldsymbol{M}(t+1)$,}

$\bullet$ \emph{Fix $\boldsymbol{M}(t+1)$, update $\boldsymbol{Z}(t+1)$,}

\noindent where $\boldsymbol{M}(t+1) = \boldsymbol{Z}^*(t+1)$, $\boldsymbol{Z}^*(t+1)$ is the normalization of $\boldsymbol{Z}(t+1)$.

\subsubsection{Update \textbf{V}}
Considering $\boldsymbol{M},\boldsymbol{w}$ are fixed, the minimization of $\boldsymbol{V}$ is an unconstrained optimization problem. For $q$th view, the partial derivatives of $J_{MvLRECM}$ and $(d_{ij}^q)^2$ with respect to the centers are given by
\begin{equation}
\label{eq:pd_d}
\frac{\partial J_{MvLRECM}}{\partial \boldsymbol{v}_l^q} = \sum_{i=1}^N\sum_{\{j|A_j \neq \phi, A_j \subseteq \Omega\}}c_j^\alpha(m_{ij}^q)^2 \frac{\partial(d_{ij}^q)^2}{\partial \boldsymbol{v}_l^q},
\end{equation}
\begin{equation}
\label{eq:pd_v}
\frac{\partial(d_{ij}^q)^2}{\partial \boldsymbol{v}_l^q} = 2(s_{lj})(\boldsymbol{x}_i^q - \overline{\boldsymbol{v}}_j^q)(-\frac{1}{c_j}).
\end{equation}
From Eqs.~(\ref{eq:pd_d})-(\ref{eq:pd_v}), we can obtain
\begin{align}
&\frac{\partial J_{MvLRECM}}{\partial \boldsymbol{v}_l^q}= \nonumber\\
& -2\sum_{i=1}^N\sum_{\{j|A_j \neq \phi, A_j \subseteq \Omega\}}c_j^{\alpha-1}(m_{ij}^q)^2s_{lj}\left(\boldsymbol{x}_i^q -\frac{1}{c_j}\sum_{k=1}^Cs_{kj}\boldsymbol{v}_k^q\right),\nonumber\\
&~~~~~~~~l = 1,2,\dots C. 
\end{align}
Setting these derivatives to zero, we obtain $l$ linear equations in $\boldsymbol{v}_k^q$ which are written as
\begin{align}
& \sum_{i=1}^N\boldsymbol{x}_i^q\sum_{\{j|A_j \neq \phi, A_j \subseteq \Omega\}}c_j^{\alpha-1}(m_{ij}^q)^2s_{lj} \nonumber \\
=& \sum_{k=1}^C\boldsymbol{v}_k^q\sum_{i=1}^N\sum_{\{j|A_j \neq \phi, A_j \subseteq \Omega\}}c_j^{\alpha-2}(m_{ij}^q)^2s_{lj}s_{kj}.
\end{align}
We define a matrix $\boldsymbol{B}^q \in \mathbb{R}^{C\times D}$ for the $q$th view as
\begin{align}
\boldsymbol{B}_{ls}^q&=\sum_{i=1}^N\boldsymbol{x}_{is}\sum_{\{j|A_j \neq \phi, A_j \subseteq \Omega\}}c_j^{\alpha-1}(m_{ij}^q)^2s_{lj}\nonumber \\
&=\sum_{i=1}^N\boldsymbol{x}_{is}\sum_{A_j\ni a_i}c_j^{\alpha - 1}(m_{ij}^q)^2, \nonumber\\
&l=1,2,\dots,C, \quad s=1,2,\dots,D, 
\end{align}	
and a matrix $\boldsymbol{H}^q \in \mathbb{R}^{C\times C}$ can be calculated as
\begin{align}
\boldsymbol{H}_{lk}^q 
&= \sum_{i=1}^N\sum_{\{j|A_j \neq \phi, A_j \subseteq \Omega\}}c_j^{\alpha-2}(m_{ij}^q)^2s_{lj}s_{kj}\nonumber\\
&=\sum_{i=1}^N\sum_{A_j \supseteq\{a_k,a_l\}}c_j^{\alpha-2}(m_{ij}^q)^2\nonumber\\
&l,k=1,2,\dots,C.
\end{align}
With these notations, the centers $\boldsymbol{V}^q$ in the $q$th view is the solution of the following linear system
\begin{equation}
\label{eq:v_c}
\boldsymbol{H}^q\cdot\boldsymbol{V}^q=\boldsymbol{B}^q.
\end{equation}

\subsubsection{Update \textbf{w}}

View weights $\boldsymbol{w}$ are updated when $\boldsymbol{M}$ and $\boldsymbol{V}$ are fixed. We introduce $Q$ Lagrange multipliers $\gamma$ and define the Lagrangian functions as follows
\begin{equation}
\mathscr{L}(\boldsymbol{w}, \gamma) = J_{MvLRECM} - \gamma\left(\sum_{q=1}^Qw_q - 1\right).
\end{equation}
By differentiating the Lagrangian with respect to $w_q$ and setting the derivatives to zero, we obtain
\begin{align}
\label{eq:w}
\frac{\partial\mathscr{L}(\boldsymbol{w}, \gamma)}{\partial w_q} &= \sum_{i=1}^N\sum_{\{j|A_j \neq \phi, A_j \subseteq \Omega\}}c_j^\alpha(m_{ij}^q)^2 \nonumber \\
& + \sum_{i=1}^N\delta^2(m_{i\phi}^q)^2 + \eta(1+\ln{w_q}) - \gamma = 0,
\end{align}
where we have
\begin{equation}
\label{eq:R1}
\Psi \triangleq \sum_{i=1}^N\sum_{\{j|A_j \neq \phi, A_j \subseteq \Omega\}}c_j^\alpha(m_{ij}^q)^2 + \sum_{i=1}^N\delta^2(m_{i\phi}^q)^2.
\end{equation}
From Eqs.~(\ref{eq:w})-(\ref{eq:R1}), we thereby obtain 
\begin{equation}
\label{eq:w1}
w_q = e^{\frac{\gamma}{\eta}} \cdot e^{\frac{-\eta - \Psi}{\eta}}.
\end{equation}
Similarly, by setting the derivatives with respect to $\gamma$ to zero, we have
\begin{equation}
\label{eq:gamma}
\frac{\partial\mathscr{L}(\boldsymbol{w},\gamma)}{\partial\gamma} = \sum_{q=1}^Qw_q - 1 = 0.
\end{equation}
Using Eqs.~(\ref{eq:w1})-~(\ref{eq:gamma}), we thus obtain
\begin{equation}
e^{\frac{\gamma}{\eta}} = \frac{1}{\sum\limits_{q=1}^Q e^{\frac{-\eta - \Psi}{\eta}}}.
\end{equation}
Returning in Eq.~(\ref{eq:w1}), we obtain the optimized $\boldsymbol{w}$
\begin{equation}
\label{eq:w_c}
w_q = \frac{e^{\frac{-\eta-\Psi}{\eta}}}{\sum\limits_{q=1}^Qe^{\frac{-\eta-\Psi}{\eta}}}.
\end{equation}

\subsubsection{Update \textbf{M}}
Let us now consider that $\boldsymbol{V}$, $\boldsymbol{Z}$, and $\boldsymbol{w}$ are fixed. We similarly solve the constrained minimization problem with respect to $\boldsymbol{M}$ and the Lagrangian functions are given as follows
\begin{align}
\mathscr{L}(\boldsymbol{M},\boldsymbol{\lambda}) &= J_{MvLRECM} \nonumber\\
&- \sum_{q=1}^Q\sum_{i=1}^N\lambda_i^q\left(\sum_{\{j|A_j \neq \phi, A_j \subseteq \Omega\}}m_{ij}^q+m_{i\phi}^q - 1\right),
\end{align}
where $\boldsymbol{\lambda}$ is the Lagrangian multiplier.  
By differentiating the Lagrangian with respect to the $m_{ij}^q$, $m_{i\phi}^q$, and $\lambda_i^q$, we obtain
\begin{align}
\label{eq:m}
\frac{\partial\mathscr{L}(\boldsymbol{M}^q,\boldsymbol{\lambda}^q)}{\partial m_{ij}^q} &= w_q \cdot \left[d_j^\alpha \cdot 2m_{ij}^q\cdot(d_{ij}^q)^2\right] \nonumber\\
&+ 2\theta(m_{ij}^q - z_{ij}^q) - \lambda_i^q,
\end{align}
\begin{equation}
\label{eq:m1}
\frac{\partial\mathscr{L}(\boldsymbol{M}^q,\boldsymbol{\lambda}^q)}{\partial m_{i\phi}^q} = w_q \cdot \delta^2 \cdot 2m_{i\phi}^q + 2\theta(m_{ij}^q - z_{ij}^q) - \lambda_i^q,
\end{equation}
\begin{equation}
\label{eq:lambda}
\frac{\partial\mathscr{L}(\boldsymbol{M}^q,\boldsymbol{\lambda}^q)}{\partial \lambda_i^q}  = \sum_{\{j|A_j \neq \phi, A_j \subseteq \Omega\}} m_{ij}^q+m_{i\phi}^q - 1.
\end{equation}
Setting Eqs.~(\ref{eq:m})-(\ref{eq:m1}) to zero, we obtain 
\begin{equation}
\label{eq:m2}
m_{ij}^q = \frac{\lambda_i^q + 2\theta z_{ij}^q}{2\left[w_qc_j^\alpha(d_{ij}^q)^2 + \theta\right]},
\end{equation}
\begin{equation}
\label{eq:m3}
m_{i\phi}^q = \frac{\lambda_i^q + 2\theta z_{i\phi}^q}{2\left(w_q\delta^2 + \theta\right)}.
\end{equation}
Using Eqs.(\ref{eq:lambda})-(\ref{eq:m3}), we have
\begin{align}
\sum_{\{j|A_j \neq \phi, A_j \subseteq \Omega\}}\frac{\lambda_i^q + 2\theta z_{ij}^q}{2\left[w_qc_j^\alpha(d_{ij}^q)^2 + \theta\right]} + \frac{\lambda_i^q + 2\theta z_{i\phi}^q}{2\left(w_q\delta^2 + \theta\right)} = 1,
\end{align}
and we thereby obtain
\begin{align}
\lambda_i^q &= 2 \cdot \frac{1-\left(\sum_{\{j|A_j \neq \phi, A_j \subseteq \Omega\}}\frac{\theta z_{ij}^q}{w_qc_j^\alpha(d_{ij}^q)^2+\theta} +\frac{\theta z_{i\phi}^q}{w_q\delta^2+\theta}\right) 
} {\sum_{\{j|A_j \neq \phi, A_j \subseteq \Omega\}}\frac{1}{w_qc_j^\alpha(d_{ij}^q)^2+\theta} +\frac{1}{w_q\delta^2+\theta}} \nonumber \\
\end{align}
where
\begin{equation}
\Delta \triangleq \frac{1-\left(\sum_{\{j|A_j \neq \phi, A_j \subseteq \Omega\}}\frac{\theta z_{ij}^q}{w_qc_j^\alpha(d_{ij}^q)^2+\theta} +\frac{\theta z_{i\phi}^q}{w_q\delta^2+\theta}\right) 
} {\sum_{\{j|A_j \neq \phi, A_j \subseteq \Omega\}}\frac{1}{w_qc_j^\alpha(d_{ij}^q)^2+\theta} +\frac{1}{w_q\delta^2+\theta}}.
\end{equation}
Returning in Eq.~(\ref{eq:m2}), we obtain the optimized solution of $\boldsymbol{M}$
\begin{equation}
\label{eq:m_c1}
m_{ij}^q = \frac{\Delta+\theta z_{ij}^q}{w_qc_j^{\alpha}(d_{ij}^q)^2 + \theta},
\end{equation}
\begin{equation}
\label{eq:m_c2}
m_{i\phi}^q = 1 - \sum_{\{j|A_j \neq \phi, A_j \subseteq \Omega\}}m_{ij}^q.
\end{equation}

\subsubsection{Update \textbf{Z}}
The optimization of $\boldsymbol{Z}_i$ is only relevant to $\boldsymbol{M}_i$, so we optimize the objective functions as follows
\begin{align}
\label{eq:NNM}
\min_{\boldsymbol{Z}} \rho\parallel \boldsymbol{Z}_i \parallel_* + \parallel\boldsymbol{M}_i - \boldsymbol{Z}_i\parallel_F^2,~~~~
i=1,2,\dots,N.
\end{align}

The nuclear norm of $\boldsymbol{Z}$ is equally calculated as the sum of its singular values~\cite{fazel2002matrix}, (\emph{i.e.}, $\parallel \boldsymbol{Z} \parallel_*=\sum_i\sigma_i(\boldsymbol{Z})$), 
where $\sigma_i(\boldsymbol{Z})$ denotes the $i$th singular value of $\boldsymbol{Z}$ in a descending order. We define
\begin{align}
\boldsymbol{M}=\boldsymbol{A}\bm{\varSigma} \boldsymbol{B}^T, ~\bm{\varSigma}=\begin{pmatrix} diag(\sigma_1,\sigma_2,\dots,\sigma_N)\\\boldsymbol{0} \end{pmatrix}, \nonumber\\
\boldsymbol{Z}=\boldsymbol{A}\boldsymbol{D} \boldsymbol{B}^T, ~\boldsymbol{D} =\begin{pmatrix} diag(d_1,d_2,\dots,d_N)\\\boldsymbol{0} \end{pmatrix} \nonumber.
\end{align}

By imposing a soft-thresholding operation on $\bm{\varSigma}$~\cite{fazel2002matrix}, the optimized solution of $\boldsymbol{Z}$ is calculated 
\begin{equation}
\label{eq:z_c}
\boldsymbol{Z}=\boldsymbol{A}\boldsymbol{S}_{\frac{\rho}{2}}(\boldsymbol{D}) \boldsymbol{B}^T,
\end{equation}
where $\boldsymbol{S}_{\frac{\rho}{2}}(\boldsymbol{D})$ is the soft-thresholding function on diagonal matrix $\boldsymbol{D}$ with Parameter $\frac{\rho}{2}$, where the diagonal element $d_i$ is
\begin{align}
d_i = \max\left(\sigma_i-\frac{\rho}{2}, 0\right), \quad i=1,2,\dots,N.
\end{align}
The proof of this optimization method, NNM, is shown in Sec.~\ref{sec:NNM} in the supplementary. More importantly, a detailed algorithmic description is presented  in \textbf{Algorithm~\ref{alg:S1}}. 
\begin{algorithm}[t]
	\caption{MvLRECM}
	\label{alg:S1}
	\textbf{Input}: Data for $Q$ views $\boldsymbol{X}^1,\boldsymbol{X}^2,\dots \boldsymbol{X}^Q \in \mathbb{R}^{N*D} $, the number
	of singleton cluster $C$, the maximum number of iteration $I_{max}$.\\
	\textbf{Parameter}: $\alpha$, $\delta$, $\theta$, and $\eta$\\
	\textbf{Output}: The solution of clustering
	\begin{algorithmic}[1] 
		\STATE Initialize the centers of singleton clusters: randomly select $C$ objects in the dataset to be the centers,
		\STATE Initialize the weight for each view, $w_q=1/Q$,
		\STATE calculte $\rho=2^{-\frac{C}{2}}$
		\WHILE{$|J_{old}-J|<10^{-4}$ or $I_{max}$ reached.}
		\STATE $J_{old}=J$
		\STATE Fix $\boldsymbol{M}$ and $\boldsymbol{w}$, update $\boldsymbol{V}$ by using Eq.~(\ref{eq:v_c}),
		\STATE Fix $\boldsymbol{M}$ and $\boldsymbol{V}$, update $\boldsymbol{w}$ by using Eq.~(\ref{eq:w_c}),
		\STATE Fix $\boldsymbol{V}$, $\boldsymbol{w}$, and $\boldsymbol{Z}$, update $\boldsymbol{M}$ by using Eqs.~(\ref{eq:m_c1})-(\ref{eq:m_c2}),
		\STATE Fix $\boldsymbol{M}$, update $\boldsymbol{Z}$ by using Eq.~(\ref{eq:z_c}),
		\STATE $\boldsymbol{M}_i=norm(\boldsymbol{Z}_i)\quad$
		\STATE Calculate $J_{MvLRECM}$.
		\ENDWHILE
		\STATE Unified mass matrix $\overline{\boldsymbol{M}}=\sum_{Q}^{q=1}w_q \times \boldsymbol{M}^q$,
		\STATE  When $\overline{m}_{ij} = \max (\overline{\boldsymbol{m}}_i)$, the $i$th object $\in A_j$. 
		\STATE \textbf{return} Unified mass matrix $\overline{\boldsymbol{M}}$, and the solution of clustering
	\end{algorithmic}
\end{algorithm}	

\begin{figure*}[b]
	\centering
	\subfloat[\label{fig:a1}]{
		\includegraphics[width=0.28\linewidth]{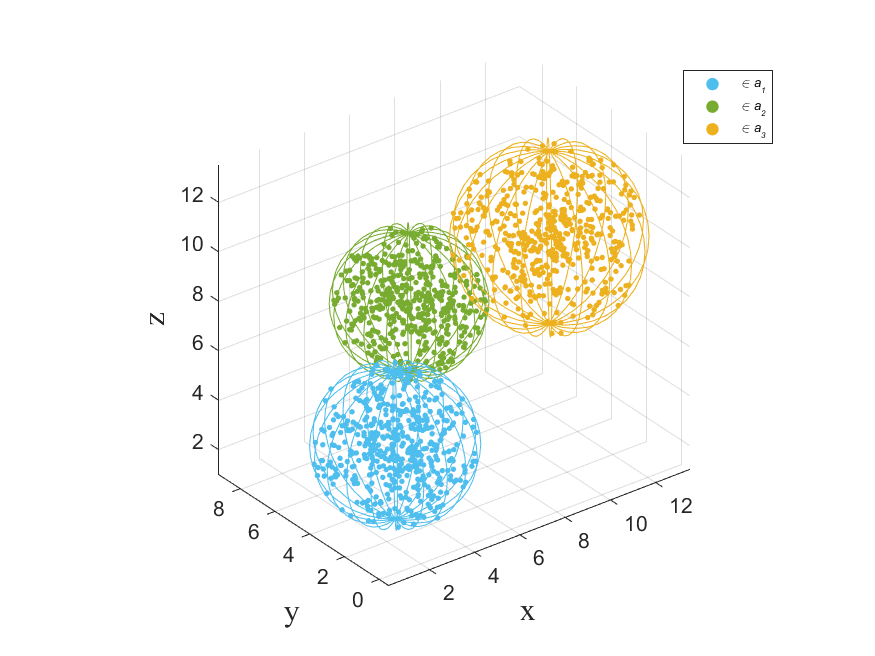}}
	\subfloat[\label{fig:a2}]{
		\includegraphics[width=0.28\linewidth]{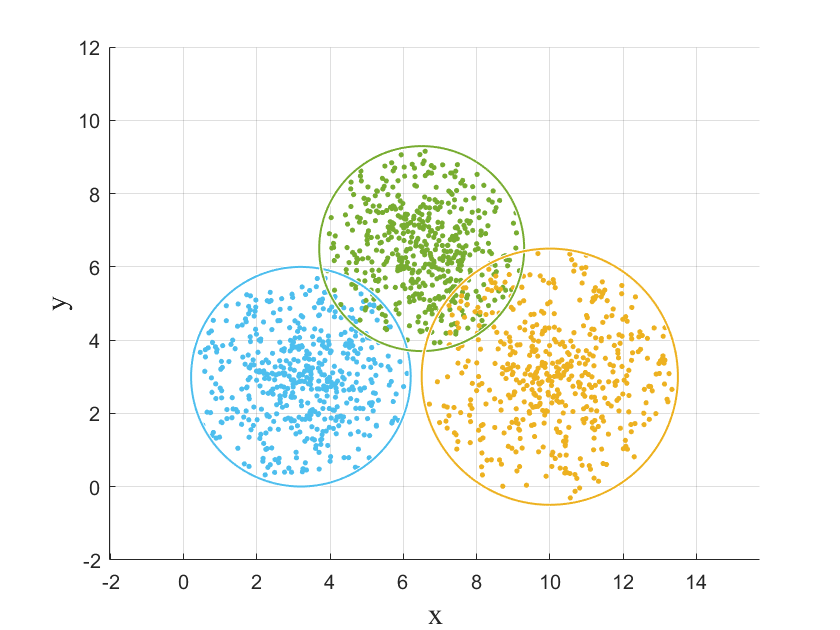}}
	\subfloat[\label{fig:a3}]{
		\includegraphics[width=0.28\linewidth]{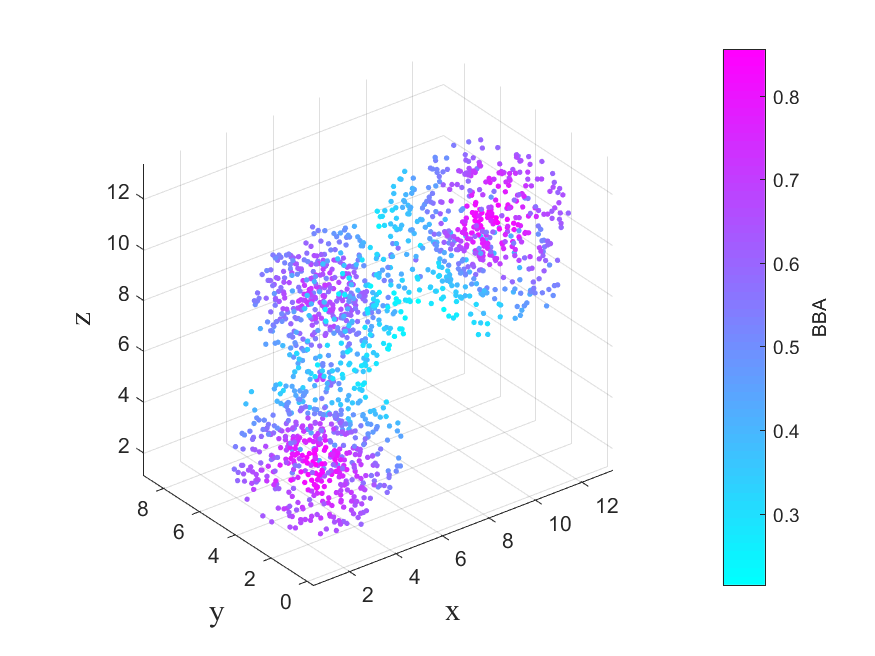}}
	\\
	\subfloat[\label{fig:b1}]{
		\includegraphics[width=0.28\linewidth]{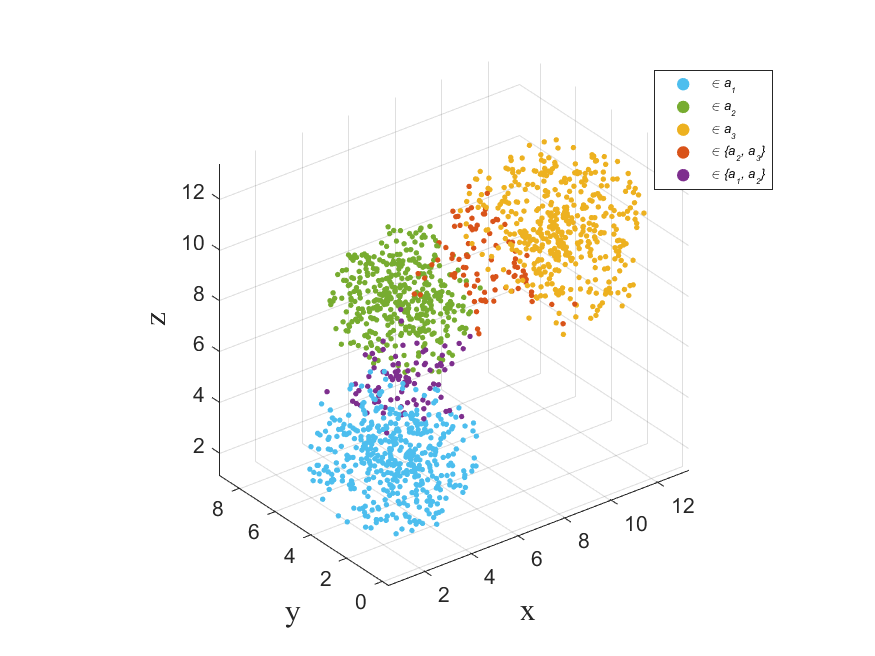}}
	\subfloat[\label{fig:b2}]{
		\includegraphics[width=0.28\linewidth]{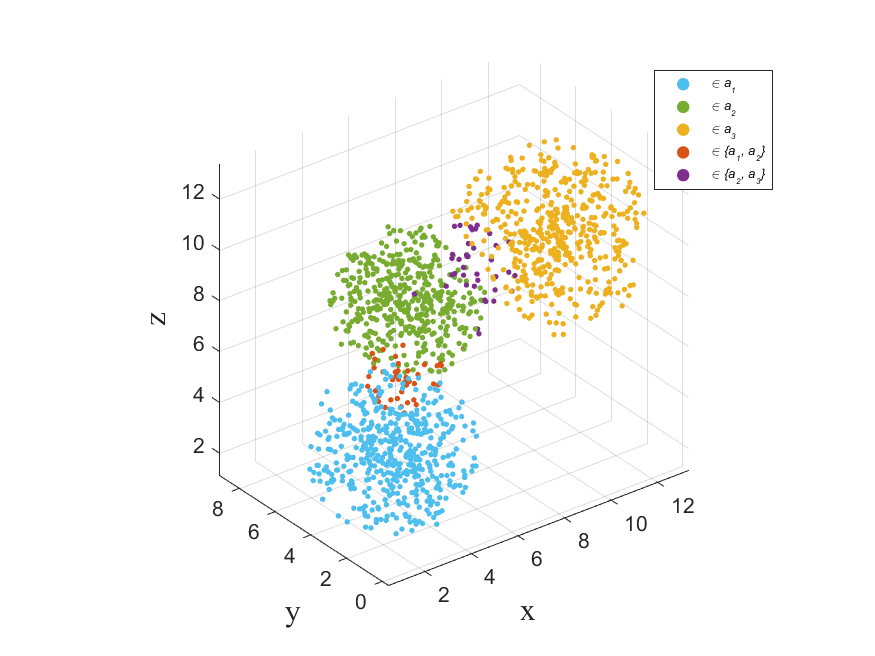}}
	\subfloat[\label{fig:b3}]{
		\includegraphics[width=0.28\linewidth]{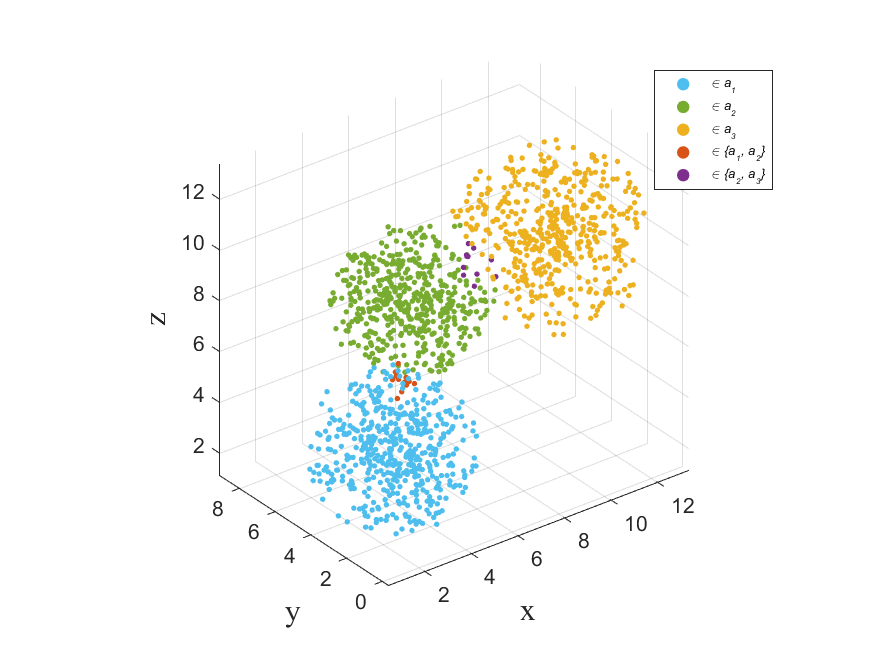}}
	\\
	\subfloat[\label{fig:c1}]{
		\includegraphics[width=0.28\linewidth]{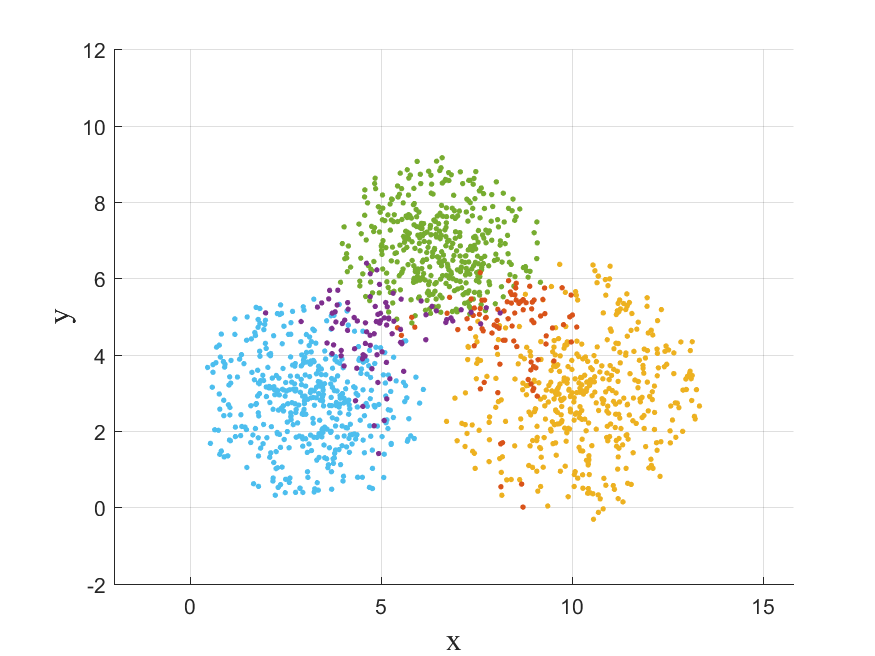}}
	\subfloat[\label{fig:c2}]{
		\includegraphics[width=0.28\linewidth]{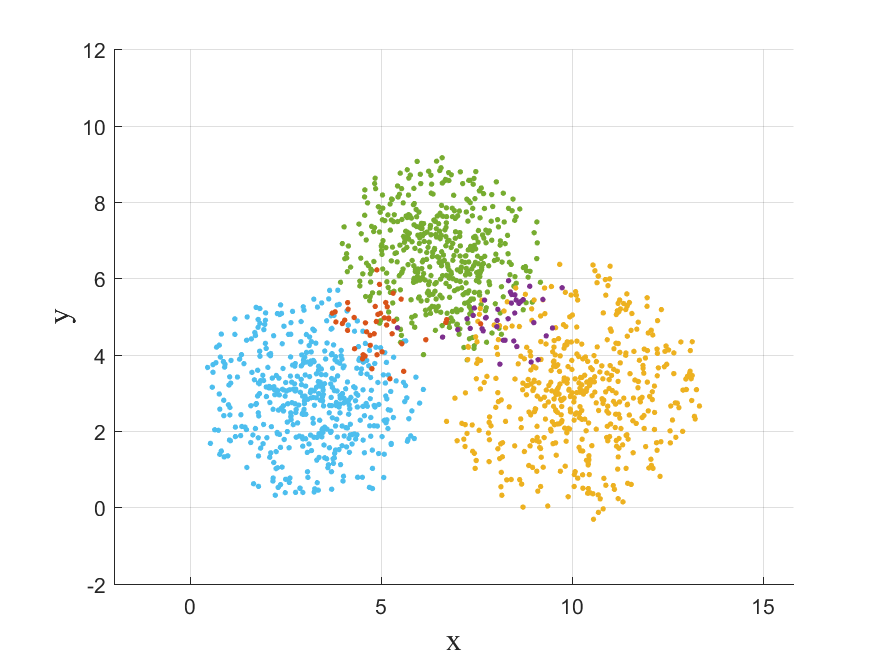}}
	\subfloat[\label{fig:c3}]{
		\includegraphics[width=0.28\linewidth]{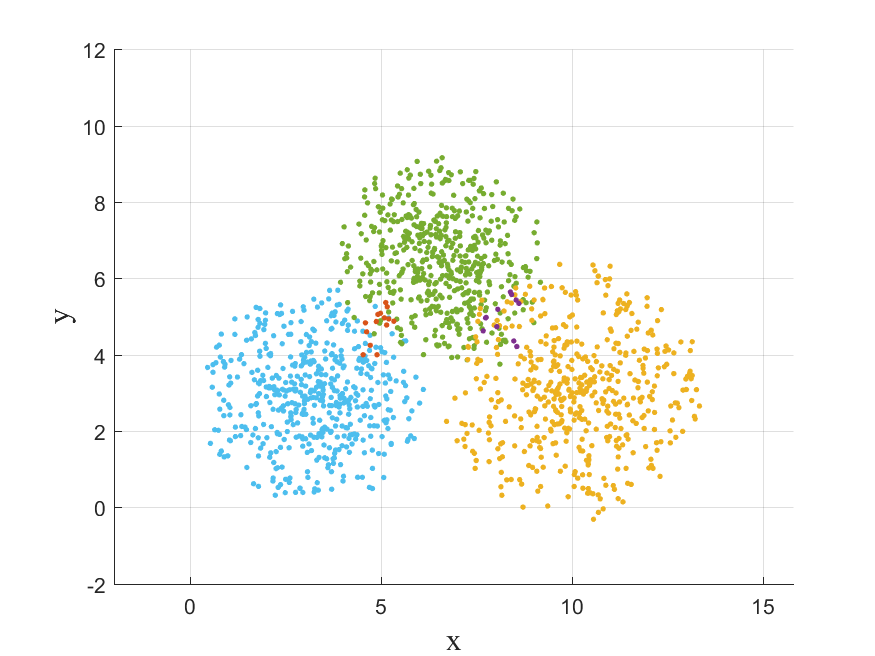}}
	\caption{Results on the 3DBall dataset, (a) the original 3D distribution, (b) the original xy-view, (c) the masses distribution of MvLREVM ($\alpha=2$),
		(d),(g) MvLREVM ($\alpha=1$) , 
		(e),(h) MvLREVM ($\alpha=2$), 
		(f),(i) MvLREVM ($\alpha=3$).}
	\label{fig:toy} 
\end{figure*}

\section{EXPERIMENTS}
\label{sec:exp}
In the experiments, we evaluate the performance of MvLRECM on one toy dataset and seven real-world datasets. 
All the experiments run on a computer with 8 Intel i9-9900k@4.8Ghz processors with 16GB DDR4 memory. The operating system is a 64-bits Windows 10 pro. We used Matlab 2020b to implement and test the methods.

In this section, we first introduce the extensions of evaluation metrics to evaluate hard, fuzzy, and credal partitions simultaneously,
and then give an example on the toy dataset to show how MvLRECM characterizes imprecision.
Finally, we present the complete results and additional analysis of the experiments on the real-world dataset.


\subsection{Metrics Study}
\label{sec:Smec}
Some evaluation metrics are used to achieve a comprehensive measurement for clustering: clustering accuracy (ACC), Normalized Mutual Information (NMI), Purity, Precision, Recall, F-score, and Rand Index (RI). However, their calculations are based on given hard and fuzzy partitions. They cannot compare credal partition to ground truth because they cannot handle the objects which are assigned to meta-clusters.
Thus we introduce how to evaluate the performance in meta-clusters, to apply these metrics for hard, fuzzy, and credal partitions simultaneously.
\begin{table}[h]\footnotesize
	\center
	\caption{pair confusion matrix in the credal partition.}
	\label{tab:STFPN}  
	\begin{tabular}{cccc}
		\toprule
		& True Label & Classical & Proposed \\
		\midrule
		TP & $\mathcal{G}_i = \mathcal{G}_j$ & $\mathcal{S}_i = \mathcal{S}_j$ & $\mathcal{S}_i \cap \mathcal{S}_j \ne \emptyset$\\
		FP & $\mathcal{G}_i \ne \mathcal{G}_j$ & $\mathcal{S}_i = \mathcal{S}_j$ & $| \mathcal{S}_i \cup \mathcal{S}_j | = 1 \text{ and } \mathcal{S}_i \ne \emptyset \text{ and } \mathcal{S}_j \ne \emptyset$ \\
		TN & $\mathcal{G}_i \ne \mathcal{G}_j$ & $\mathcal{S}_i \ne \mathcal{S}_j$ & $| \mathcal{S}_i \cup \mathcal{S}_j | \ne 1 \text{ or } \mathcal{S}_i = \emptyset \text{ or } \mathcal{S}_j = \emptyset$ \\
		FN & $\mathcal{G}_i = \mathcal{G}_j$ & $\mathcal{S}_i \ne \mathcal{S}_j$ & $\mathcal{S}_i \cap \mathcal{S}_j \ne \emptyset$ \\
		\bottomrule
	\end{tabular}
\end{table}

\begin{table*}[hb]\scriptsize
	\caption{Clustering performance on the real-world datasets (mean $\pm$ standard deviation).}
	\label{tab:RealW}
	\resizebox{\linewidth}{!}{
		\begin{tabular}{ccccccccccc}
			\toprule
			Datasets & \multicolumn{2}{c}{Methods} & ACC & NMI & Purity &  F-Score & Precision & Recall & RI & IR\\
			\midrule
			\multirow{11}*{abalone} & \multirow{1}*{Single-view} & \boldmath$\operatorname{ECM}$ & 0.4225$\pm$0.0409 & 0.0438$\pm$0.0294 & 0.4307$\pm$0.0402 & 0.6622$\pm$0.0304 & 0.6869$\pm$0.0380 & 0.6353$\pm$0.0231 & 0.7816$\pm$0.0219 & 0.3698$\pm$0.0110 \\
			\cmidrule(lr){2-11}
			& \multirow{10}*{Multi-view} &\boldmath$\operatorname{CDMGC}$ & 0.3776$\pm$0.0011 & 0.0180$\pm$0.0027 & 0.3848$\pm$0.0028 & 0.4977$\pm$0.0000 & 0.3357$\pm$0.0002 & \textbf{0.9655$\pm$0.0000} & 0.3532$\pm$0.0024 & 0.0000$\pm$0.0000 \\ 
			& & \boldmath$\operatorname{CGD}$ & 0.3637$\pm$0.0000 & 0.0045$\pm$0.0000 & 0.3735$\pm$0.0000 & 0.3427$\pm$0.0000 & 0.3369$\pm$0.0000 & 0.3487$\pm$0.0000 & 0.5522$\pm$0.0000 & 0.0000$\pm$0.0000 \\ 
			& & \boldmath$\operatorname{Co-FW-MVFCM}$ & 0.4880$\pm$0.0000 & 0.0916$\pm$0.0000 & 0.4897$\pm$0.0000 & 0.4231$\pm$0.0000 & 0.3934$\pm$0.0000 & 0.4577$\pm$0.0000 & 0.5822$\pm$0.0000 & 0.0000$\pm$0.0000 \\ 
			& & \boldmath$\operatorname{CoFKM}$ & 0.5139$\pm$0.0000 & 0.1206$\pm$0.0000 & 0.5149$\pm$0.0000 & 0.4578$\pm$0.0000 & 0.4092$\pm$0.0000 & 0.5195$\pm$0.0000 & 0.5880$\pm$0.0000 & 0.0000$\pm$0.0000 \\ 
			& & \boldmath$\operatorname{LMSC}$ & 0.4763$\pm$0.0000 & 0.1010$\pm$0.0000 & 0.4839$\pm$0.0000 & 0.4068$\pm$0.0000 & 0.4056$\pm$0.0000 & 0.4080$\pm$0.0000 & 0.6017$\pm$0.0000 & 0.0000$\pm$0.0000 \\ 
			& & \boldmath$\operatorname{LTMSC}$ & 0.5165$\pm$0.0000 & 0.1180$\pm$0.0000 & 0.5201$\pm$0.0000 & 0.4264$\pm$0.0000 & 0.4211$\pm$0.0000 & 0.4318$\pm$0.0000 & 0.6111$\pm$0.0000 & 0.0000$\pm$0.0000 \\ 
			& & \boldmath$\operatorname{MCSSC}$ & 0.4590$\pm$0.0000 & 0.0534$\pm$0.0000 & 0.4590$\pm$0.0000 & 0.3931$\pm$0.0000 & 0.3705$\pm$0.0000 & 0.4186$\pm$0.0000 & 0.5673$\pm$0.0000 & 0.0000$\pm$0.0000 \\ 
			& & \boldmath$\operatorname{TBGL-MVC}$ & 0.4669$\pm$0.0000 & 0.1093$\pm$0.0000 & 0.4902$\pm$0.0000 & 0.4928$\pm$0.0000 & 0.3714$\pm$0.0000 & 0.7322$\pm$0.0000 & 0.4955$\pm$0.0000 & 0.0000$\pm$0.0000 \\ 
			& & \boldmath$\operatorname{MVASM}$ & - & - & - & - & - & - & - & - \\ 
			& & \boldmath$\operatorname{MvLRECM}$ & \textbf{0.6301$\pm$0.0510} & \textbf{0.1391$\pm$0.0007} & \textbf{0.6785$\pm$0.0677} & \textbf{0.6933$\pm$0.0000} & \textbf{0.7250$\pm$0.0000} & 0.6643$\pm$0.0000 & \textbf{0.8033$\pm$0.0000} & \textbf{0.2851$\pm$0.0006} \\ 
			\cmidrule(lr){1-11}

			\multirow{11}*{contraceptive} & \multirow{1}*{Single-view} & \boldmath$\operatorname{ECM}$ & 0.3469$\pm$0.0015 & 0.0168$\pm$0.0019 & 0.4177$\pm$0.0008 & 0.5828$\pm$0.0813 & 0.5530$\pm$0.0671 & 0.6160$\pm$0.0974 & 0.6904$\pm$0.0513 & 0.2795$\pm$0.0003 \\
			\cmidrule(lr){2-11}
			& \multirow{10}*{Multi-view} &\boldmath$\operatorname{CDMGC}$ & 0.4460$\pm$0.0000 & 0.0202$\pm$0.0000 & 0.4467$\pm$0.0000 & 0.4376$\pm$0.0000 & 0.3657$\pm$0.0000 & 0.5446$\pm$0.0000 & 0.5053$\pm$0.0000 & 0.0000$\pm$0.0000 \\ 
			& & \boldmath$\operatorname{CGD}$ & 0.4338$\pm$0.0000 & \textbf{0.0352$\pm$0.0000} & 0.4494$\pm$0.0000 & 0.3630$\pm$0.0000 & 0.3730$\pm$0.0000 & 0.3535$\pm$0.0000 & 0.5616$\pm$0.0000 & 0.0000$\pm$0.0000 \\ 
			& & \boldmath$\operatorname{Co-FW-MVFCM}$ & 0.3944$\pm$0.0000 & 0.0320$\pm$0.0000 & 0.4569$\pm$0.0000 & 0.3652$\pm$0.0000 & 0.3714$\pm$0.0000 & 0.3592$\pm$0.0000 & 0.5588$\pm$0.0000 & 0.0000$\pm$0.0000 \\ 
			& & \boldmath$\operatorname{CoFKM}$ & 0.3938$\pm$0.0000 & 0.0330$\pm$0.0000 & 0.4562$\pm$0.0004 & 0.3671$\pm$0.0000 & 0.3714$\pm$0.0001 & 0.3632$\pm$0.0000 & 0.5580$\pm$0.0002 & 0.0000$\pm$0.0000 \\ 
			& & \boldmath$\operatorname{LMSC}$ & 0.3754$\pm$0.0000 & 0.0199$\pm$0.0001 & 0.4270$\pm$0.0000 & 0.3957$\pm$0.0000 & 0.3567$\pm$0.0000 & 0.4444$\pm$0.0000 & 0.5204$\pm$0.0001 & 0.0000$\pm$0.0000 \\ 
			& & \boldmath$\operatorname{LTMSC}$ & 0.3761$\pm$0.0000 & 0.0166$\pm$0.0000 & 0.4270$\pm$0.0000 & 0.4072$\pm$0.0000 & 0.3541$\pm$0.0000 & 0.4791$\pm$0.0000 & 0.5071$\pm$0.0000 & 0.0000$\pm$0.0000 \\ 
			& & \boldmath$\operatorname{MCSSC}$ & 0.3652$\pm$0.0000 & 0.0042$\pm$0.0000 & 0.4270$\pm$0.0000 & 0.3467$\pm$0.0000 & 0.3537$\pm$0.0000 & 0.3401$\pm$0.0000 & 0.5472$\pm$0.0000 & 0.0000$\pm$0.0000 \\ 
			& & \boldmath$\operatorname{TBGL-MVC}$ & 0.3788$\pm$0.0000 & 0.0314$\pm$0.0000 & 0.4331$\pm$0.0000 & 0.3787$\pm$0.0000 & 0.3588$\pm$0.0000 & 0.4009$\pm$0.0000 & 0.5351$\pm$0.0000 & 0.0000$\pm$0.0000 \\ 
			& & \boldmath$\operatorname{MVASM}$ & 0.3788$\pm$0.0000 & 0.0194$\pm$0.0000 & 0.4270$\pm$0.0000 & 0.3586$\pm$0.0000 & 0.3554$\pm$0.0000 & 0.3619$\pm$0.0000 & 0.5425$\pm$0.0000 & 0.0000$\pm$0.0000 \\ 
			& & \boldmath$\operatorname{MvLRECM}$ & \textbf{0.5709$\pm$0.0606} & 0.0205$\pm$0.0000 & \textbf{0.7515$\pm$0.1279} & \textbf{0.6689$\pm$0.0000} & \textbf{0.6333$\pm$0.0000} & \textbf{0.7543$\pm$0.0000} & \textbf{0.7361$\pm$0.0000} & \textbf{0.1283$\pm$0.0000} \\ 
			\cmidrule(lr){1-11}

			\multirow{11}*{foresttype} & \multirow{1}*{Single-view} & $\mathbf{ECM}$ & 0.0101$\pm$0.0012 & 0.1895$\pm$0.0665 & 0.0146$\pm$0.0019 & 0.0005$\pm$0.0001 & \textbf{0.8803$\pm$0.0809} & 0.0002$\pm$0.0001 & 0.7178$\pm$0.0000 & 0.0000$\pm$0.0000 \\
			\cmidrule(lr){2-11}
			& \multirow{10}*{Multi-view} & 
			\boldmath$\operatorname{CDMGC}$ & 0.3674$\pm$0.0052 & 0.0782$\pm$0.0076 & 0.4114$\pm$0.0049 & 0.4360$\pm$0.0018 & 0.2901$\pm$0.0019 & \textbf{0.7675$\pm$0.0452} & 0.3589$\pm$0.0070 & 0.0000$\pm$0.0000 \\
			&  & \boldmath$\operatorname{CGD}$ & 0.6883$\pm$0.0000 & 0.4601$\pm$0.0000 & 0.6883$\pm$0.0000 & 0.5583$\pm$0.0000 & 0.5167$\pm$0.0000 & 0.6072$\pm$0.0000 & 0.7288$\pm$0.0000 & 0.0000$\pm$0.0000 \\
			
			&  & \boldmath$\operatorname{Co-FW-MVFCM}$ & 0.7323$\pm$0.0000 & 0.4897$\pm$0.0000 & 0.7323$\pm$0.0000 & 0.5970$\pm$0.0000 & 0.5727$\pm$0.0000 & 0.6234$\pm$0.0000 & 0.7624$\pm$0.0000 & 0.0000$\pm$0.0000 \\
			&  & \boldmath$\operatorname{CoFKM}$ & 0.6381$\pm$0.0257 & 0.2964$\pm$0.0208 & 0.6384$\pm$0.0250 & 0.5406$\pm$0.0052 & 0.4478$\pm$0.0106 & 0.6789$\pm$0.0000 & 0.6724$\pm$0.0116 & 0.0000$\pm$0.0000 \\
			& & \boldmath$\operatorname{LMSC}$ & 0.5182$\pm$0.0030 & 0.3711$\pm$0.0051 & 0.6405$\pm$0.0020 & 0.5060$\pm$0.0031 & 0.4990$\pm$0.0018 & 0.5133$\pm$0.0039 & 0.7171$\pm$0.0010 & 0.0000$\pm$0.0000 \\ 
			&  & \boldmath$\operatorname{LTMSC}$ & 0.5605$\pm$0.0085 & 0.4149$\pm$0.0047 & 0.6209$\pm$0.0017 & 0.4976$\pm$0.0023 & 0.5007$\pm$0.0017 & 0.4946$\pm$0.0049 & 0.7181$\pm$0.0010 & 0.0000$\pm$0.0000 \\
			&  & \boldmath$\operatorname{MCSSC}$ & 0.6157$\pm$0.0000 & 0.3123$\pm$0.0000 & 0.6233$\pm$0.0000 & 0.4560$\pm$0.0000 & 0.4772$\pm$0.0000 & 0.4366$\pm$0.0000 & 0.7059$\pm$0.0000 & 0.0000$\pm$0.0000 \\
			&  & \boldmath$\operatorname{TBGL-MVC}$ & 0.3652$\pm$0.0000 & 0.0542$\pm$0.0000 & 0.3977$\pm$0.0000 & 0.3932$\pm$0.0000 & 0.2772$\pm$0.0000 & 0.6758$\pm$0.0000 & 0.4110$\pm$0.0000 & 0.0000$\pm$0.0000 \\
			& & \boldmath$\operatorname{MVASM}$ & 0.3270$\pm$0.0000 & 0.0153$\pm$0.0000 & 0.3862$\pm$0.0000 & 0.2976$\pm$0.0000 & 0.2856$\pm$0.0000 & 0.3106$\pm$0.0000 & 0.5861$\pm$0.0000 & 0.0000$\pm$0.0000 \\ 
			&  & \boldmath$\operatorname{MvLRECM}$ & \textbf{0.7552$\pm$0.0009} & \textbf{0.5184$\pm$0.0034} & \textbf{0.7572$\pm$0.0009} & \textbf{0.6100$\pm$0.0022} & 0.6272$\pm$0.0021 & 0.5947$\pm$0.0015 & \textbf{0.7857$\pm$0.0011} & \textbf{0.0000$\pm$0.0000} \\
			\cmidrule(lr){1-11}

			\multirow{11}*{hayes} & \multirow{1}*{Single-view} & \boldmath$\operatorname{ECM}$ & 0.2608$\pm$0.0127 & 0.0225$\pm$0.0031 & 0.3045$\pm$0.0080 & 0.3803$\pm$0.0100 & 0.6554$\pm$0.0253 & 0.2670$\pm$0.0101 & 0.6987$\pm$0.0060 & 0.1878$\pm$0.0450 \\
			\cmidrule(lr){2-11}
			& \multirow{10}*{Multi-view} & 
			\boldmath$\operatorname{CDMGC}$ & 0.4015$\pm$0.0000 & 0.0123$\pm$0.0000 & 0.4394$\pm$0.0000 & 0.3445$\pm$0.0000 & 0.3439$\pm$0.0000 & 0.3451$\pm$0.0000 & 0.5372$\pm$0.0084 & 0.0000$\pm$0.0000 \\
			&  & \boldmath$\operatorname{CGD}$ & 0.3864$\pm$0.0000 & 0.0084$\pm$0.0000 & 0.4242$\pm$0.0000 & 0.3574$\pm$0.0000 & 0.3418$\pm$0.0000 & 0.3745$\pm$0.0000 & 0.5350$\pm$0.0000 & 0.0000$\pm$0.0000 \\
			
			&  & \boldmath$\operatorname{Co-FW-MVFCM}$ & \textbf{0.4394$\pm$0.0000} & 0.0280$\pm$0.0000 & 0.4621$\pm$0.0000 & 0.3476$\pm$0.0000 & 0.3548$\pm$0.0000 & 0.3407$\pm$0.0000 & 0.5585$\pm$0.0000 & 0.0000$\pm$0.0000 \\
			&  & \boldmath$\operatorname{CoFKM}$ & 0.4242$\pm$0.0000 & 0.0238$\pm$0.0000 & \textbf{0.4697$\pm$0.0000} & 0.3465$\pm$0.0000 & 0.3554$\pm$0.0000 & 0.3380$\pm$0.0000 & 0.5598$\pm$0.0000 & 0.0000$\pm$0.0000 \\
			& & \boldmath$\operatorname{LMSC}$ & 0.3788$\pm$0.0000 & 0.0087$\pm$0.0000 & 0.4015$\pm$0.0000 & 0.3448$\pm$0.0000 & 0.3439$\pm$0.0000 & 0.3457$\pm$0.0000 & 0.5464$\pm$0.0000 & 0.0000$\pm$0.0000 \\ 
			&  & \boldmath$\operatorname{LTMSC}$ & 0.3837$\pm$0.0038 & 0.0085$\pm$0.0003 & 0.4161$\pm$0.0020 & 0.3350$\pm$0.0004 & 0.3421$\pm$0.0003 & 0.3283$\pm$0.0006 & 0.5501$\pm$0.0001 & 0.0000$\pm$0.0000 \\
			&  & \boldmath$\operatorname{MCSSC}$ & 0.4167$\pm$0.0000 & 0.0411$\pm$0.0000 & 0.4470$\pm$0.0000 & 0.3970$\pm$0.0000 & 0.3519$\pm$0.0000 & 0.4553$\pm$0.0000 & 0.5224$\pm$0.0000 & 0.0000$\pm$0.0000 \\
			&  & \boldmath$\operatorname{TBGL-MVC}$ & 0.4242$\pm$0.0000 & 0.0490$\pm$0.0000 & 0.4242$\pm$0.0000 & 0.4074$\pm$0.0000 & 0.3537$\pm$0.0000 & \textbf{0.4804$\pm$0.0000} & 0.5176$\pm$0.0000 & 0.0000$\pm$0.0000 \\
			& & \boldmath$\operatorname{MVASM}$ & 0.4394$\pm$0.0000 & \textbf{0.1170$\pm$0.0000} & 0.4697$\pm$0.0000 & 0.4615$\pm$0.0000 & 0.4218$\pm$0.0000 & \textbf{0.5095$\pm$0.0000} & 0.5895$\pm$0.0000 & 0.0000$\pm$0.0000 \\ 
			&  & \boldmath$\operatorname{MvLRECM}$ & 0.3295$\pm$0.0127 & \textbf{0.0601$\pm$0.0054} & 0.3485$\pm$0.0083 & \textbf{0.4886$\pm$0.0222} & \textbf{0.8386$\pm$0.0504} & 0.3772$\pm$0.0186 & \textbf{0.7509$\pm$0.0123} & \textbf{0.1364$\pm$0.0097} \\
			\cmidrule(lr){1-11}

			\multirow{11}*{Ionosphere} & \multirow{1}*{Single-view} & \boldmath$\operatorname{ECM}$ & 0.5481$\pm$0.0164 & 0.0081$\pm$0.0041 & 0.6410$\pm$0.0000 & 0.5461$\pm$0.0176 & 0.5526$\pm$0.0017 & 0.5332$\pm$0.0224 & 0.5173$\pm$0.0027 & 0.0144$\pm$0.0008 \\
			\cmidrule(lr){2-11}
			& \multirow{10}*{Multi-view} & 
			\boldmath$\operatorname{CDMGC}$ & 0.5783$\pm$0.0000 & 0.0326$\pm$0.0000 & 0.6410$\pm$0.0000 & 0.6533$\pm$0.0000 & 0.5283$\pm$0.0000 & \textbf{0.8558$\pm$0.0000} & 0.5109$\pm$0.0000 & 0.0000$\pm$0.0000 \\
			&  & \boldmath$\operatorname{CGD}$ & 0.6673$\pm$0.0012 & 0.1170$\pm$0.0023 & 0.6673$\pm$0.0012 & 0.5738$\pm$0.0009 & 0.5920$\pm$0.0008 & 0.5566$\pm$0.0010 & 0.5547$\pm$0.0008 & 0.0000$\pm$0.0000 \\
			
			&  & \boldmath$\operatorname{Co-FW-MVFCM}$ & 0.7009$\pm$0.0000 & 0.1063$\pm$0.0000 & 0.7009$\pm$0.0000 & 0.5989$\pm$0.0000 & 0.6156$\pm$0.0000 & 0.5831$\pm$0.0000 & 0.5795$\pm$0.0000 & 0.0000$\pm$0.0000 \\
			&  & \boldmath$\operatorname{CoFKM}$ & 0.6581$\pm$0.0000 & 0.0460$\pm$0.0000 & 0.6581$\pm$0.0000 & 0.5863$\pm$0.0000 & 0.5789$\pm$0.0000 & 0.5939$\pm$0.0000 & 0.5487$\pm$0.0000 & 0.0000$\pm$0.0000 \\
			& & \boldmath$\operatorname{LMSC}$ & - & - & - & - & - & - & - & - \\ 
			&  & \boldmath$\operatorname{LTMSC}$ & \textbf{0.8946$\pm$0.0000} & 0.4946$\pm$0.0000 & \textbf{0.8946$\pm$0.0000} & \textbf{0.8222$\pm$0.0000} & \textbf{0.8325$\pm$0.0000} & 0.8121$\pm$0.0000 & \textbf{0.8109$\pm$0.0000} & 0.0000$\pm$0.0000 \\
			&  & \boldmath$\operatorname{MCSSC}$ & 0.5783$\pm$0.0000 & 0.0656$\pm$0.0000 & 0.6410$\pm$0.0000 & 0.5540$\pm$0.0000 & 0.5442$\pm$0.0000 & 0.5641$\pm$0.0000 & 0.5109$\pm$0.0000 & 0.0000$\pm$0.0000 \\
			&  & \boldmath$\operatorname{TBGL-MVC}$ & 0.6097$\pm$0.0000 & \textbf{0.1875$\pm$0.0000} & 0.6410$\pm$0.0000 & 0.5871$\pm$0.0000 & 0.5495$\pm$0.0000 & 0.6303$\pm$0.0000 & 0.5227$\pm$0.0000 & 0.0000$\pm$0.0000 \\
			& & \boldmath$\operatorname{MVASM}$ & 0.6667$\pm$0.0000 & 0.0854$\pm$0.0000 & 0.6667$\pm$0.0000 & 0.5703$\pm$0.0000 & 0.5930$\pm$0.0000 & 0.5494$\pm$0.0000 & 0.5543$\pm$0.0000 & 0.0000$\pm$0.0000 \\ 
			&  & \boldmath$\operatorname{MvLRECM}$ & 0.7094$\pm$0.0009 & 0.1277$\pm$0.0007 & 0.7094$\pm$0.0009 & 0.6028$\pm$0.0009 & 0.6244$\pm$0.0006 & 0.5826$\pm$0.0011 & 0.5865$\pm$0.0007 & \textbf{0.0000$\pm$0.0000} \\
			\cmidrule(lr){1-11}

			\multirow{11}*{segment} & \multirow{1}*{Single-view} & \boldmath$\operatorname{ECM}$ & 0.0881$\pm$0.0112 & 0.4511$\pm$0.0885 & 0.1005$\pm$0.0141 & 0.0506$\pm$0.0098 & 0.8240$\pm$0.0756 & 0.0262$\pm$0.0052 & 0.8604$\pm$0.0009 & 0.1452$\pm$0.0029 \\ 
			
			\cmidrule(lr){2-11}
			& \multirow{10}*{Multi-view} &\boldmath$\operatorname{CDMGC}$ & 0.4420$\pm$0.0297 & 0.3351$\pm$0.0292 & 0.4446$\pm$0.0294 & 0.3522$\pm$0.0112 & 0.2490$\pm$0.0136 & \textbf{0.9655$\pm$0.0000} & 0.6847$\pm$0.0385 & 0.0000$\pm$0.0000 \\ 
			& & \boldmath$\operatorname{CGD}$ & 0.5442$\pm$0.0000 & 0.3198$\pm$0.0006 & 0.5935$\pm$0.0000 & 0.5400$\pm$0.0007 & 0.4165$\pm$0.0007 & 0.7680$\pm$0.0006 & 0.8137$\pm$0.0004 & 0.0000$\pm$0.0000 \\ 
			& & \boldmath$\operatorname{Co-FW-MVFCM}$ & 0.4853$\pm$0.0000 & 0.4631$\pm$0.0000 & 0.5113$\pm$0.0000 & 0.4511$\pm$0.0000 & 0.3728$\pm$0.0000 & 0.5710$\pm$0.0000 & 0.8020$\pm$0.0000 & 0.0000$\pm$0.0000 \\ 
			& & \boldmath$\operatorname{CoFKM}$ & - & - & - & - & - & - & - & - \\ 
			& & \boldmath$\operatorname{LMSC}$ & - & - & - & - & - & - & - & - \\ 
			& & \boldmath$\operatorname{LTMSC}$ & \textbf{0.5991$\pm$0.0003} & 0.5220$\pm$0.0003 & \textbf{0.6017$\pm$0.0003} & 0.5145$\pm$0.0003 & 0.5058$\pm$0.0003 & 0.5236$\pm$0.0004 & 0.8592$\pm$0.0001 & 0.0000$\pm$0.0000 \\ 
			& & \boldmath$\operatorname{MCSSC}$ & 0.2723$\pm$0.0000 & 0.1213$\pm$0.0000 & 0.2879$\pm$0.0000 & 0.2427$\pm$0.0000 & 0.2148$\pm$0.0000 & 0.2790$\pm$0.0000 & 0.7519$\pm$0.0000 & 0.0000$\pm$0.0000 \\ 
			& & \boldmath$\operatorname{TBGL-MVC}$ & - & - & - & - & - & - & - & - \\ 
			& & \boldmath$\operatorname{MVASM}$ & 0.5706$\pm$0.0485 & \textbf{0.5799$\pm$0.0561} & \textbf{0.6126$\pm$0.0333} & 0.5362$\pm$0.0423 & 0.4979$\pm$0.0551 & 0.5810$\pm$0.0201 & 0.8568$\pm$0.0220 & 0.0000$\pm$0.0000 \\ 
			& & \boldmath$\operatorname{MvLRECM}$ & 0.4512$\pm$0.0316 & \textbf{0.5773$\pm$0.0145} & 0.4805$\pm$0.0175 & \textbf{0.8149$\pm$0.0151} & \textbf{0.8846$\pm$0.0191} & 0.7975$\pm$0.0267 & \textbf{0.9495$\pm$0.0041} & \textbf{0.1420$\pm$0.0251} \\ 
			
			\bottomrule
	\end{tabular}}
\end{table*}

For the $i$th object, let $\mathcal{G}_i \in \Omega$ be the provided ground-truth label. 
$\mathcal{S}_i \in \Omega$ is its clustering solution. 
In the hard and fuzzy partitions, $\mathcal{S}_i\in \Omega$. In the credal partition, $\mathcal{S}_i\in 2^\Omega$
The clustering accuracy (ACC) is redefined in the credal partition as follows
\begin{equation}
ACC=\frac{\sum_{i=1}^{N}\phi(\mathcal{G}_i,map(\mathcal{S}_i))}{N},
\end{equation}
where 
\begin{equation}
\phi(\mathcal{G}_i,map(\mathcal{S}_j)) = 
\begin{cases}
1  & \text{if $\mathcal{G}_i \in \mathcal{S}_j$} \\ 
0  & \text{else},
\end{cases}
\end{equation}
where $map(\mathcal{S}_j)$ is the permutation mapping function that maps each label $\mathcal{G}_i$ to the equivalent label from data. The best mapping can be obtained by using the Hungarian Algorithm~\cite{kuhn1955hungarian}.

Precision, Recall, F-score, and Rand Index (RI) can be calculated as follows
\begin{equation}
Precision=\frac{TP}{TP+FP},\nonumber
\end{equation}
\begin{equation}
Recall=\frac{TP}{TP+FN},\nonumber
\end{equation}
\begin{equation}
F\mbox{-}sore=\frac{2 \times Precision \times Recall}{Precision + Recall},\nonumber
\end{equation}
\begin{equation}
RI=\frac{TP+TN}{TP+FP+FN+TN}.\nonumber
\end{equation}
When applying to credal partition, the extensions of definitions of TP, FP, FN, and TN are shown in Table~\ref{tab:STFPN}.

In addition, Imprecision Rate (IR) is used to describe the degree of imprecision of results
\begin{equation}
IR=\frac{N_{im}}{N},
\end{equation}
where $N_{im}$ is the number of objects assigned to meta-clusters. For hard and fuzzy partitions which cannot describe imprecision, their IRs are always equal to zero.

\begin{figure*}[ht]
	\centering
	\includegraphics[width=1\linewidth]{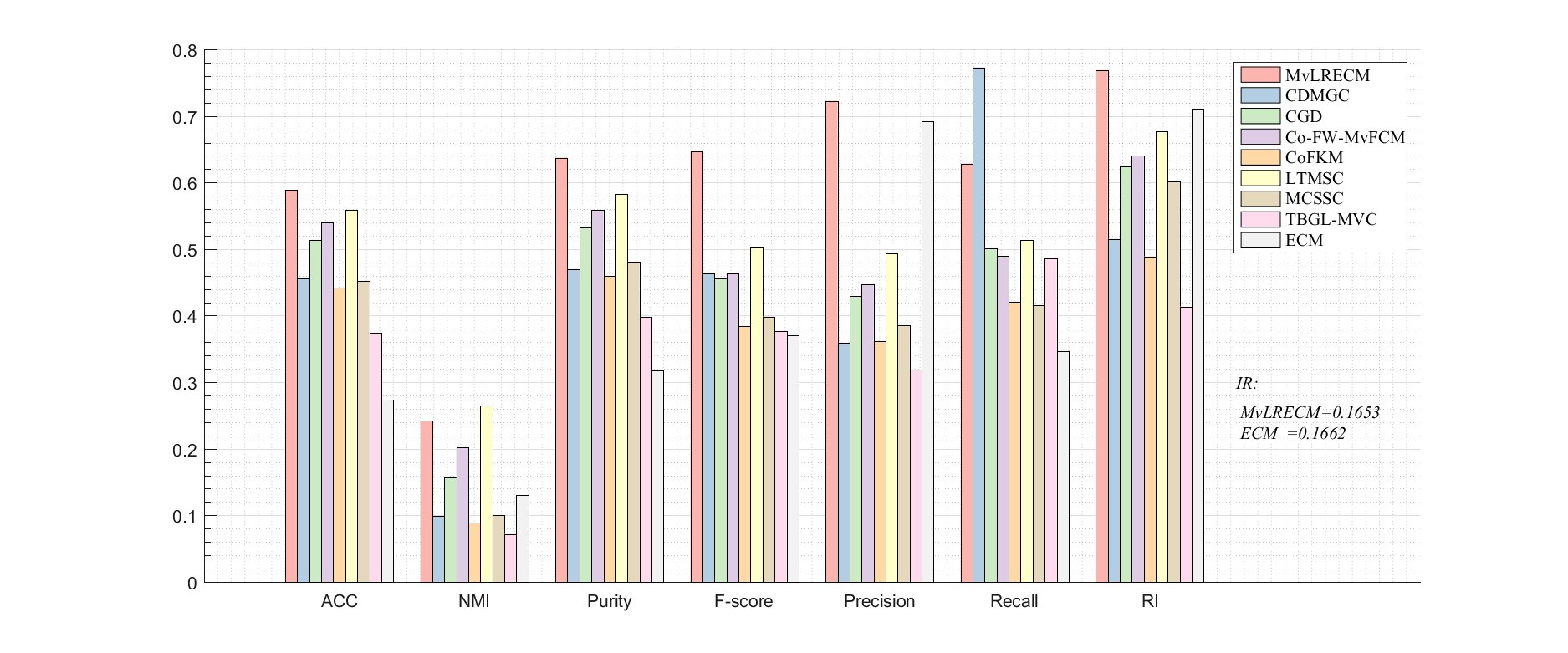}
	\caption{Average clustering performance on the real-world datasets.} 
	\label{fig:aver}
\end{figure*}
\begin{table*}[hb]\scriptsize
	\begin{center}
	\caption{Examples on the Mnist179 Dataset.}
		\begin{tabular}{cccccccccc}
			\toprule
			& \includegraphics{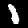} & \includegraphics{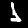} & \includegraphics{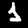} & \includegraphics{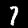} & \includegraphics{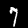} & \includegraphics{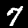} & \includegraphics{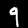} & \includegraphics{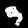} & \includegraphics{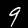} \\
			\midrule
			TrueLable & 1 & 1 & 1 & 7 & 7 & 7 & 9 & 9 & 9 \\
			CDMGC & 1 & 7 & 7 & 1 & 1 & 1 & 1 & 1 & 1 \\
			CGD & 9 & 9 & 9 & 7 & 7 & 7 & 7 & 7 & 7 \\
			CoFKM & 7 & 9 & 7 & 9 & 1 & 9 & 7 & 7 & 1 \\
			LMSC  & 1 & 7 & 1 & 1 & 7 & 1 & 9 & 9 & 1 \\
			LTMSC & 7 & 7 & 7 & 1 & 9 & 9 & 9 & 9 & 7 \\
			MCSSC & 9 & 9 & 9 & 9 & 9 & 9 & 9 & 9 & 1 \\
			TBGL-MVC & 1 & 9 & 1 & 1 & 1 & 1 & 1 & 1 & 1 \\
			MVASM & 7 & 7 & 1 & 7 & 9 & 9 & 9 & 7 & 1  \\
			MvLRECM & \{1,7\} & \{1,7\} & \{1,9\} & \{7,9\} & \{1,7\} & \{7,9\} & \{7,9\} & \{7,9\} & \{7,9\} \\	
			\bottomrule
		\end{tabular}

		\label{tab:num}   
	\end{center}
\end{table*}
\subsection{Running example on toy dataset}
\label{sec:toy}
In this subsection, we generate one toy dataset, named the 3DBall dataset, to validate the effectiveness of MvLRECM and reveal the limitations of hard and fuzzy partitions

The 3DBall dataset contains 1500 objects in 3 clusters. The objects of each cluster follow a combination of 3D Gaussian distribution and 3D spherical uniform distribution. We use their projection coordinates in the xy plane, yz plane, and xz plane as three views of this dataset. The original distribution of objects and clustering results are shown
in Fig.~\ref{fig:toy}. 
We give the results of MvLRECM when $\alpha=\{1,2,3\}$ in Figs.~\ref{fig:b1}-\ref{fig:c3} and Fig.~\ref{fig:a3} shows the mass distribution when $\alpha=2$.
Figs.~\ref{fig:a1}-\ref{fig:a2} show that three clusters $a_1,a_2,a_3$ have clear overlapping regions in which the objects are indistinguishable. From Figs.~\ref{fig:b1}-\ref{fig:c3}, we can see MvLRECM clearly characterizes imprecision of the cluster information. That is, it carefully assigns these indistinguishable objects to appropriate meta-clusters. 
In addition, the sizes of meta-clusters are determined by Parameter $\alpha$, which controls the number of objects in meta-clusters. Large $\alpha$ leads to large size of meta-clusters and a great IR. In the following experiments, we take $\alpha=2$, as suggested in~\cite{masson2008ecm}.
The performance of hard and fuzzy partitions on the 3DBall dataset is shown in Subsec.~\ref{sec:Stoy} in the supplementary.

\begin{figure*}[ht]
	\centering
	\subfloat[\label{fig:pa1}]{
		\includegraphics[width=0.25\linewidth]{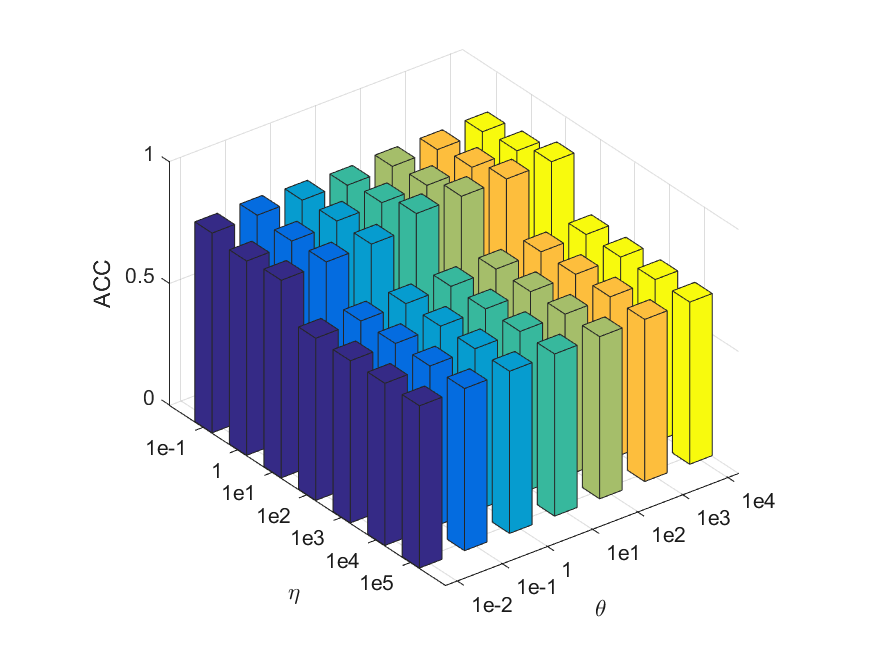}}
	\subfloat[\label{fig:pa2}]{
		\includegraphics[width=0.25\linewidth]{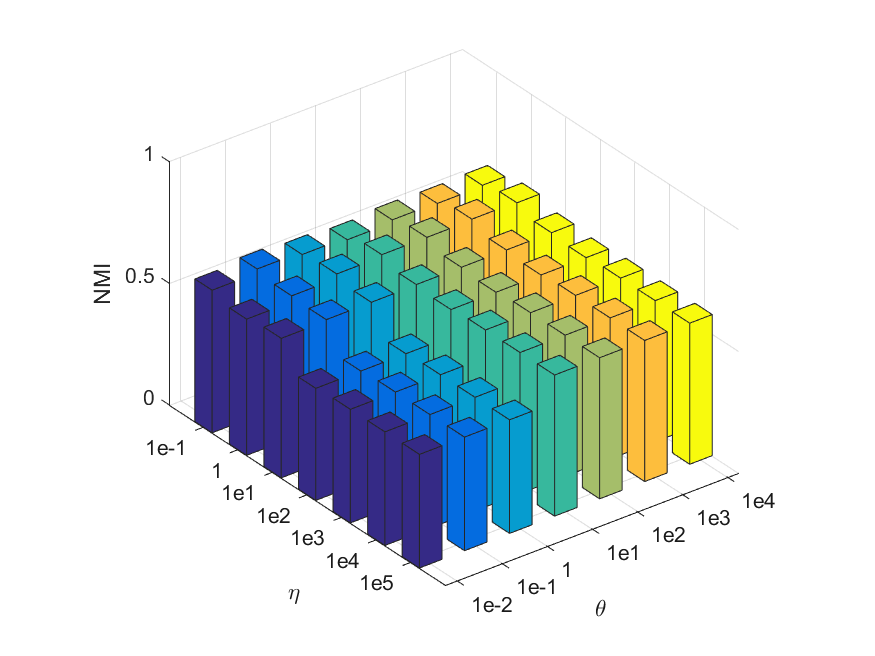}}
	\subfloat[\label{fig:pa3}]{
		\includegraphics[width=0.25\linewidth]{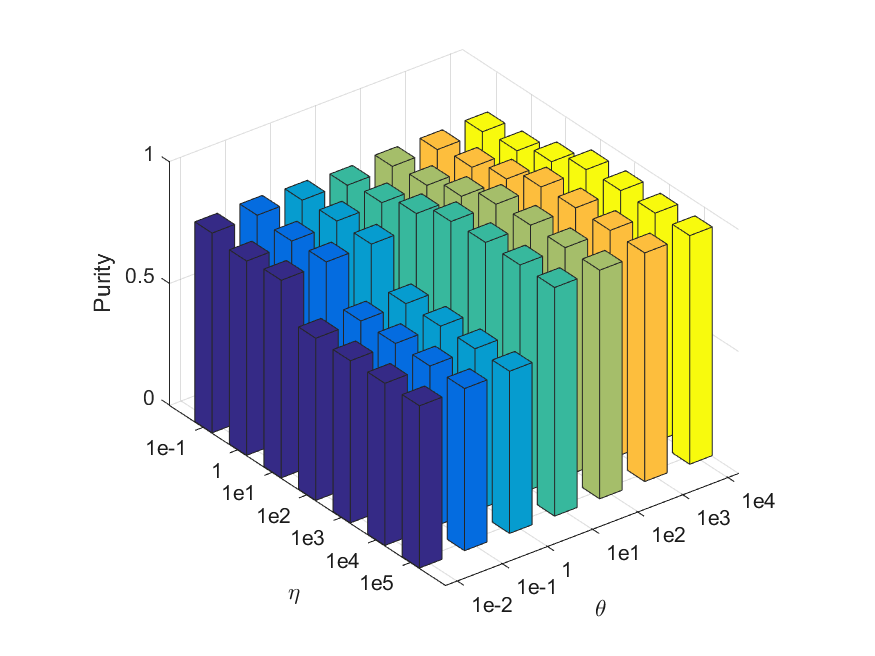}}
	\subfloat[\label{fig:pa4}]{
		\includegraphics[width=0.25\linewidth]{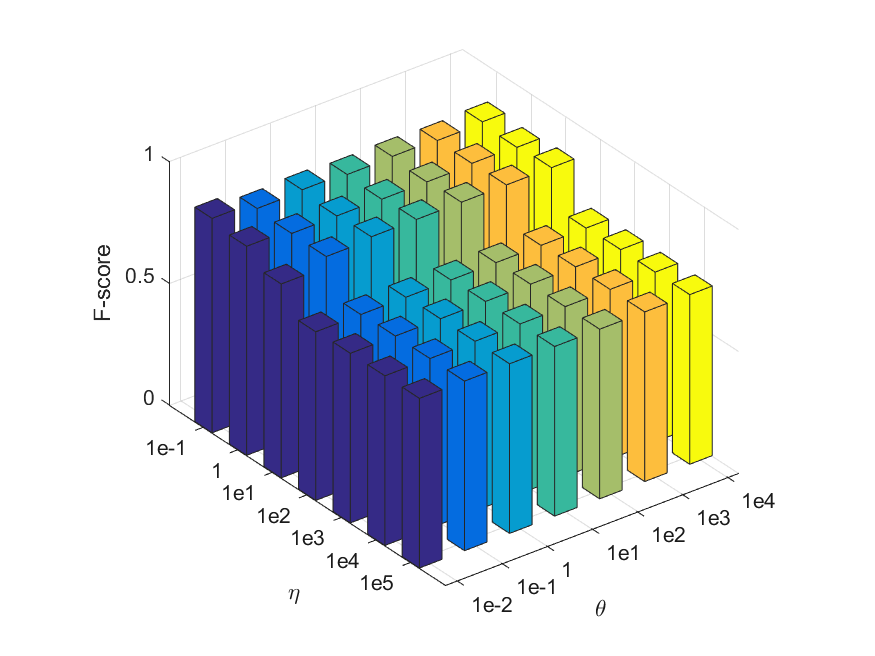}}
	\\
	\subfloat[\label{fig:pb1}]{
		\includegraphics[width=0.25\linewidth]{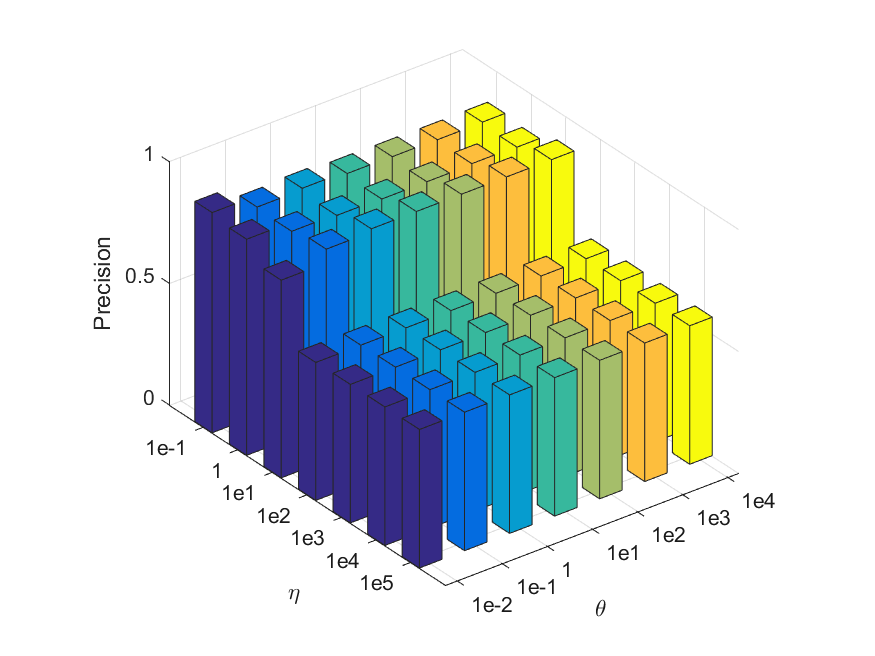}}
	\subfloat[\label{fig:pb2}]{
		\includegraphics[width=0.25\linewidth]{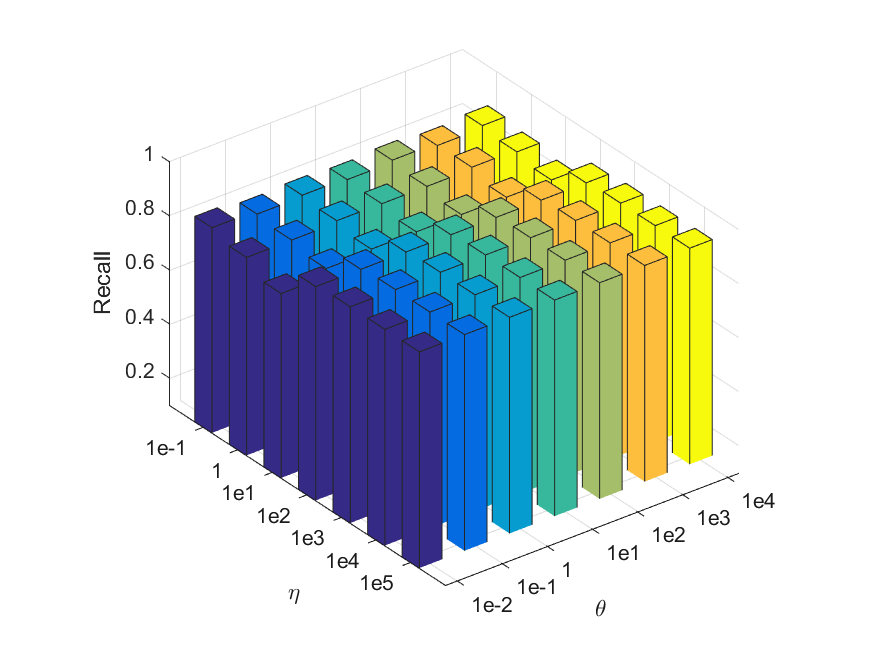}}
	\subfloat[\label{fig:pb3}]{
		\includegraphics[width=0.25\linewidth] {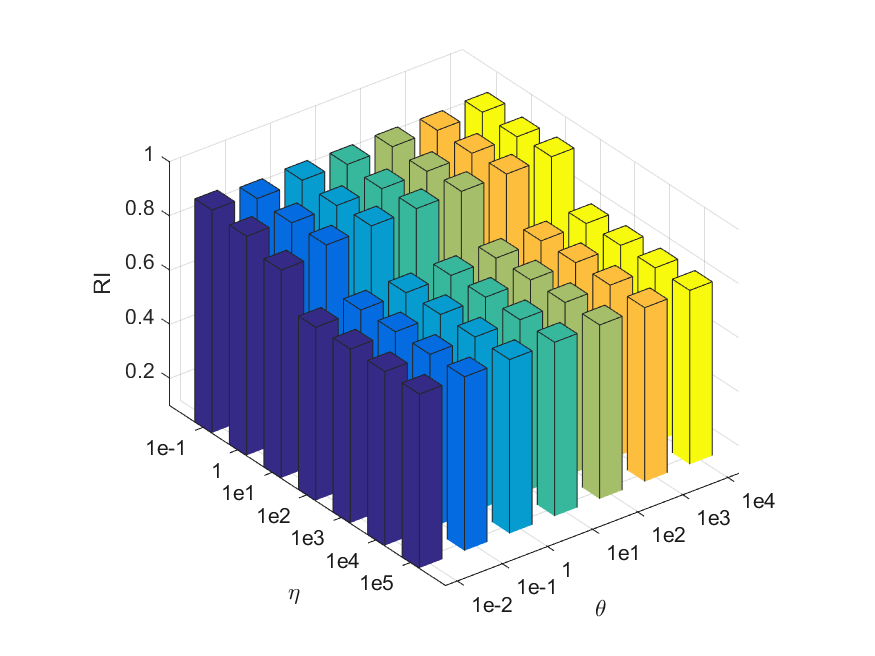}}
	\subfloat[\label{fig:pb4}]{
		\includegraphics[width=0.25\linewidth]{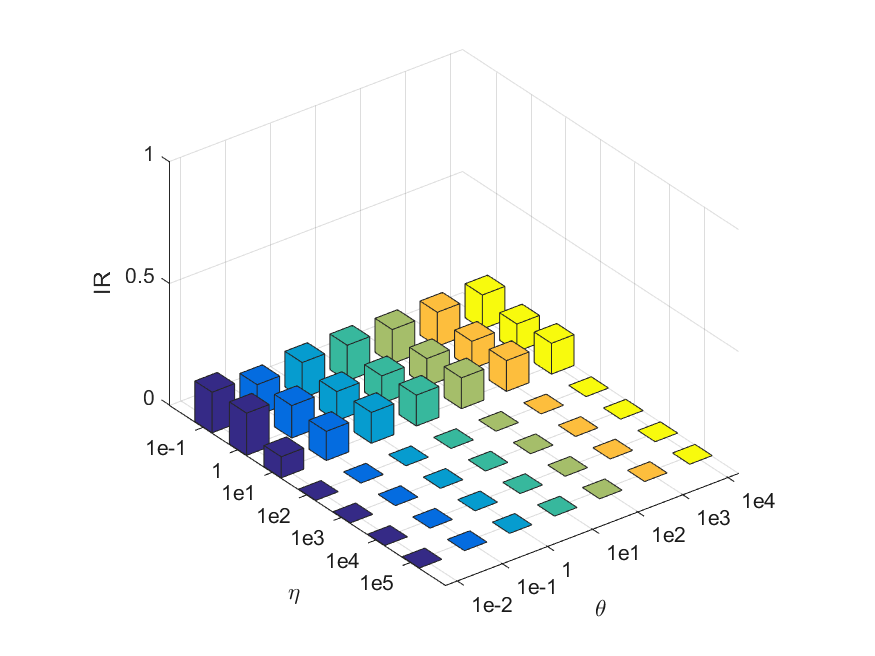}}
	\caption{How Parameter $\theta$ and $\eta$ affect the performance of MvLRECM on the Iris dataset, (a)ACC, (b)NMI, (c)Purity, (d)F-score, (e)Precision, (f)Recall, (g)RI, (h)IR, where $x$ and $y$ axis are $\theta$ and $\eta$.}
	\label{fig:para} 
\end{figure*}
\subsection{Experiments on real-world dataset}
\label{sec:real}

To demonstrate the performance of MvLRECM,
six popular real-world datasets are used, including Abalone, Contraceptive, Foresttype, Hayes, Ionosphere, and Segment, from UCI Machine Learning Repository\footnote{https://archive.ics.uci.edu/ml/datasets.}.
We compare MvLRECM with eight state-of-the-art clustering methods, including both single-view clustering and multi-view clustering methods: ECM~\cite{masson2008ecm},  CDMGC~\cite{huang2021measuring}, CGD~\cite{tang2020cgd}, Co-FW-MVFCM~\cite{yang2021collaborative}, LMSC~\cite{zhang2018generalized}, LTMSC~\cite{zhang2015low}, MCSSC~\cite{gao2021multi}, TBGL-MVC~\cite{xia2022tensorized}, and MVASM~\cite{han2020multi}. 
As a baseline, we apply the single-view evidential method, ECM.
The multi-view result of ECM is based on the average mass of belief of different views.

The details of these datasets and methods are shown in Subsec.~\ref{sec:Sinf} in the supplementary. For all methods, we repeat each experiment 30 times and report the average result and standard deviation.

The performance of these methods is depicted in Table~\ref{tab:RealW}, where the best performance is labeled with bold font\footnote{"-" indicates that the method is unable to obtain a valid clustering result on this dataset}. MvLRECM achieves an outstanding performance. 
To summarize the performance of the methods in all six datasets visually, we give Fig.~\ref{fig:aver} to depict the average performance, which shows that MvLRECM achieves best performance in five metrics (ACC, Purity, F-score, Precision, RI) and ranks second in two others (NMI, Recall). In addition, MvLRECM can achieve smaller IR compared to ECM while outperforming ECM in the other six metrics.

There are two reasons to explain why MvLRECM could outperform. First, hard and fuzzy partitions assign all the objects to singleton clusters. They cannot describe imprecision so the results have a high risk of errors when the cluster information of some objects is indistinguishable. MvLRECM overcomes this issue by applying credal partition and thus reduces the risk of errors. Second, compare to ECM, MvLRECM uses multi-view data in a more reasonable and efficient way. The combination of low-rank and Shannon entropy constraints allows MvLRECM better to exploit the consistency and diversity of multi-view data globally. Therefore, MvLRECM guarantees better performance while reducing IR compared to ECM.

However, MvLRECM cannot obtain the good results in all the datasets (\emph{i.e.} Ionosphere)
because the credal partition strategy is centroid-based, which is more suitable for spherical distributions and cannot get efficient performance in the datasets with arbitrary shapes and sizes in the space, such as non-spherical clusters~\cite{rodriguez2014clustering} or imbalanced clusters~\cite{jain2010data,liang2012k}.

Moreover, Mnist Handwrite Dataset~\cite{deng2012mnist}, a widespread benchmark for machine learning, is applied to further show the potential of MvLRECM. We select 3000 objects of the digits 1, 7, and 9, which are hard to identify, called Mnist179, and show partial results.
For these hard-to-identify objects, current methods force them into a singleton cluster, most of which are misclassifications. In contrast, MvLRECM does not assign them to a singleton cluster precisely but narrows down the range of singleton clusters to which objects may belong, and the correct singleton cluster is always included.

\subsection{Parameter study}
\label{sec:para}

MvLRECM introduces two regularization parameters $\theta$ and $\eta$. Parameter $\theta$ determines the degree of mass complementarity in the fusion strategy and parameter $\eta$ is for the auto-determination of view weights.  
We conduct experiments to analyze the effect of these parameters. We let $\theta=\{1e-2,1e-1,1,1e1,1e2,1e3,1e4\}$ and $\eta$ vary from $1e-1$ to $1e5$. Fig.~\ref{fig:para} shows the performance of eight metrics on the Dataset Iris
We observe that the values of metrics (NMI, Purity, Precision, F-score, Recall, and IR) fluctuate in the range of 0.2 when $\eta$ changes. Better performance is presented when $\eta=\{1e-1,1,1e1\}$.
In addition, MvLRECM demonstrates stable performance across a wide range of $\theta$. Only the performance of Purity and Precision shows a small increase as $\theta$ increases from $1$ to $1e1$.
This experiment demonstrates the robustness of MvLRECM to parameters. 	

The complete experimental results and more analysis of Subsecs.~\ref{sec:toy}-\ref{sec:para} are presented in Subsecs.~\ref{sec:Stoy}-\ref{sec:Spara} in the supplementary.

\section{Conclusion}
\label{sec:con}
In this paper, we propose a multi-view clustering method, MvLRECM, to characterize uncertainty and imprecision. The characterization of uncertainty and imprecision is achieved by applying credal partition based on the theory of belief functions. An entropy-weighting fusion strategy based on low-rank constraints is employed to integrate the multi-view masses to reduce the risk of errors and improve accuracy.
The experiments demonstrate that MvLRECM greatly improves the performance by assigning a few objects to meta-clusters. It significantly reduces the risk of errors, which, in some cases, is worse than imprecision. Meanwhile, MvLRECM shows some limitations. It cannot handle the datasets well when the clusters have arbitrary shapes and sizes. 
This is one of the problems for our future work. For the study of multi-view clustering, our theoretical analysis and experimental results validate the importance of characterizing uncertainty and imprecision. Many perspectives and generalizations can be expected in future works. 
The associated code for MvLRECM is available on GitHub at https://github.com/JinyiXUres/MvLRECM. 
	
	\bibliographystyle{IEEEtran}
	\bibliography{TETCI}

\end{document}


\title{Supplementary for the Manuscript:\\
		``How to characterize imprecision in multi-view clustering?"}
	
	\maketitle
	
	\section{NNM}
\label{sec:NNM}
Since direct rank minimization is NP-hard and is difficult to solve, the low-rank problem is generally relaxed by substitutively minimizing the nuclear norm of the estimated matrix, which is a convex relaxation of minimizing the matrix rank~\cite{fazel2002matrix}. 
\cite{cai2010singular} proves that the nuclear norm proximal (NNP) problem 
\begin{align}
min_{\boldsymbol{X}}\|\boldsymbol{Y} - \boldsymbol{X}\|_F^2 + \lambda\|\boldsymbol{X}\|_*
\label{eq:NNP}
\end{align}
can be solved in closed form by imposing a soft-thresholding operation on the singular values of the observation matrix
\begin{align}	\boldsymbol{X}=\boldsymbol{U}\mathcal{S}_{\frac{\lambda}{2}}\boldsymbol{(\varSigma)V}^T,&
\end{align}
where the nuclear norm of a matrix $\boldsymbol{X}$ is denoted by $\|\boldsymbol{X}\|_*$. 
$\boldsymbol{Y}=\boldsymbol{U\varSigma V}^T$ is the singular value decomposition (SVD) of $\boldsymbol{Y}$ and $\mathcal{S}_{\frac{\lambda}{2}}\boldsymbol{(\varSigma)}$ is the soft-thresholding function on diagonal matrix $\boldsymbol{\varSigma}$ with parameter $\frac{\lambda}{2}$. For each diagonal element $\boldsymbol{\varSigma_{\text{\em ii}}}$ in $\boldsymbol{\varSigma}$, there is
\begin{align}
\mathcal{S}_{\frac{\lambda}{2}}(\boldsymbol{\varSigma})_{\text{\em ii}} = \text{max}\boldsymbol{ (\varSigma}_{\text{\em ii}} - \frac{\lambda}{2}, 0\boldsymbol{ )}.
\end{align}

This methodology is called nuclear norm minimization (NNM).  NNM has been attracting significant attention due to its rapid development in both theory and implementation. 

\subsubsection{The proof of NNM}

A lemma is presented to analyze the NNP problem~\cite{mirsky1975trace}. 
\begin{lemma}
	\label{le:1}
	For any $m \times n$ matrices $\mathbf{A}$ and $\mathbf{B}$, $tr \left(\mathbf{A}^T\mathbf{B}\right) \le\sum_i\sigma_id_i$, where $\sigma_1 \ge \sigma_2 \ge\cdots\ge0$ and $d_1 \ge d_2 \ge\cdots\ge0$ are the descending singular values of $\mathbf{A}$ and $\mathbf{B}$, respectively. The case of equality occurs if and only if it is possible to find unitaries $\mathbf{U}$ and $\mathbf{V}$ that simultaneously singular value decompose $\mathbf{A}$ and $\mathbf{B}$ in the sense that
	\begin{equation}
	\mathbf{A}=\mathbf{U}\boldsymbol{\varSigma}\mathbf{V}^T, and \ \mathbf{B}=\mathbf{U}\boldsymbol{D}\mathbf{V}^T,\nonumber
	\end{equation}
	where 
	\begin{equation}
	\boldsymbol{\varSigma} = \left( \begin{array}{c} diag(\sigma_1,\sigma_2,\dots,\sigma_n) \\ \textbf{0} \end{array} \right)\in\mathbb{R}^{m\times n},
	\end{equation}
	\begin{equation}
	\boldsymbol{D} = \left( \begin{array}{c} diag(d_1, d_2,\dots,d_n) \\ \textbf{0} \end{array} \right).
	\end{equation}
\end{lemma}

Based on the result of Lemma~\ref{le:1}, we deduce the following important theorem.
\begin{theorem}
	Given a $m \times n$ matrix $\boldsymbol{A}$ without loss of generality, we assume that $m\ge n$.
	The optimized solution of Eq.~(\ref{eq:NNP}) can be expressed as $\hat{\mathbf{B}}=\mathbf{U}\hat{\boldsymbol{D}}\mathbf{V}^T$ and $(d_1, d_1, ..., d_n)$ is the solution to the following convex optimization problem
	\begin{align}
	\label{eq:sd}
	\min\limits_{d_1,d_2,\dots,d_n}&\sum^n_{i=1}\left(\sigma_i-d_i\right)^2+\lambda d_i,&\nonumber\\
	s.t. \quad&d_1\ge d_2\ge\dots\ge d_n\ge0.&
	\end{align}
	\label{the:1}
\end{theorem}

\begin{proof}[The proof of Theorem~\ref{the:1}.]
	For any $\boldsymbol{X}$, $\boldsymbol{Y}\in\mathbb{R}^{m\times n}\left(m > n\right)$, denote by $\boldsymbol{\bar{U}D\bar{V}}^T$ and $\boldsymbol{U\varSigma V^T}$ the singular value decomposition of matrix $\boldsymbol{X}$ and $\boldsymbol{Y}$, respectively. Based on the property of Frobenius norm, the following derivations hold		
	\begin{align}
	&\|\boldsymbol{Y} - \boldsymbol{X}\|^2_F + \lambda\|\boldsymbol{X}\|_*&\nonumber\\
	&=Tr\left(\boldsymbol{Y}^T\boldsymbol{Y}\right)- 2Tr\left(\boldsymbol{Y}^T\boldsymbol{X}\right) + Tr\left(\boldsymbol{X}^T\boldsymbol{X}\right) + \sum\limits_{i}^n \lambda d_i&\nonumber\\
	&=\sum\limits_{i}^n\sigma_i^2-2Tr\left(\boldsymbol{Y}^T\boldsymbol{X}\right) + \sum\limits_{i}^nd_i^2+\sum\limits_{i}^n \lambda d_i.&
	\end{align}
	Based on the von Neumann trace inequality in Lemma~\ref{le:1}, we know that $Tr\left(\boldsymbol{X}^T\boldsymbol{X}\right)$ achieves its upper bound $\sum_i^n\sigma_id_i$ if $\boldsymbol{\varSigma} = \bar{\boldsymbol{\varSigma}}$ and $\boldsymbol{D} = \bar{\boldsymbol{D}}$ and we have
	\begin{align}
	&\min\limits_{\boldsymbol{X}}\|\boldsymbol{Y}-\boldsymbol{X}\|^2_F+\lambda\|\boldsymbol{X}\|_*& \nonumber\\
	&\Leftrightarrow\min\limits_{\boldsymbol{D}}\sum\limits_{i}^n\sigma_i^2-2\sum\limits_{i}^n\sigma_id_i+\sum\limits_{i}^nd_i^2+\sum\limits_{i}^n\lambda d_i&\\
	&s.t.\ d_1\ge d_2\ge...\ge d_n\ge0&\\
	&\Leftrightarrow\min\limits_{\boldsymbol{D}}\sum\limits_{i}\left(d_i-\sigma_i\right)^2+\lambda d_i&\\
	&s.t.\ d_1\ge d_2\ge...\ge d_n\ge0,&
	\end{align}
\end{proof}

Eq.~(\ref{eq:NNP}) degenerates to the following unconstrained equation
\begin{align}
&\min\limits_{d_i\ge0}\left(d_i-\sigma_i\right)^2 + \lambda d_i\Leftrightarrow\min\limits_{d_i\ge0}\left(d_i-\left(\sigma_i-\frac{\lambda}{2}\right)\right)^2,&
\end{align}
thus we obtain
\begin{align}
&\bar{d}_i=\max\left(\sigma_i-\frac{\lambda}{2}, 0\right),\quad i=1,2,...,n.&
\label{eq:sd2}
\end{align}
Since we have $\sigma_1\ge\sigma_2\ge...\ge\sigma_n$, it is easy to see that $\bar{d}_1\ge\bar{d}_2\ge...\ge\bar{d}_n$. Thus, $\bar{d}_{i=1,2,...,n}$ satisfy the constraint of Eq.~(\ref{eq:sd}), and the solution in Eq.~(\ref{eq:sd2}) is the globally optimized solution of Eq.~(\ref{eq:sd}).

\section{EXPERIMENTS}
\label{sec:Sexp}
\subsection{Datasets and methods}
\label{sec:Sinf}
The six real-world datasets used in the main manuscript are Abalone, Contraceptive, Foresttype, Hayes, Ionosphere, and Segment.
The details of these datasets are shown in Table~\ref{tab:Sdataset}.
\begin{table}[h]
	\center
	\caption{Details of the real-world datasets.}
	\label{tab:Sdataset}
	\begin{tabular}{lllllllll}  
		\toprule
		Dataset & $N$ & $Q$ & $C$ & $D_1$ & $D_2$ & $D_3$ & $D_4$ & $D_5$\\
		\midrule
		Abalone & 4,174 & 3 & 3 & 3 & 2 & 3 & - & -\\
		Contraceptive & 1,473 & 2 & 3 & 7 & 2 & - & - & -\\
		Foresttype & 523 & 2 & 4 & 13 & 10 & 4 & - & -\\
		Hayes & 132 & 2 & 3 & 3 & 2 & - & - & -\\
		Ionosphere & 351 & 4 & 2 & 14 & 7 & 9 & 4 & -\\
		Segment & 2310 & 5 & 7 & 5 & 5 & 4 & 2 & 3\\
		\bottomrule
	\end{tabular}
\end{table}

Eight state-of-the-art clustering methods, including both single-view clustering and multi-view clustering methods are used to compare the performance: ECM~\cite{masson2008ecm},  CDMGC~\cite{huang2021measuring}, CGD~\cite{tang2020cgd}, Co-FW-MVFCM~\cite{yang2021collaborative}, LMSC~\cite{zhang2018generalized}, LTMSC~\cite{zhang2015low}, MCSSC~\cite{gao2021multi}, TBGL-MVC~\cite{xia2022tensorized}, and MVASM~\cite{han2020multi}. 
\begin{enumerate}
	\item ECM: It is a single-view clustering method, called evidential \emph{c}-means (ECM)~\cite{masson2008ecm}, in the theoretical framework of belief functions. It can characterize uncertainty and imprecision. ECM is based on a classical alternating minimization method. The first step of ECM is the determination of the centers of the clusters, and the second step is the allocation of the masses to the different clusters;
	
	\item Consistent and Divergent Multi-view Graph Clustering (CDMGC): This method can formulate both the multi-view consistency and diversity in a unified framework. It automatically assigns suitable weights for different views based on their clustering capacity;
	
	\item Cross-view Graph Diffusion (CGD): It employs a diffusion process for multi-view clustering. It takes the traditional predefined graph matrices of different views as input, and learns an improved graph for every view via an iterative cross-diffusion process;	
	
	\item Collaborative Feature-weighted Multi-view Fuzzy C-means(Co-FW-MVFCM): This method contains two steps, including a local step and a collaborative step. The local step is a single-view partition procedure to produce a local partition in each view, and the collaborative step is sharing information about their memberships between different views;
	
	\item CoFKM: It's a centralized strategy based on the fuzzy \emph{k}-means method for multi-view scenarios;
	
	\item Low-rank Tensor constrained Multiview Subspace Clustering (LTMSC): This method regards the subspace representation matrices of different views as a tensor, which captures dexterously the high order correlations underlying multi-view data;
	
	\item Low-rank Tensor constrained Multiview Subspace Clustering (LMSC): It is a subspace clustering model for multi-view data. This method explores underlying complementary information from multiple views and simultaneously seeks the underlying latent representation. In this paper, we use linear
	LMSC (lLMSC), based on linear correlations between latent representation in the experiment;
	
	\item Multi-View Clustering With Self-Representation and Structural Constraint (MCSSC): This method is a network-based method. It fuses matrix factorization and low-rank representation of different views to remove heterogeneity of multi-view;
	
	\item Tensorized Bipartite Graph Learning for
	Multi-view Clustering (TBGL-MVC): this method is based on a variance-based de-correlation anchor selection strategy. It exploits the similarity of inter-view by minimizing the tensor Schatten \emph{p}-norm, which well exploits both the spatial structure and complementary information embedded in the bipartite graphs of views.
	
	
	\item Multi-View clustering with Adaptive Sparse Memberships and Weight Allocation (MVASM): A common and flexible sparse membership matrix is learned to indicate the clustering in this method. It pays more attention to constructing a common membership matrix with proper sparseness over different views and learns the centroid matrix and its corresponding weight of each view.
\end{enumerate}

All the parameters of these methods are set as suggested in their papers, shown in Table~\ref{tab:Sparaset}.
\begin{table}[h]
	\center
	\caption{Information of the methods.}
	\label{tab:Sparaset} 
	\begin{tabular}{lll}
		\toprule
		Method & Partition& Parameter Set \\
		\midrule
		ECM & credal &$\alpha = 2, \beta = 2, \delta = 20$ \\
		\multirow{2}*{CDMGC} & \multirow{2}*{hard} & $\alpha = 1e5, \beta = 1e-5,$\\ &&$knn = 9$ \\
		CGD & hard&$\sigma = 0.5$ \\
		Co-FW-MVFCM &fuzzy& $m = 2, \beta = 4$  \\
		CoFKM &fuzzy& $\beta = 1.25$  \\
		LMSC & hard& $\lambda = 0.1$  \\
		LTMSC & hard& $\lambda = 0.1, k = 100$  \\
		MCSSC & fuzzy& $\lambda = 10, \gamma = 1$  \\
		TBGL-MVC & hard& $p = 0.9, \text{anchor rate} = 0.5$  \\
		MVASM & hard & $\gamma=0.5, q=2$ \\
		MvLRECM & credal& $\alpha=2, \theta=1e1, \eta=1e1$ \\
		\bottomrule
	\end{tabular}
\end{table}

\subsection{Running Example on toy dataset}
\label{sec:Stoy}
\begin{figure*} [t!]
	\centering
	\subfloat[\label{fig:Sa1}]{
		\includegraphics[width=3.5cm]{figure/toy/3DO2.png}}
	\subfloat[\label{fig:Sa2}]{
		\includegraphics[width=0.25\linewidth]{figure/toy/xyO1.png}}
	\subfloat[\label{fig:Sa3}]{
		\includegraphics[width=0.25\linewidth]{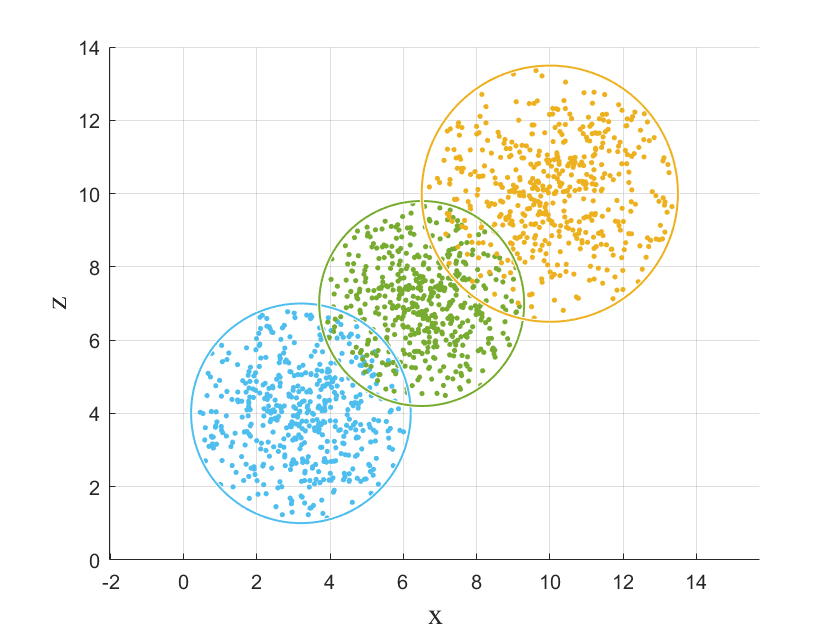}}
	\subfloat[\label{fig:Sa4}]{
		\includegraphics[width=0.25\linewidth]{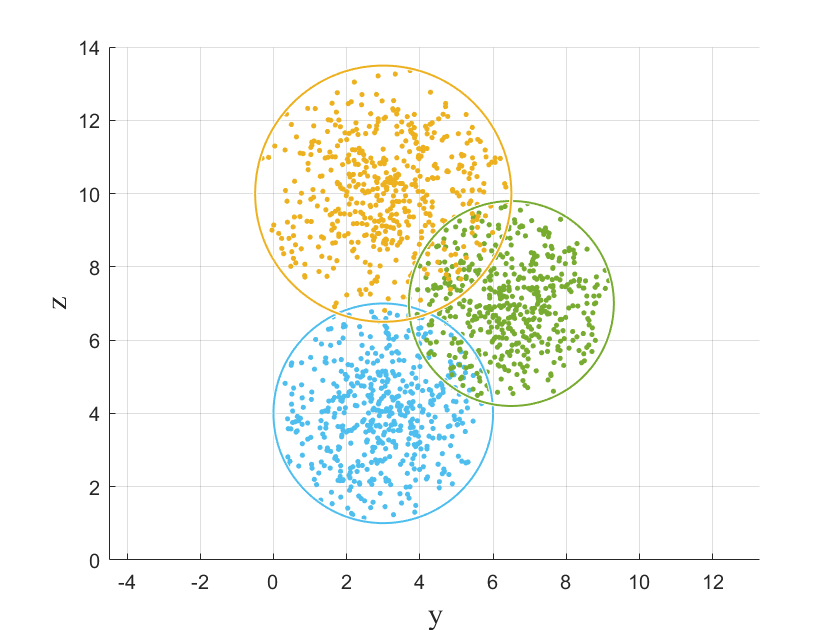}}
	\\
	\subfloat[\label{fig:Sb1}]{
		\includegraphics[width=0.25\linewidth]{figure/toy/3DM1.png}}
	\subfloat[\label{fig:Sb2}]{
		\includegraphics[width=0.25\linewidth]{figure/toy/xyM1.png}}
	\subfloat[\label{fig:Sb3}]{
		\includegraphics[width=0.25\linewidth]{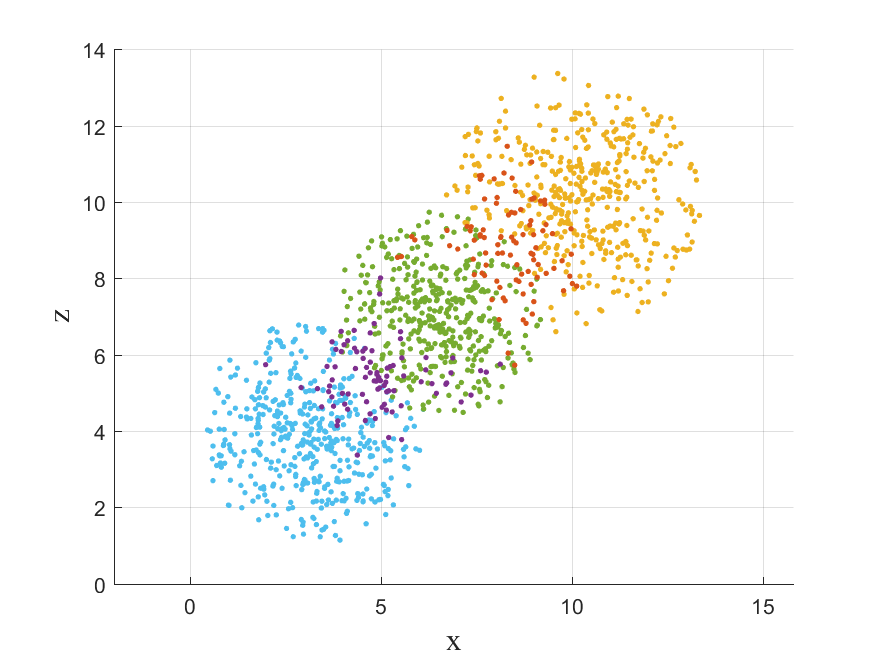}}
	\subfloat[\label{fig:Sb4}]{
		\includegraphics[width=0.25\linewidth]{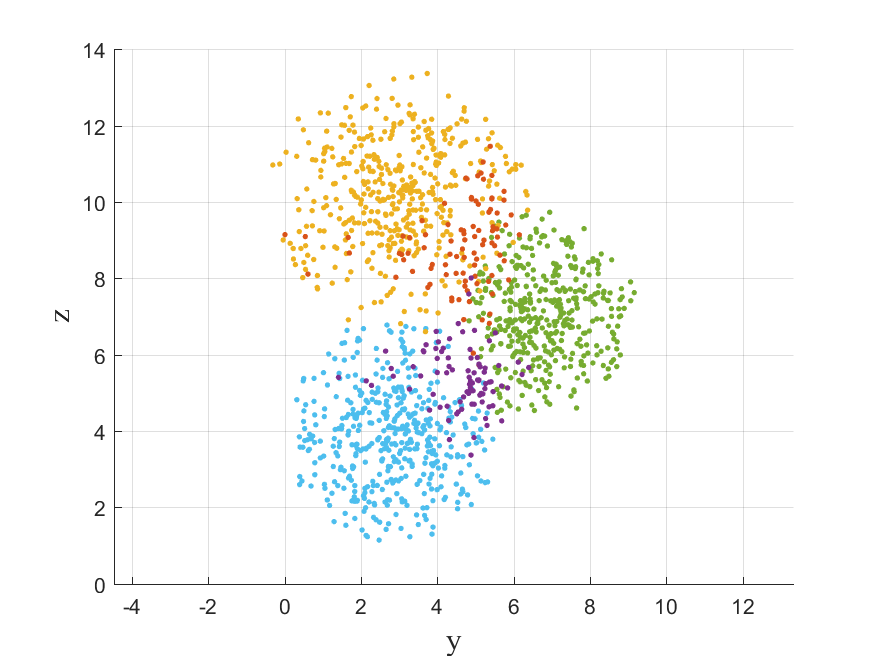}}
	\\
	\subfloat[\label{fig:Sc1}]{
		\includegraphics[width=0.25\linewidth]{figure/toy/3DM2.png}}
	\subfloat[\label{fig:Sc2}]{
		\includegraphics[width=0.25\linewidth]{figure/toy/xyM2.png}}
	\subfloat[\label{fig:Sc3}]{
		\includegraphics[width=0.25\linewidth]{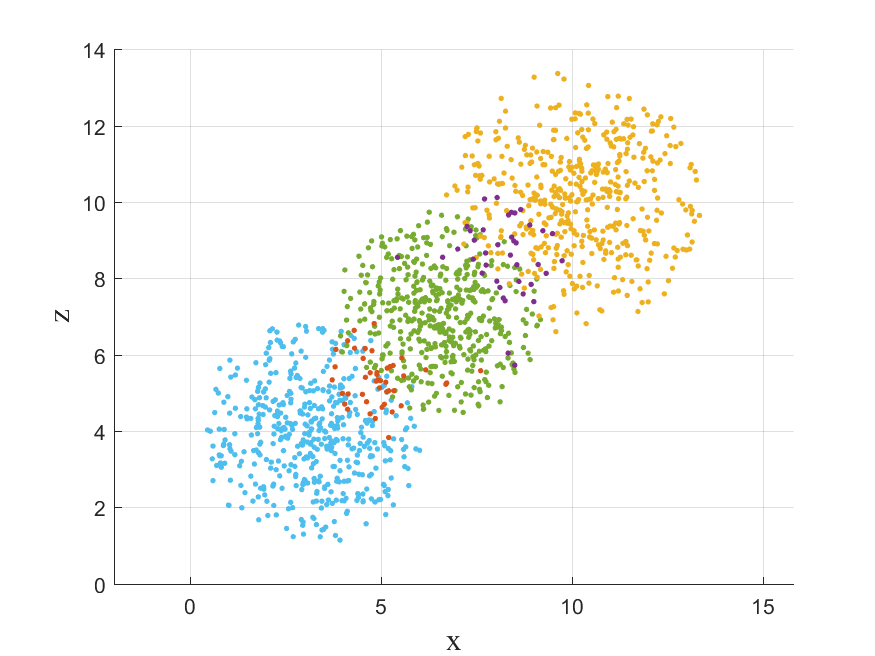}}
	\subfloat[\label{fig:Sc4}]{
		\includegraphics[width=0.25\linewidth]{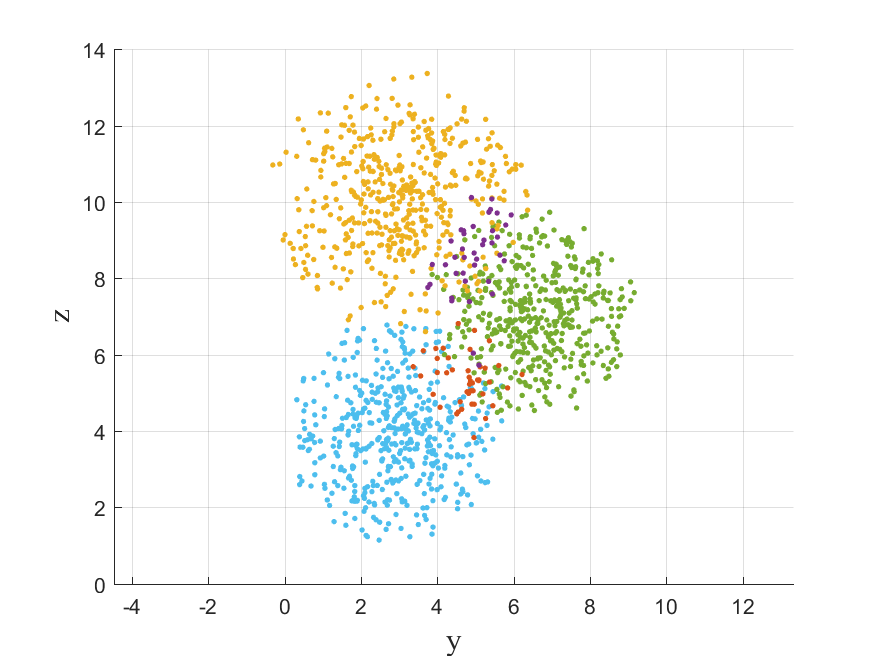}}
	\\
	\subfloat[\label{fig:Sd1}]{
		\includegraphics[width=0.25\linewidth]{figure/toy/3DM3.png}}
	\subfloat[\label{fig:Sd2}]{
		\includegraphics[width=0.25\linewidth]{figure/toy/xyM3.png}}
	\subfloat[\label{fig:Sd3}]{
		\includegraphics[width=0.25\linewidth]{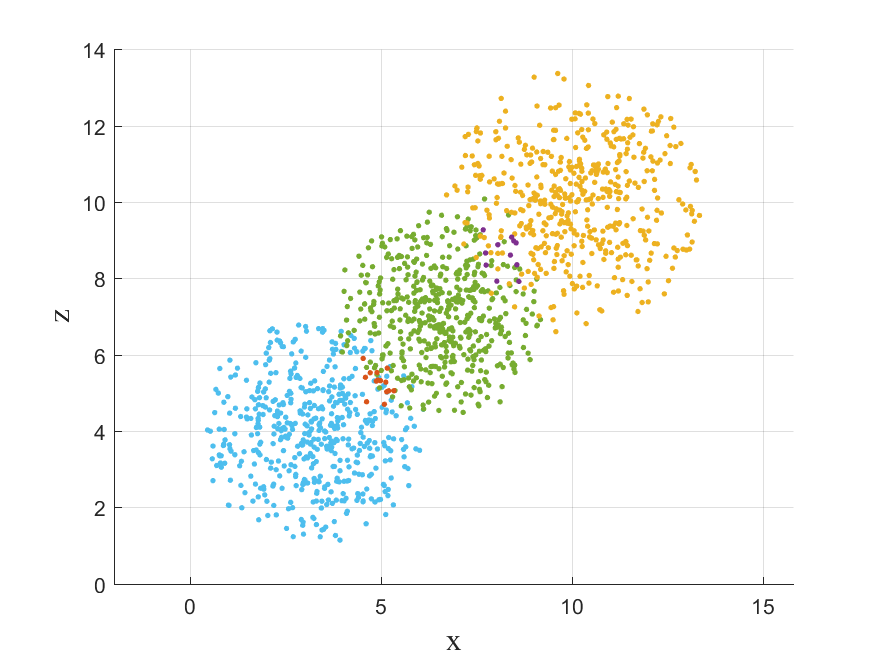}}
	\subfloat[\label{fig:Sd4}]{
		\includegraphics[width=0.25\linewidth]{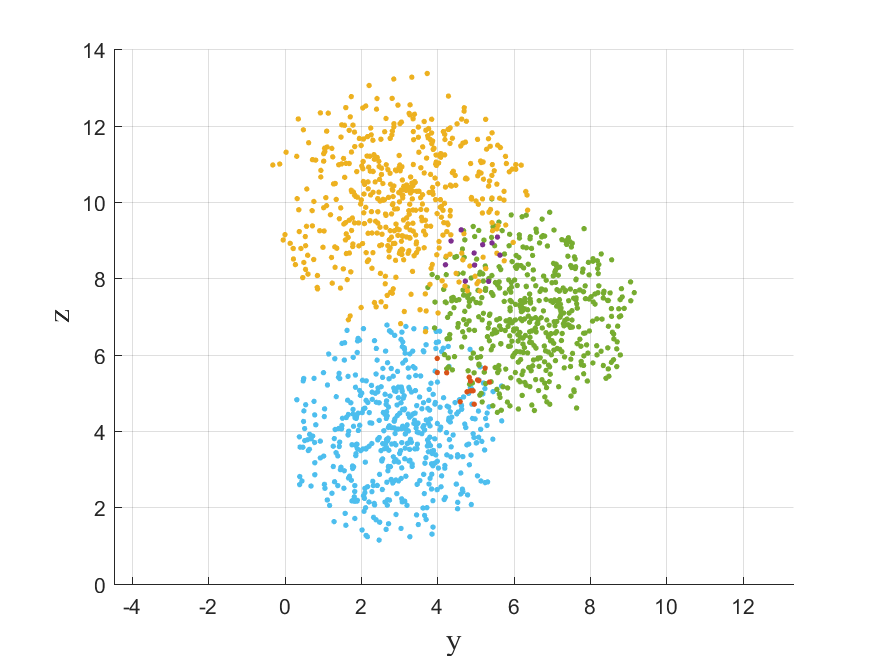}}
	
	\caption{Results on the 3DBall dataset, (a) the original 3D distribution, (b)-(c) are the three views (xy plane, yz plane, and xz plane) of the original dataset, respectively,
		(e)-(h) MvLREVM ($\alpha=1$), 
		(i)-(l) MvLREVM ($\alpha=2$), 
		(m)-(p) MvLREVM ($\alpha=3$).}
	\label{fig:toyS} 
\end{figure*}

\begin{figure*} [t!]
	\centering
	\subfloat[\label{fig:Ha1}]{
		\includegraphics[width=0.25\linewidth]{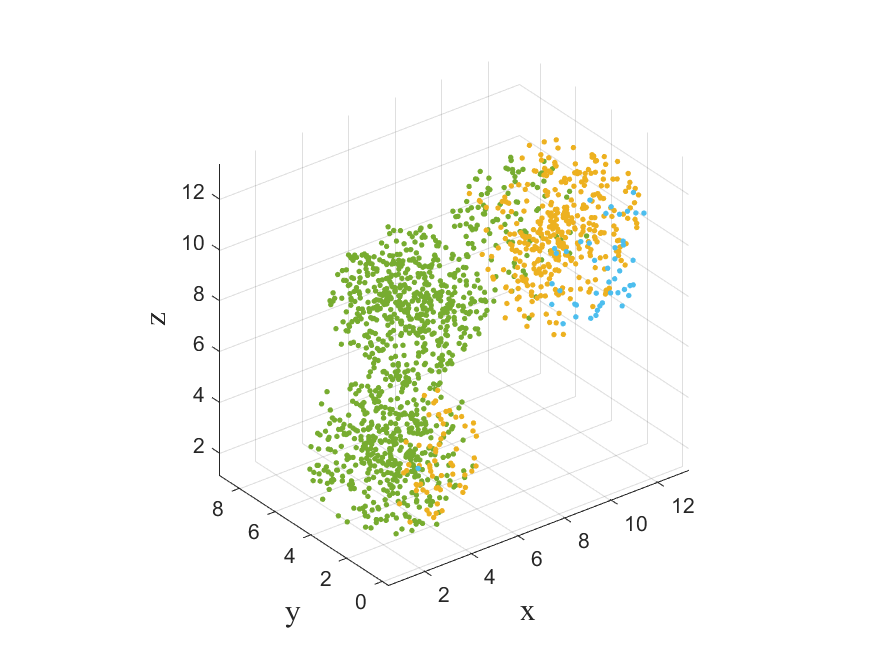}}
	\subfloat[\label{fig:Ha2}]{
		\includegraphics[width=0.25\linewidth]{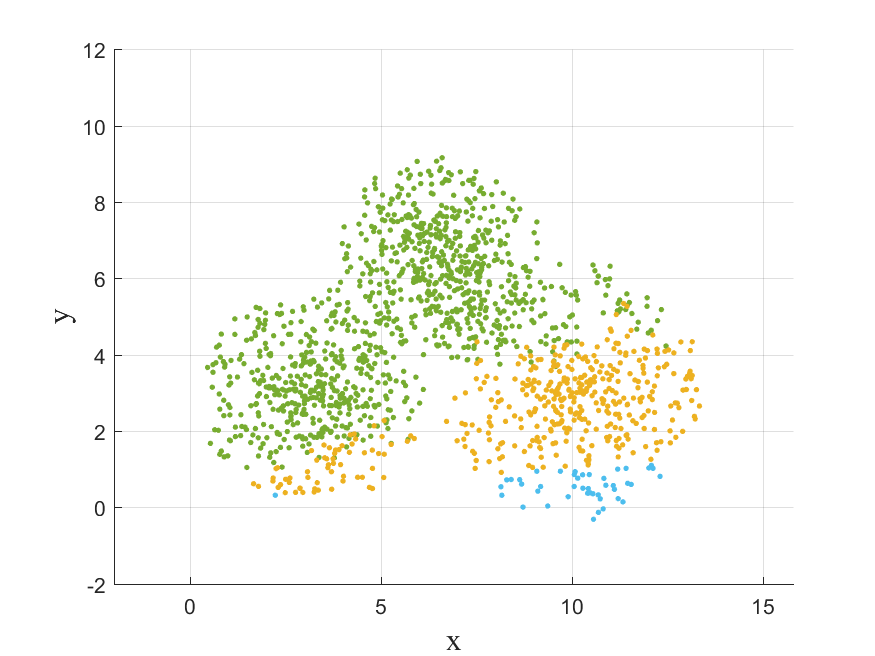}}
	\subfloat[\label{fig:Ha3}]{
		\includegraphics[width=0.25\linewidth]{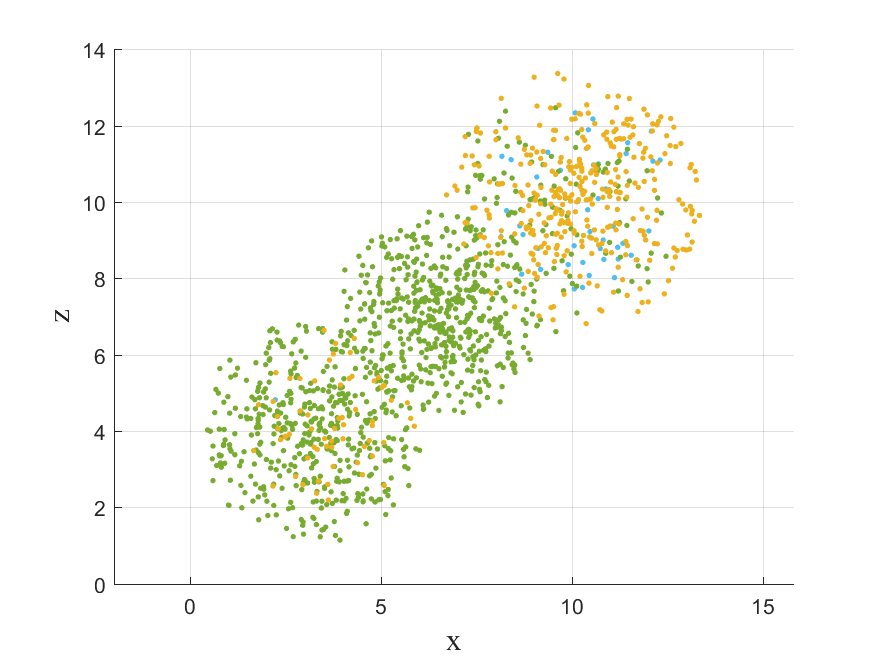}}
	\subfloat[\label{fig:Ha4}]{
		\includegraphics[width=0.25\linewidth]{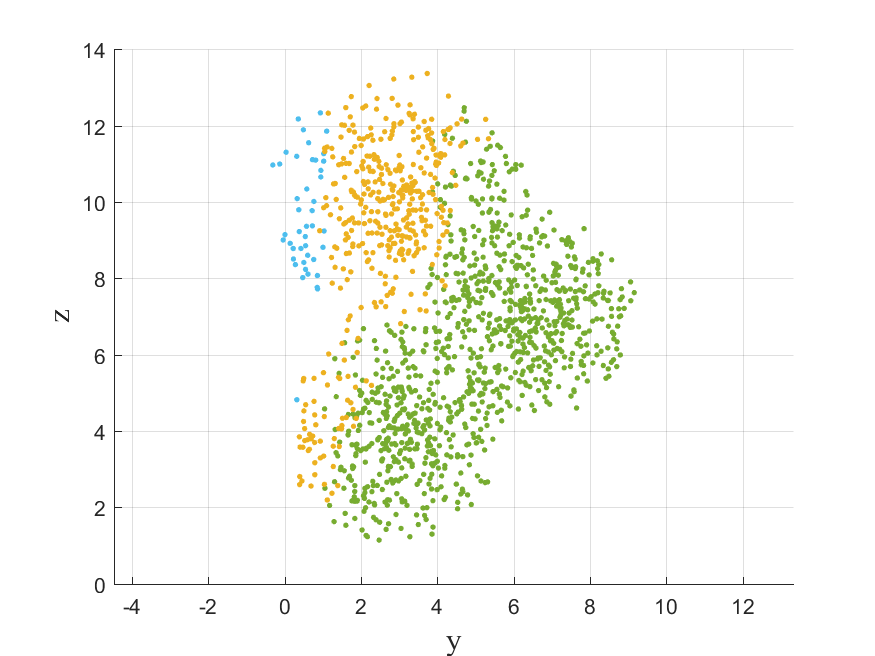}}
	\\
	\subfloat[\label{fig:Hb1}]{
		\includegraphics[width=0.25\linewidth]{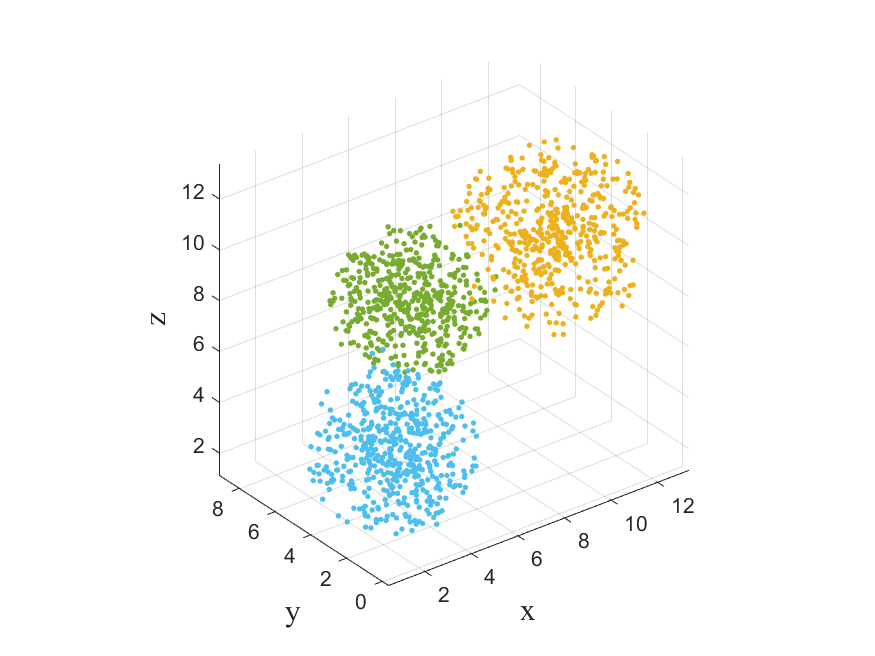}}
	\subfloat[\label{fig:Hb2}]{
		\includegraphics[width=0.25\linewidth]{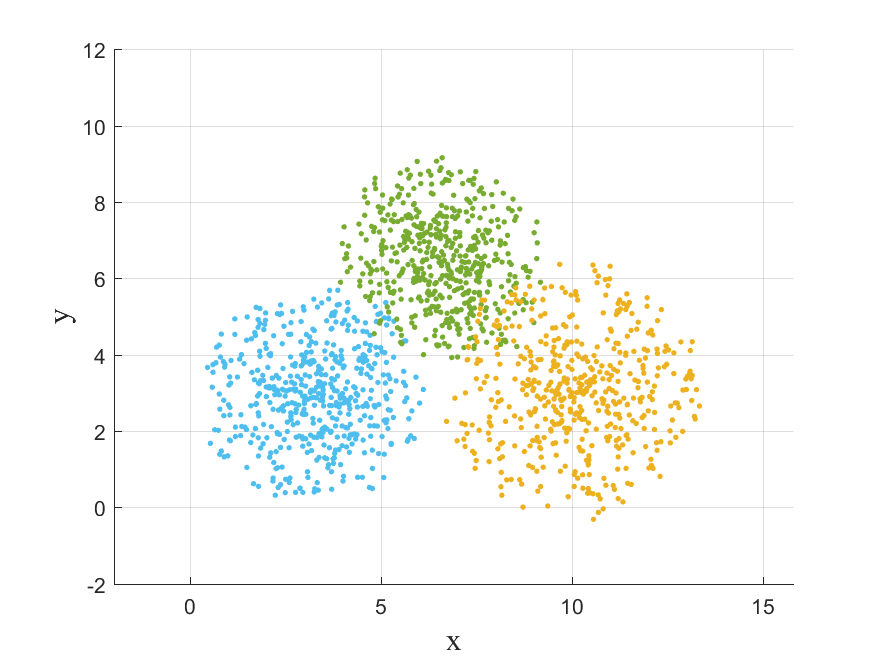}}
	\subfloat[\label{fig:Hb3}]{
		\includegraphics[width=0.25\linewidth]{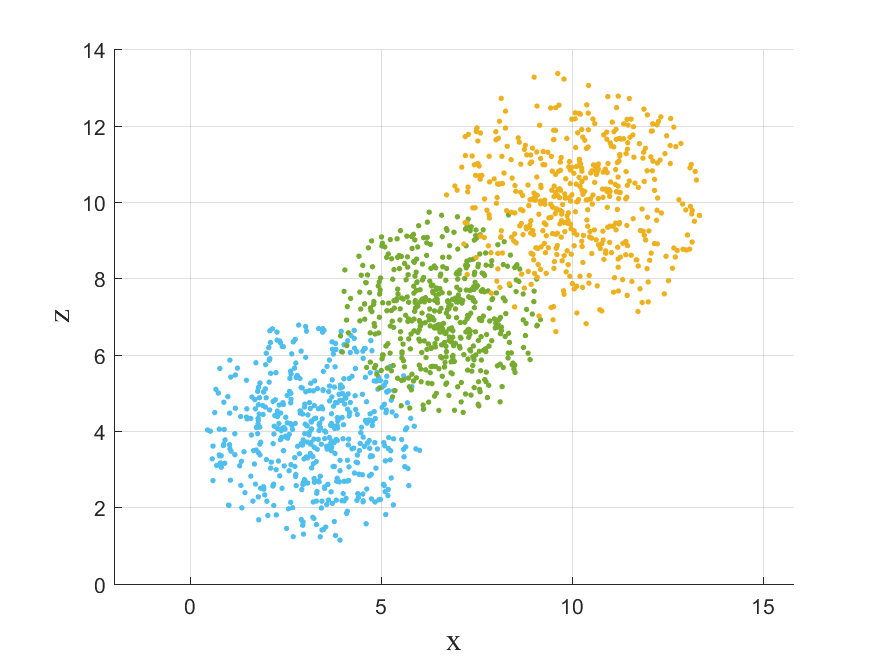}}
	\subfloat[\label{fig:Hb4}]{
		\includegraphics[width=0.25\linewidth]{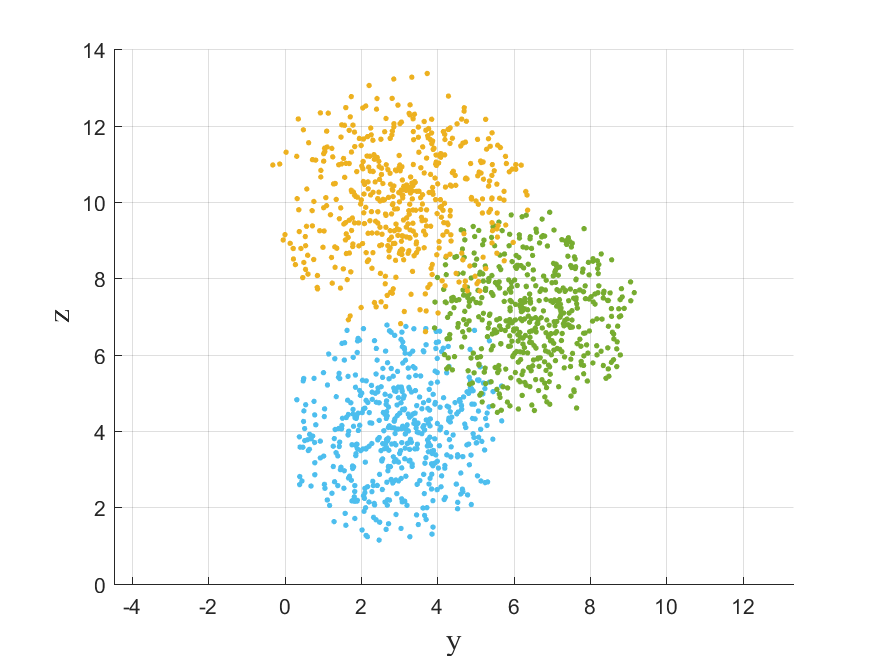}}
	\\
	\subfloat[\label{fig:Hc1}]{
		\includegraphics[width=0.25\linewidth]{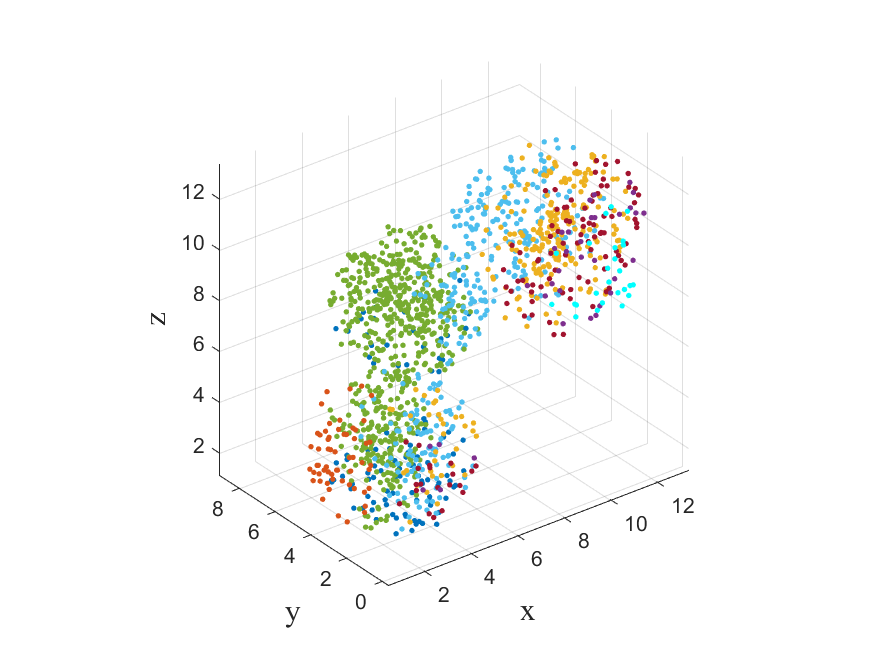}}
	\subfloat[\label{fig:Hc2}]{
		\includegraphics[width=0.25\linewidth]{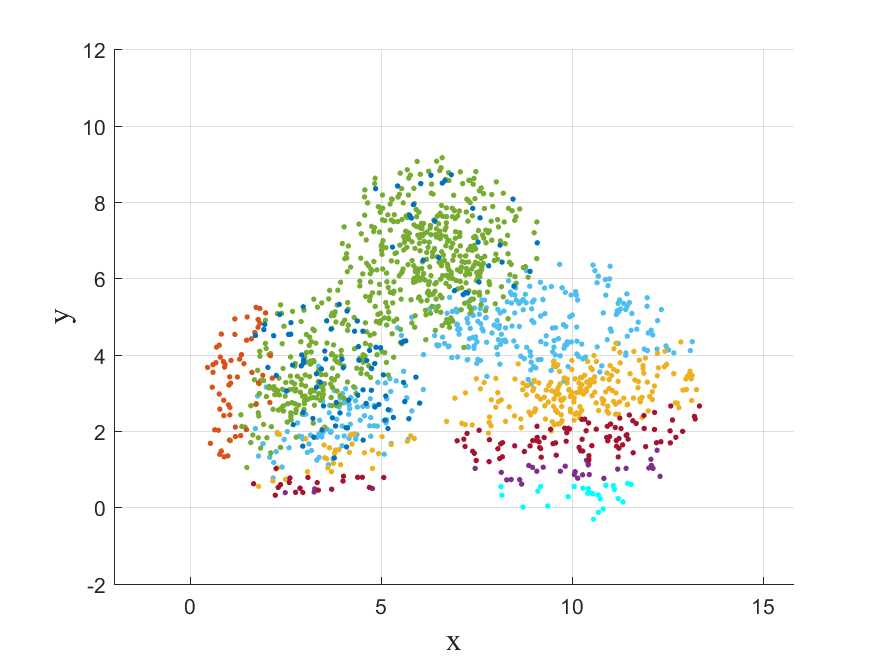}}
	\subfloat[\label{fig:Hc3}]{
		\includegraphics[width=0.25\linewidth]{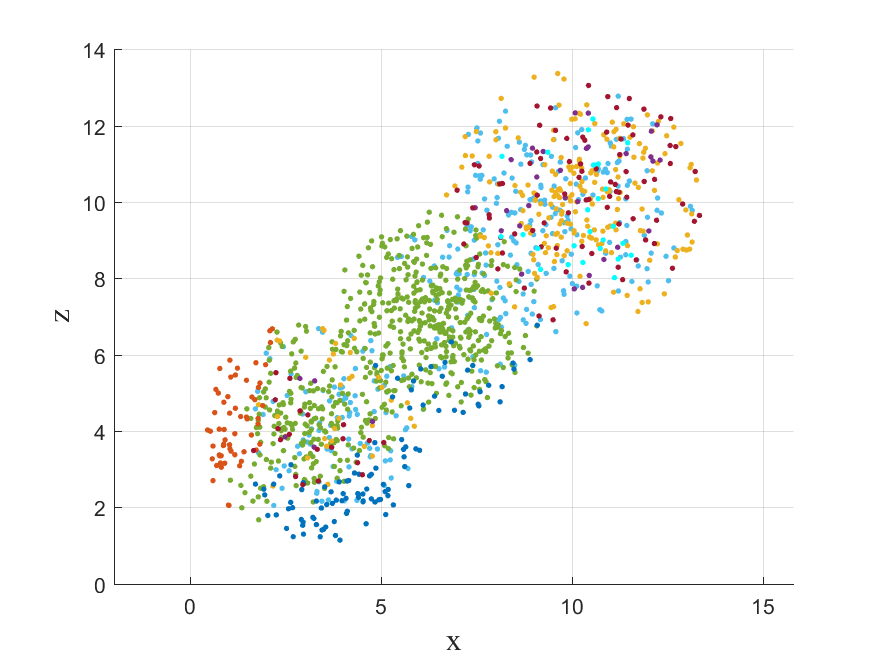}}
	\subfloat[\label{fig:Hc4}]{
		\includegraphics[width=0.25\linewidth]{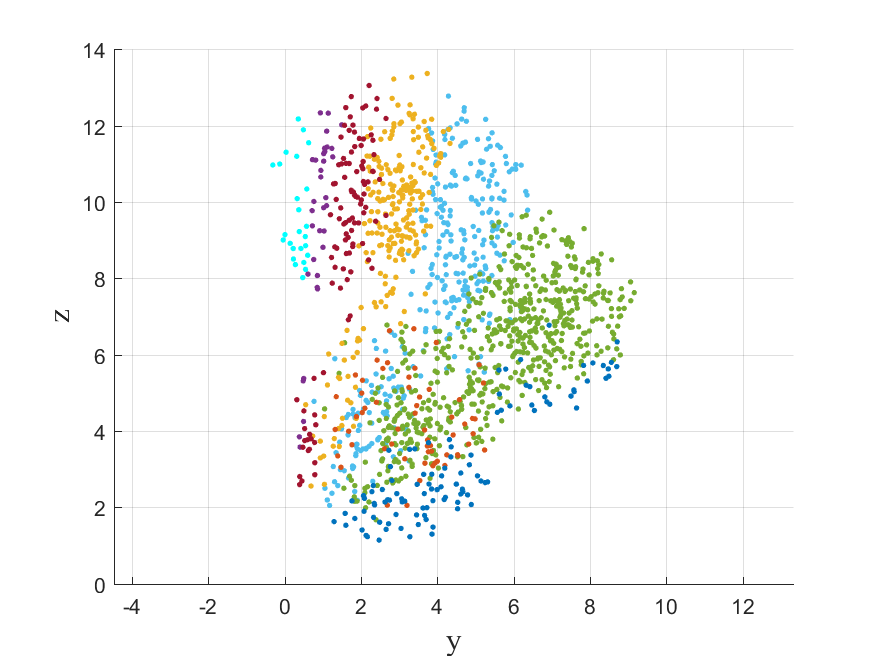}}
	\\
	\subfloat[\label{fig:Hd1}]{
		\includegraphics[width=0.25\linewidth]{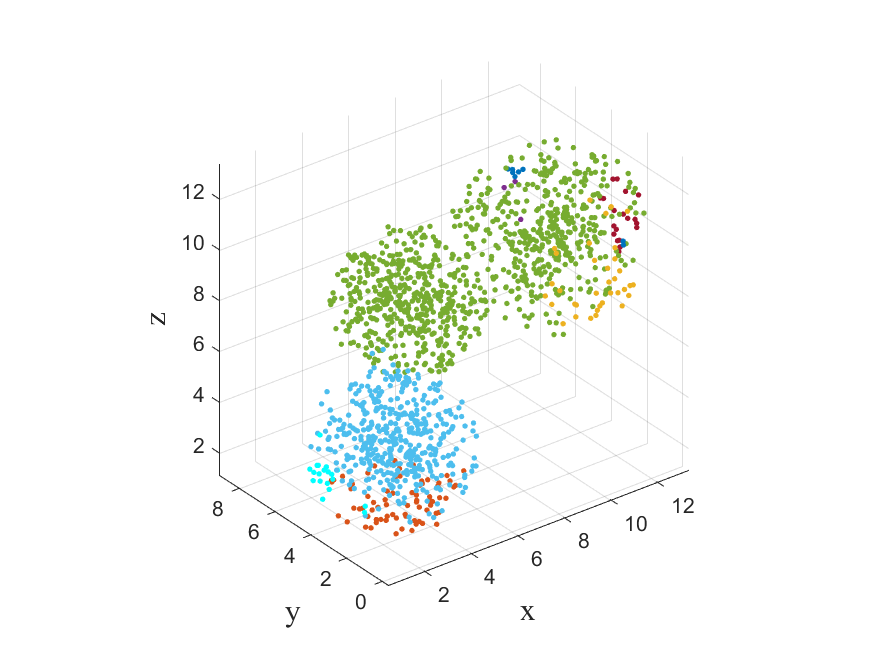}}
	\subfloat[\label{fig:Hd2}]{
		\includegraphics[width=0.25\linewidth]{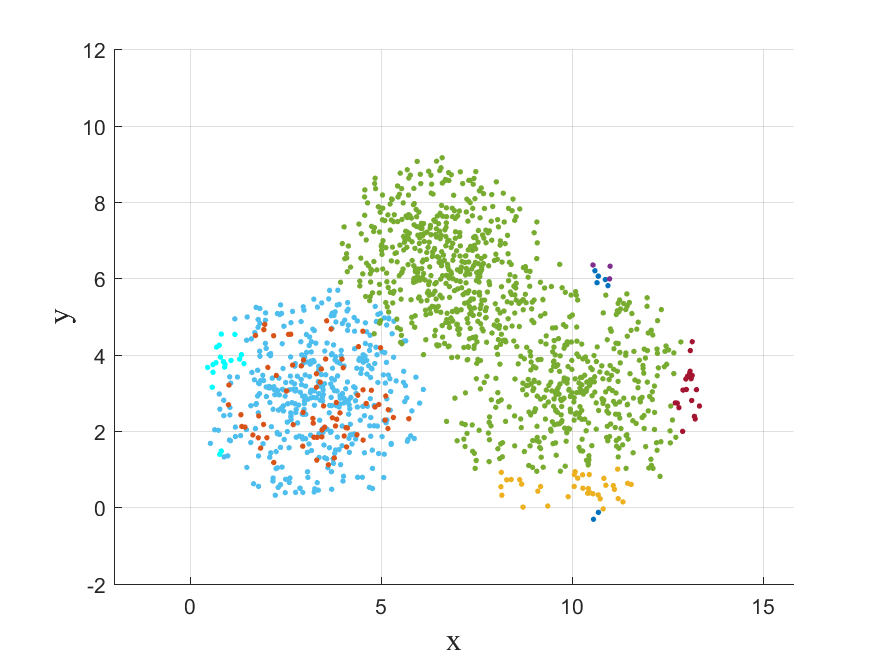}}
	\subfloat[\label{fig:Hd3}]{
		\includegraphics[width=0.25\linewidth]{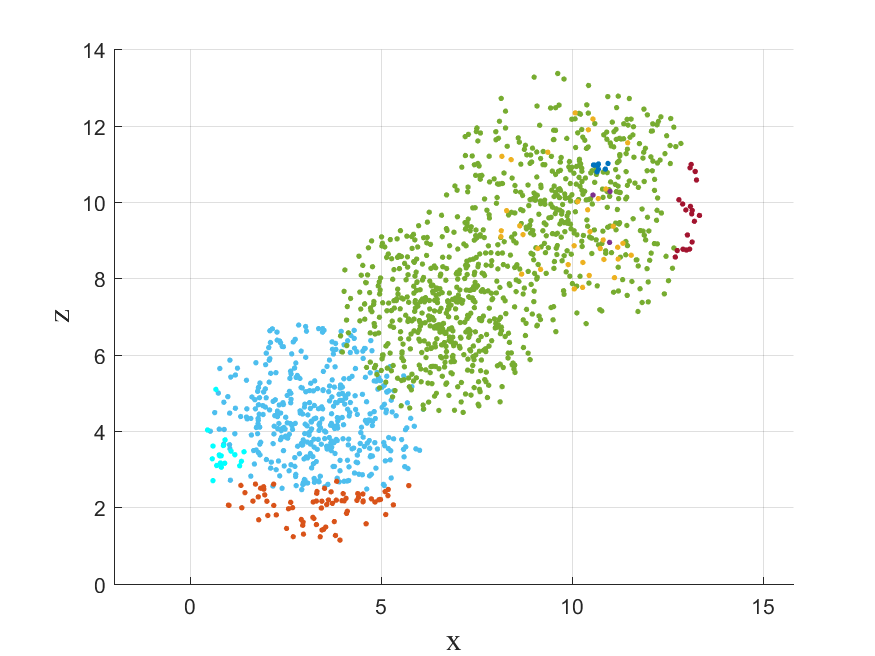}}
	\subfloat[\label{fig:Hd4}]{
		\includegraphics[width=0.25\linewidth]{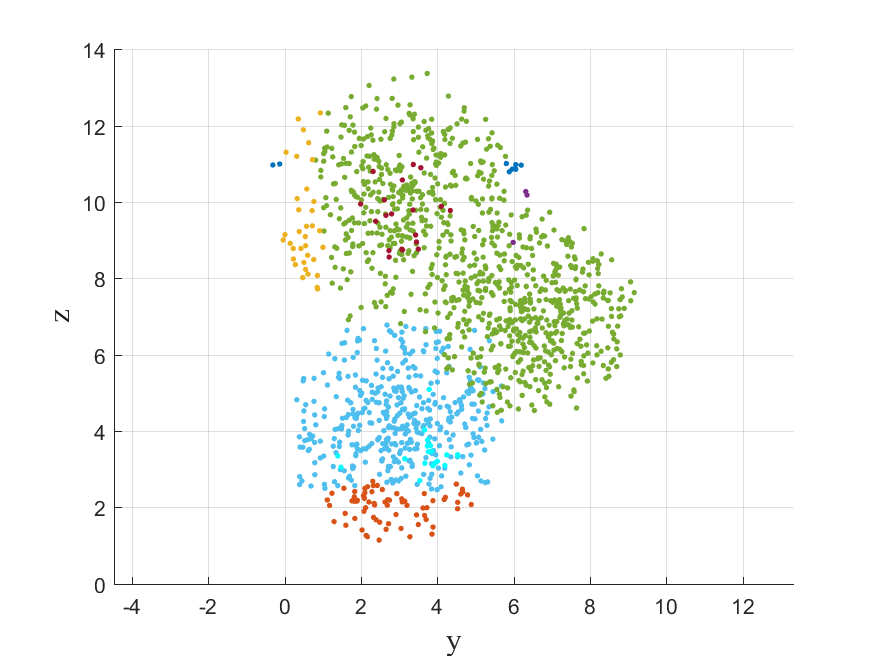}}
	\caption{Results of hard partitions on the 3DBall dataset. The first column is the 3D distribution, and the second to the fourth column are the three views (xy plane, yz plane, and xz plane), respectively,
		(a)-(d) CGD ($C=3$); 
		(e)-(h) CDMGC ($C=3$); 
		(i)-(l) CGD ($C=2^3$); 
		(m)-(p) CDMGC ($C=2^3$).}
	\label{fig:toyH} 
\end{figure*}

\begin{figure*} [t!]
	\centering
	\subfloat[\label{fig:Fa1}]{
		\includegraphics[width=0.25\linewidth]{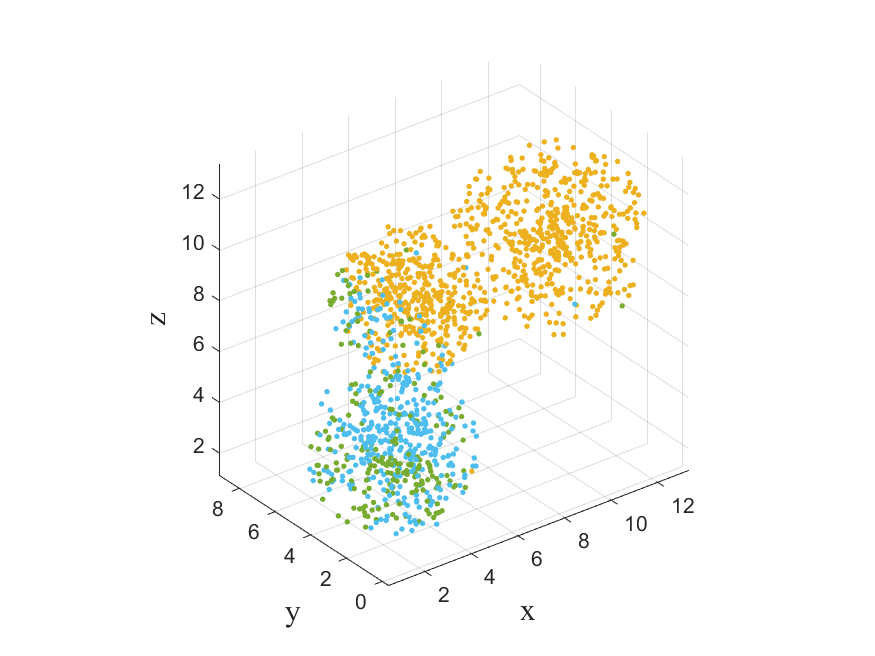}}
	\subfloat[\label{fig:Fa2}]{
		\includegraphics[width=0.25\linewidth]{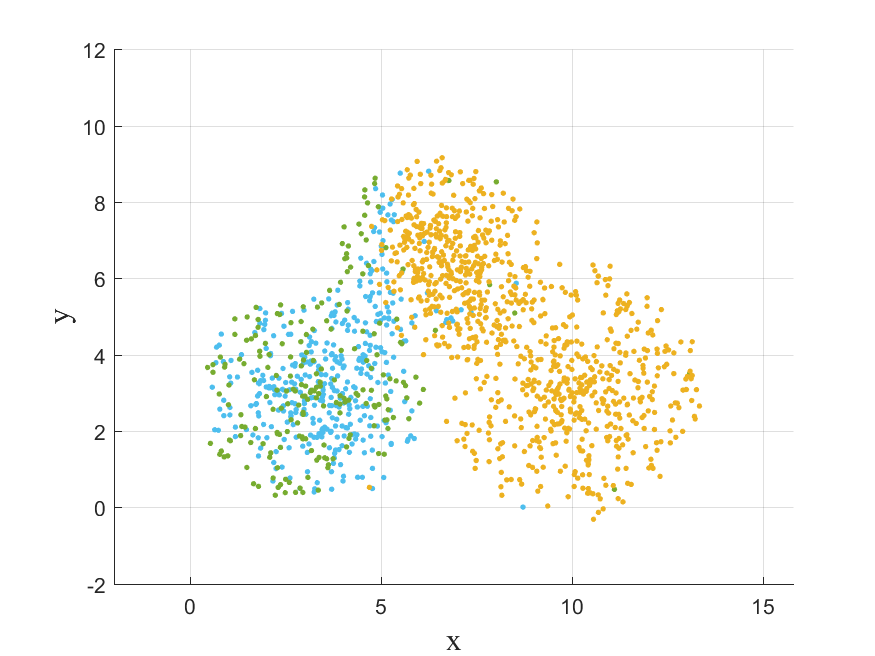}}
	\subfloat[\label{fig:Fa3}]{
		\includegraphics[width=0.25\linewidth]{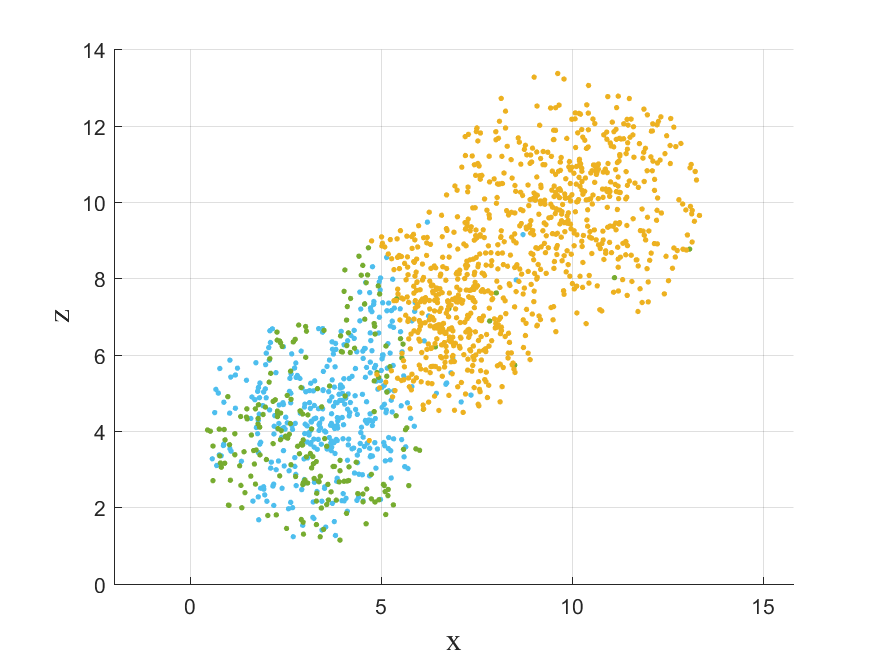}}
	\subfloat[\label{fig:Fa4}]{
		\includegraphics[width=0.25\linewidth]{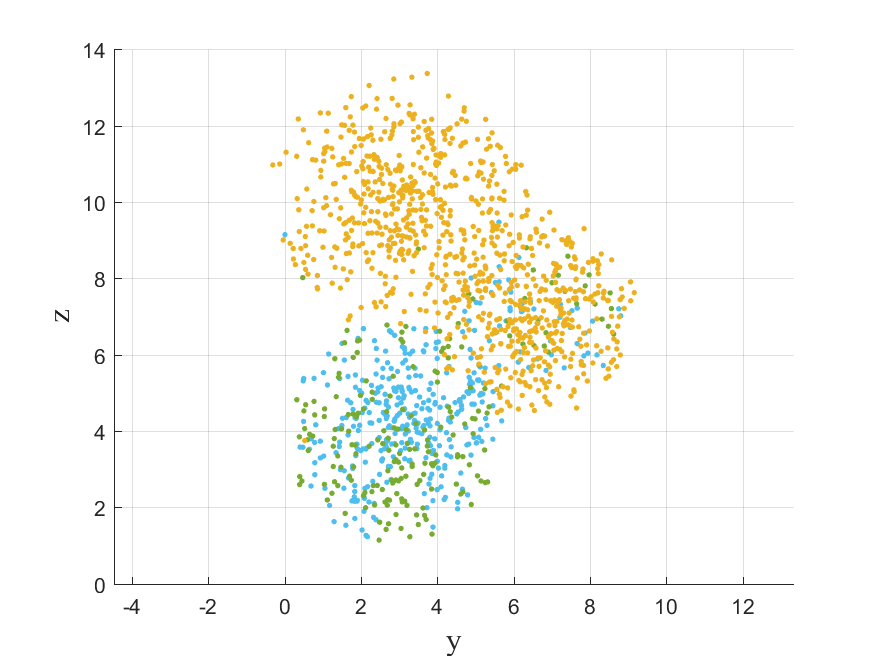}}
	\\
	\subfloat[\label{fig:Fb1}]{
		\includegraphics[width=0.25\linewidth]{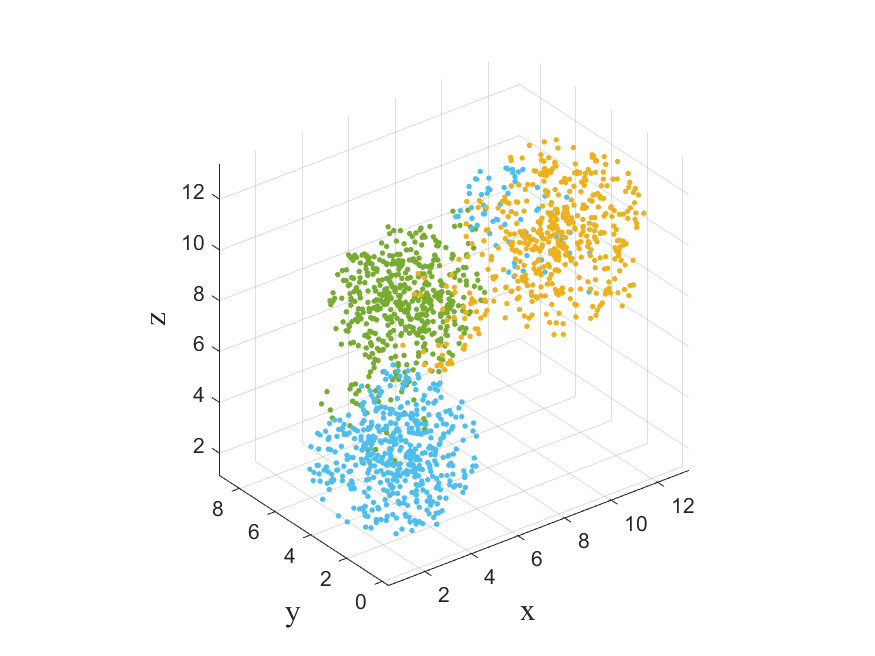}}
	\subfloat[\label{fig:Fb2}]{
		\includegraphics[width=0.25\linewidth]{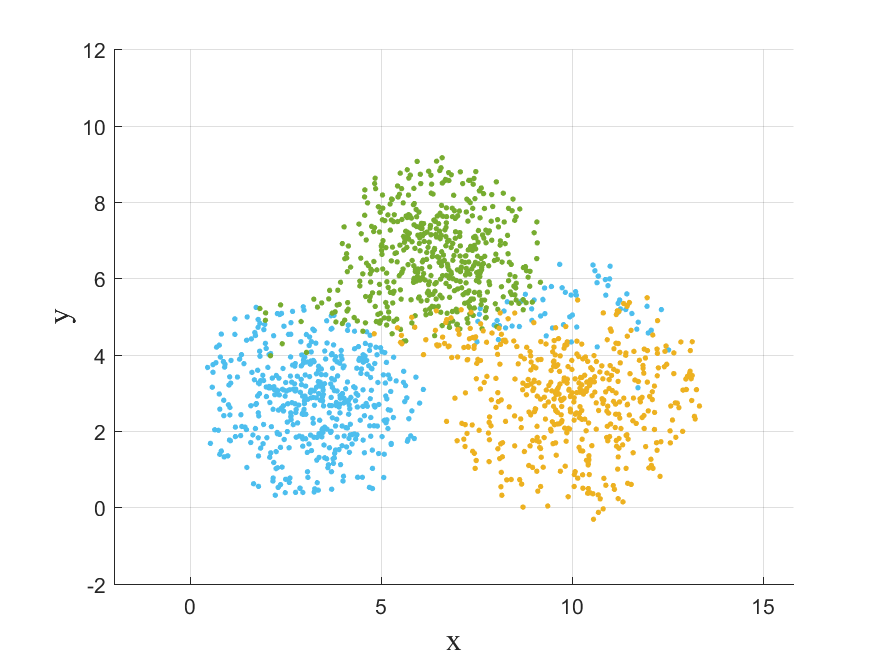}}
	\subfloat[\label{fig:Fb3}]{
		\includegraphics[width=0.25\linewidth]{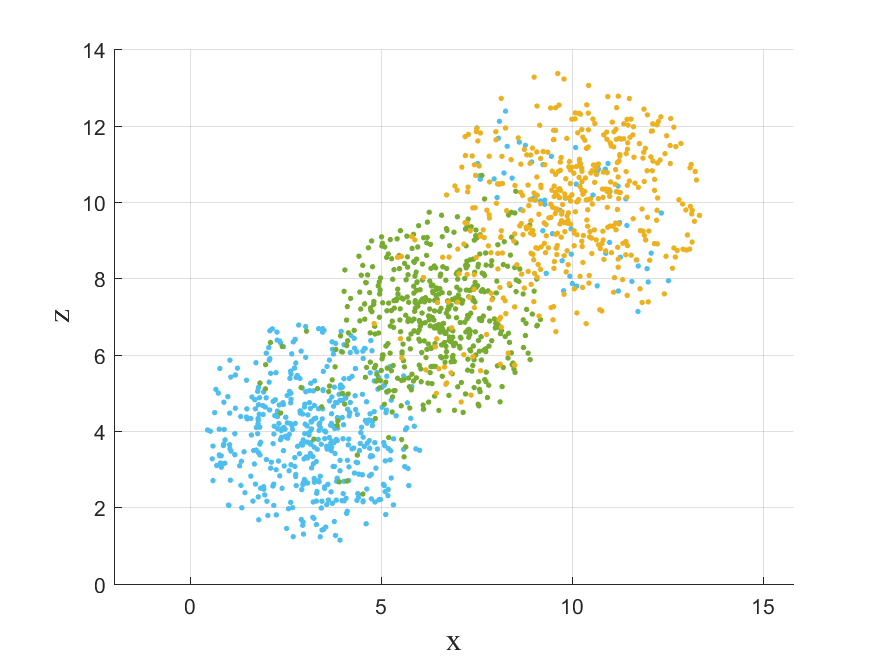}}
	\subfloat[\label{fig:Fb4}]{
		\includegraphics[width=0.25\linewidth]{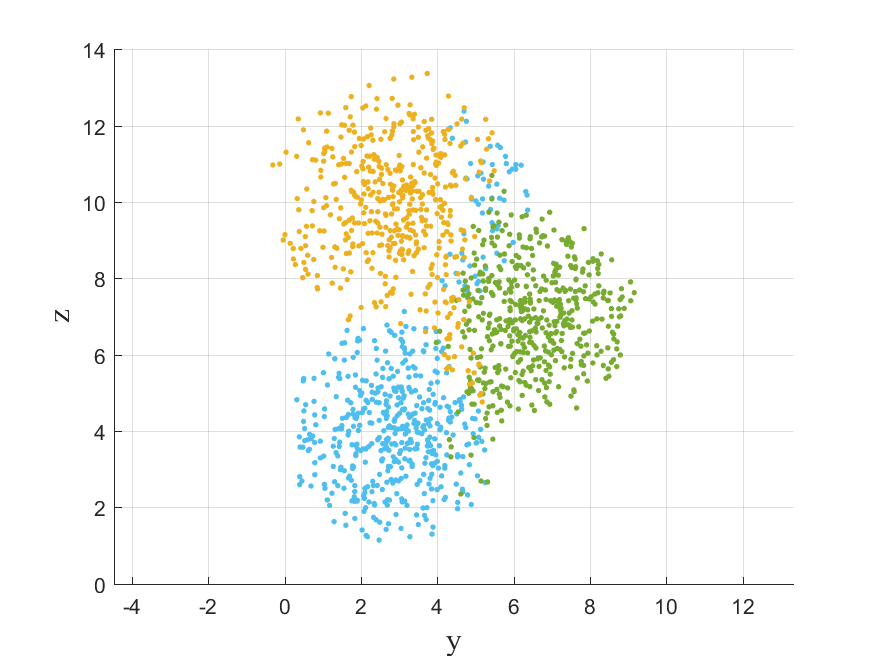}}
	\\
	\subfloat[\label{fig:Fc1}]{
		\includegraphics[width=0.25\linewidth]{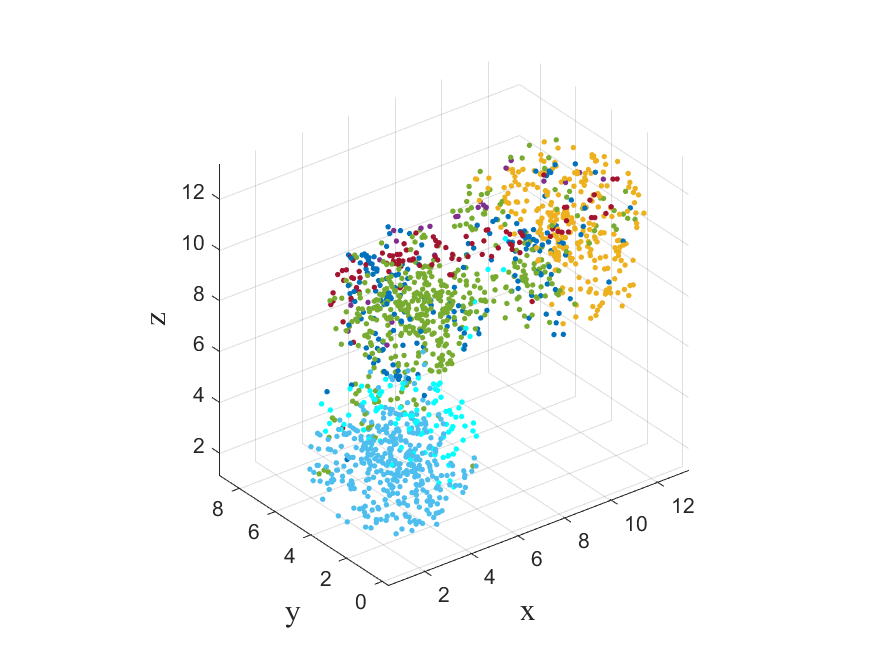}}
	\subfloat[\label{fig:Fc2}]{
		\includegraphics[width=0.25\linewidth]{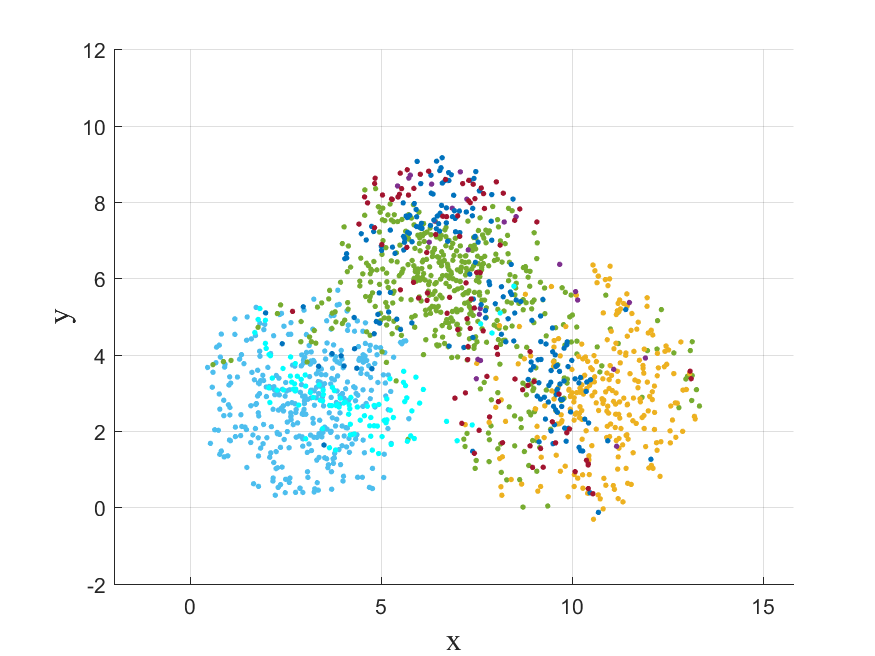}}
	\subfloat[\label{fig:Fc3}]{
		\includegraphics[width=0.25\linewidth]{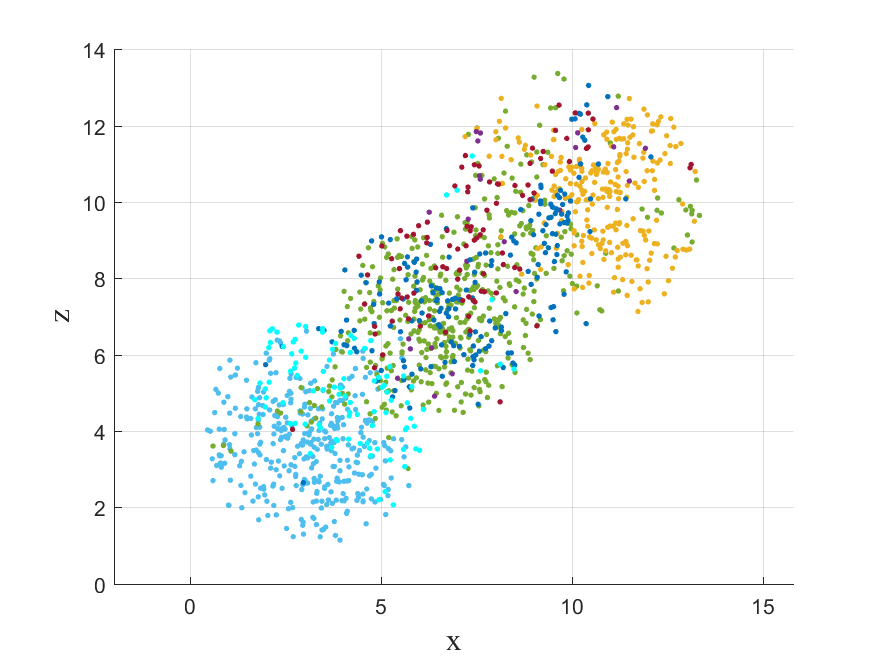}}
	\subfloat[\label{fig:Fc4}]{
		\includegraphics[width=0.25\linewidth]{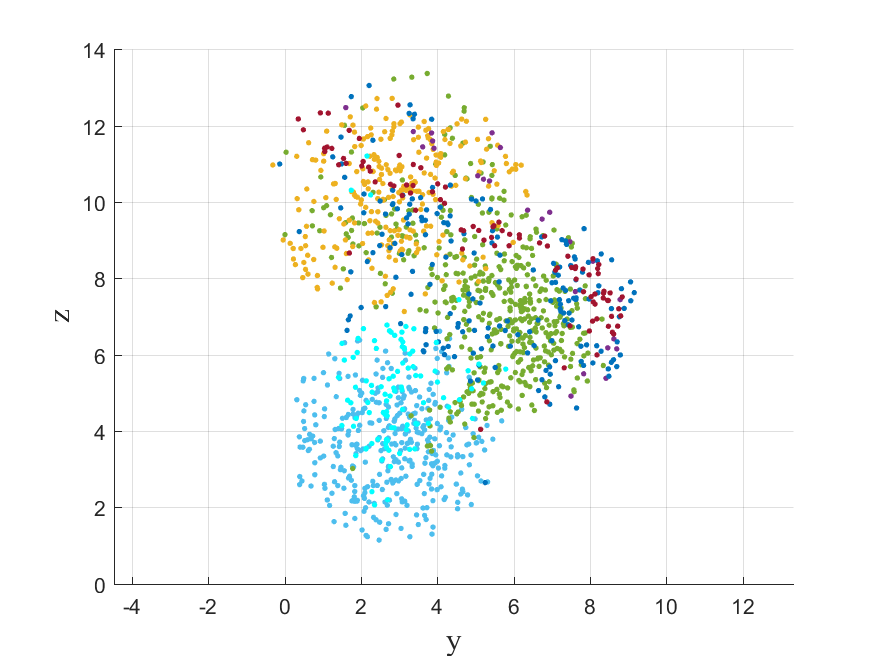}}
	\\
	\subfloat[\label{fig:Fd1}]{
		\includegraphics[width=0.25\linewidth]{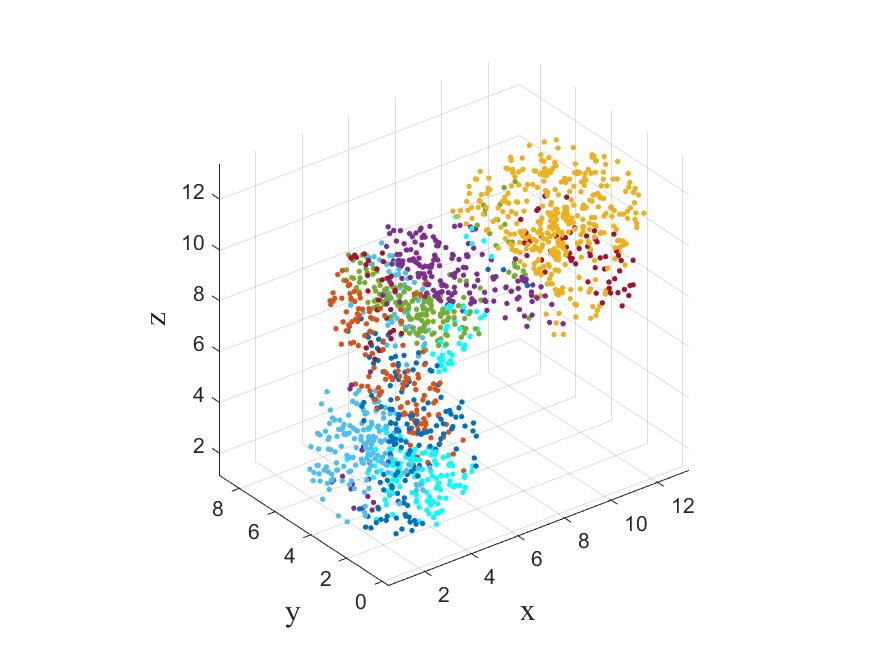}}
	\subfloat[\label{fig:Fd2}]{
		\includegraphics[width=0.25\linewidth]{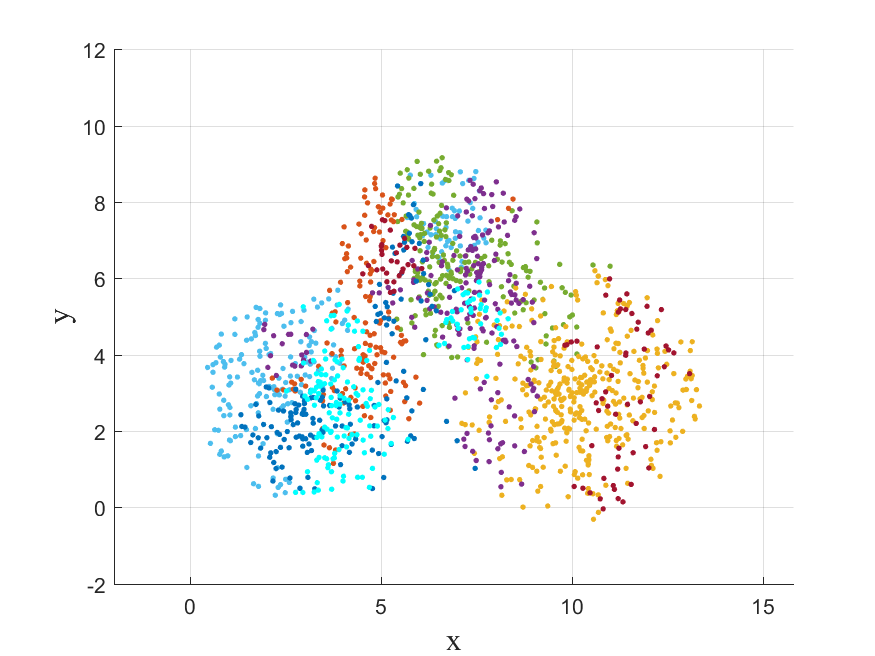}}
	\subfloat[\label{fig:Fd3}]{
		\includegraphics[width=0.25\linewidth]{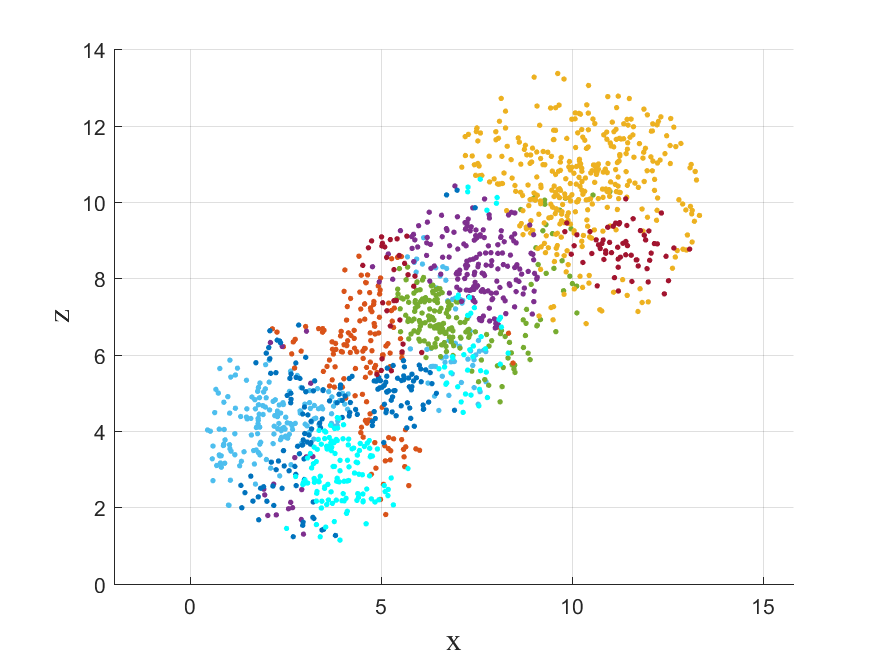}}
	\subfloat[\label{fig:Fd4}]{
		\includegraphics[width=0.25\linewidth]{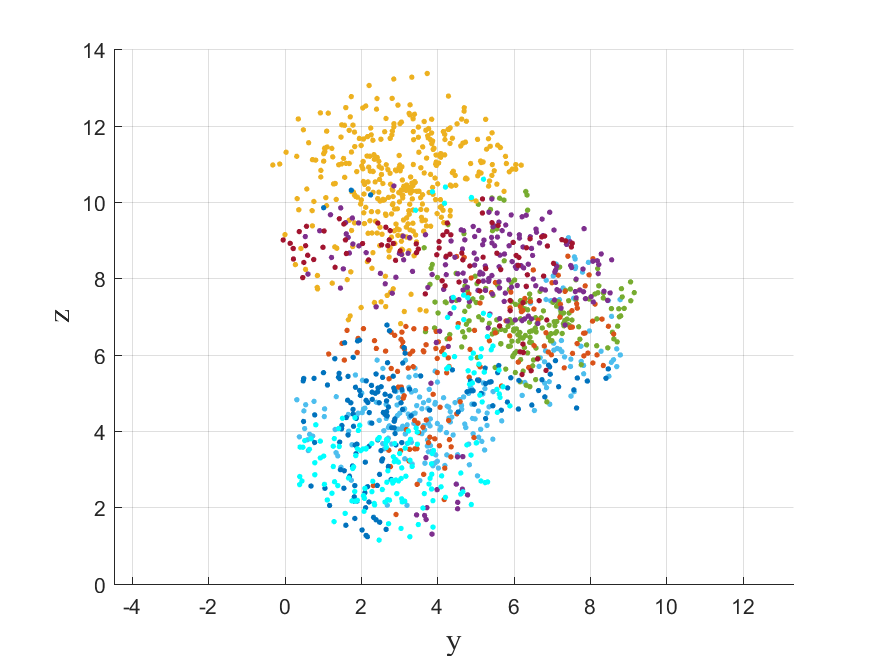}}
	\caption{Results of fuzzy partitions on the 3DBall dataset. The first column is the 3D distribution, and the second to the fourth column are the three views (xy plane, yz plane, and xz plane), respectively;
		(a)-(d) CoFKM ($C=3$), 
		(e)-(h) Co-FW-MVFCM ($C=3$),
		(i)-(l) CoFKM ($C=2^3$), 
		(m)-(p) Co-FW-MVFCM ($C=2^3$).}
	\label{fig:toyF} 
\end{figure*}

First, Fig.~\ref{fig:toyS} gives the complete version of Fig.~\ref{fig:toy} in the main manuscript, which includes the 3D distribution and three views of the original dataset and the results of MvLRECM when $\alpha=\{1,2,3\}$. 
In addition, we analyze the performance of hard and fuzzy partitions on the 3DBall dataset to show how they deal with the dataset which has overlappings between clusters.
We apply two hard partitions, CDG and CDMGC, and two fuzzy partitions, CoFKM and Co-FW-MVFCM, to the 3DBall dataset, and set the number of clusters $C=3$ and $C=2^3$ for each method. The results are shown in Fig.~\ref{fig:toyH} and Fig.~\ref{fig:toyF}. 
it is demonstrated that both hard and fuzzy partitions cannot characterize imprecision whether we set $C=\Omega$ or $C=2^\Omega$ without defining singleton cluster and meta-cluster separately. 
The invalid results of hard and fuzzy partitions show the necessity of credal partition for characterizing imprecision. 

\subsection{Experiments on Mnist179 dataset}
\label{sec:Sreal}
%
As a supplement to Table~\ref{tab:num} in the main manuscript, Table~\ref{tab:SMnist} shows the clustering performance on the Mnist179 dataset. It should be noted that Co-FW-MVFCM is not included because it cannot obtain valid results on the Mnist179 dataset, and we introduce another hard partition method, MVASM~\cite{han2020multi}\footnote{We set the parameters of MVASM as $\gamma=0.5, q=2$.}, for comparison.
\begin{table*}
	\begin{center}
		\caption{Clustering performance on the Mnist179 dataset.}
		\label{tab:SMnist}
		\begin{tabular}{cccccccccc}
			\toprule
			&Methods& ACC & NMI & Purity &  F-Score & Precision & Recall & RI & IR\\
			\midrule
			&\boldmath$\operatorname{CDMGC}$&0.3446&0.0193&0.3578&0.4946&0.3336&\textbf{0.9561}&0.3473&0.0000\\
			&\boldmath$\operatorname{CGD}$&0.4770&0.3222&0.5952&0.5543&0.4468&0.7300&0.6079&0.0000\\
			&\boldmath$\operatorname{CoFKM}$&0.7844&\textbf{0.5276}&0.7844&0.6991&0.6951&0.7031&0.7978&0.0000\\
			&\boldmath$\operatorname{LMSC}$&\textbf{0.7900}&0.4555&\textbf{0.7900}& 0.6624 &0.6601&0.6648&0.7749&0.0000\\
			&\boldmath$\operatorname{LTMSC}$&0.5145&0.4508&0.6579&0.6239&0.5512&0.7188&0.7105&0.0000\\
			&\boldmath$\operatorname{MCSSC}$&0.5214&0.2100&0.5347&0.4528&0.4385&0.4680&0.6221&0.0000\\
			&\boldmath$\operatorname{TBGL-MVC}$&0.4987&0.2085&0.5142&0.5187&0.3869&0.7867&0.5123&0.0000\\
			&\boldmath$\operatorname{MVASM}$&0.3966&0.0098&0.3966&0.3428&0.3411&0.3444&0.5588&0.0000\\
			&\boldmath$\operatorname{MvLRECM}$&0.3519&0.0018&0.3669&\textbf{0.8509}&\textbf{0.8518}&0.8499&\textbf{0.9055}&\textbf{0.3212}\\
			\bottomrule
		\end{tabular}
	\end{center}
\end{table*}

\subsection{Parameter study}
\label{sec:Spara}
\begin{figure*}
	\centering
	\subfloat[\label{fig:p1a1}]{
		\includegraphics[width=0.25\linewidth]{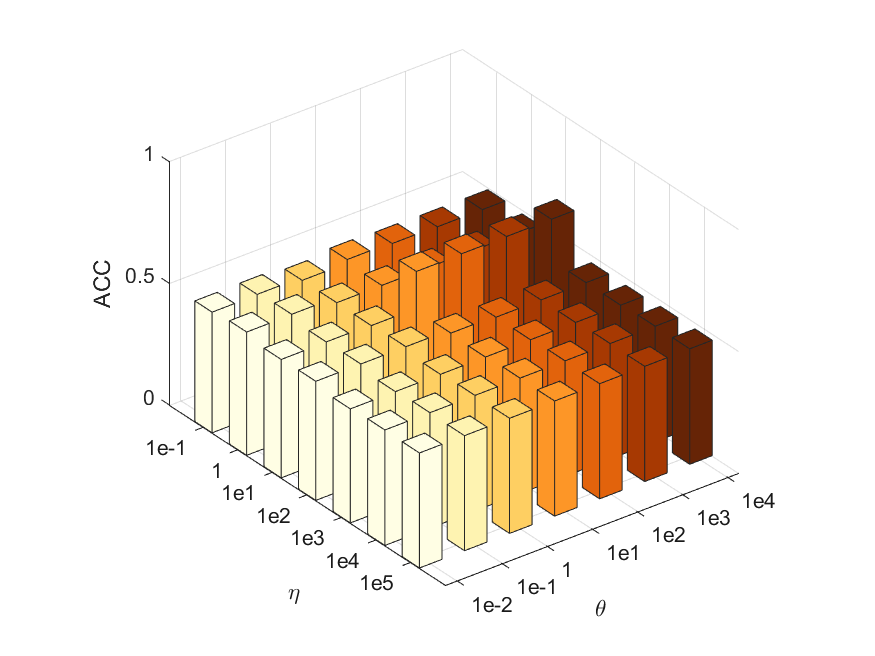}}
	\subfloat[\label{fig:p1a2}]{
		\includegraphics[width=0.25\linewidth]{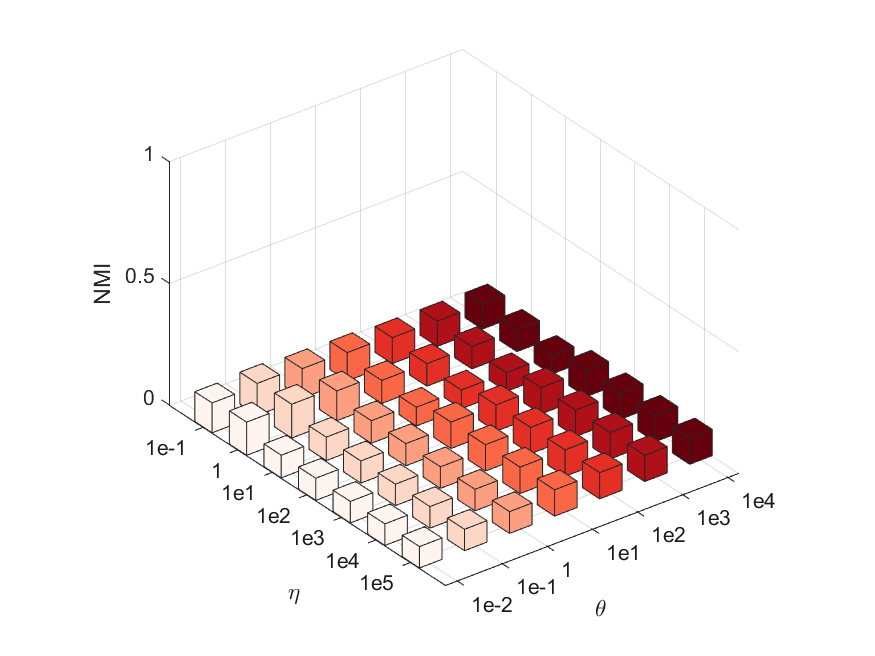}}
	\subfloat[\label{fig:p1a3}]{
		\includegraphics[width=0.25\linewidth]{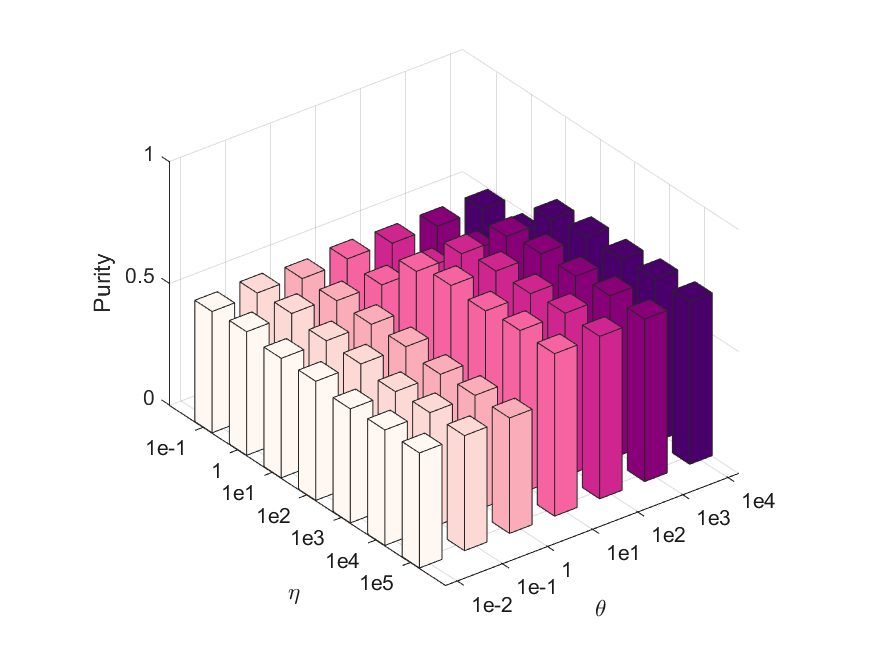}}
	\subfloat[\label{fig:p1a4}]{
		\includegraphics[width=0.25\linewidth]{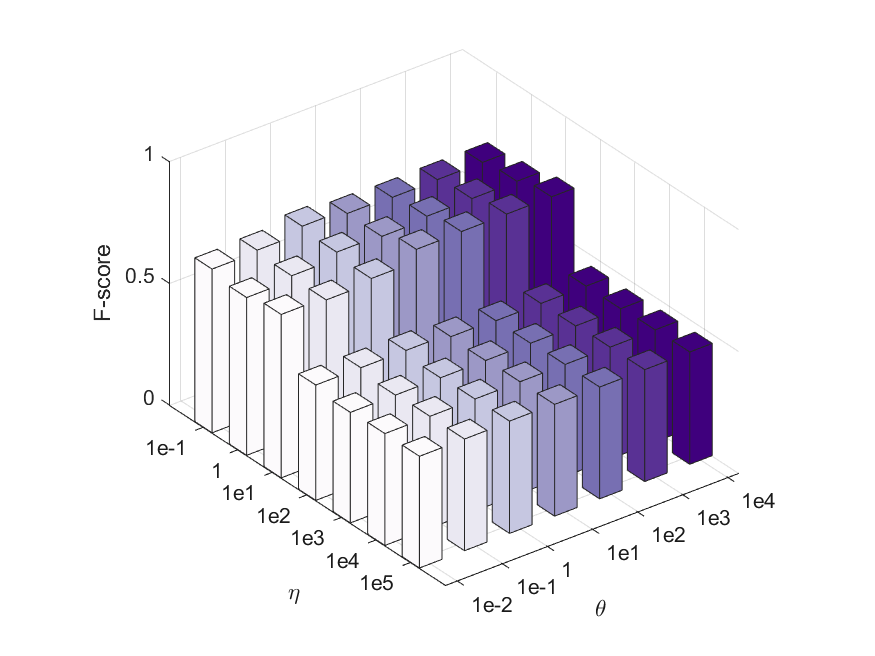}}
	\\
	\subfloat[\label{fig:p1b1}]{
		\includegraphics[width=0.25\linewidth]{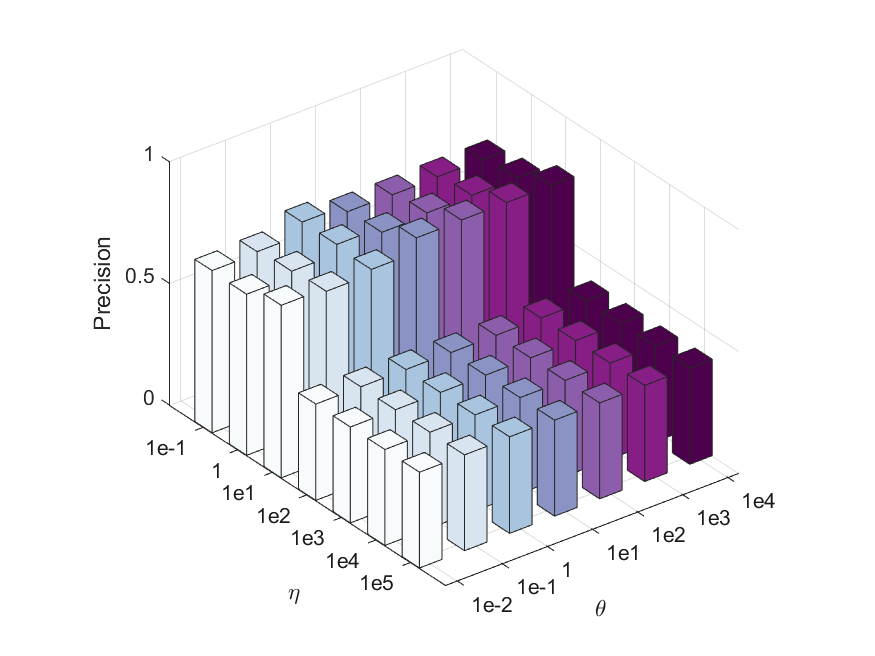}}
	\subfloat[\label{fig:p1b2}]{
		\includegraphics[width=0.25\linewidth]{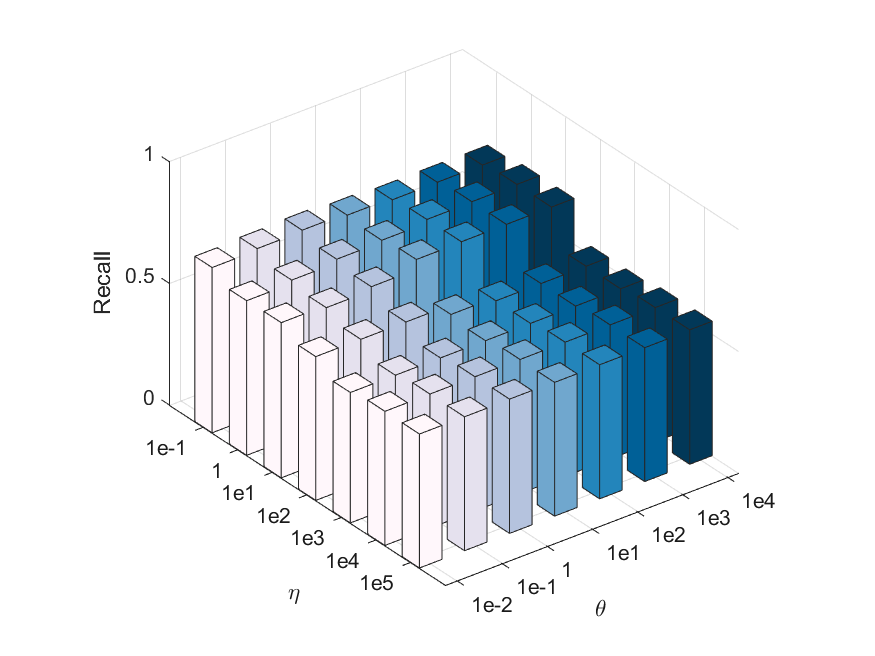}}
	\subfloat[\label{fig:p1b3}]{
		\includegraphics[width=0.25\linewidth]{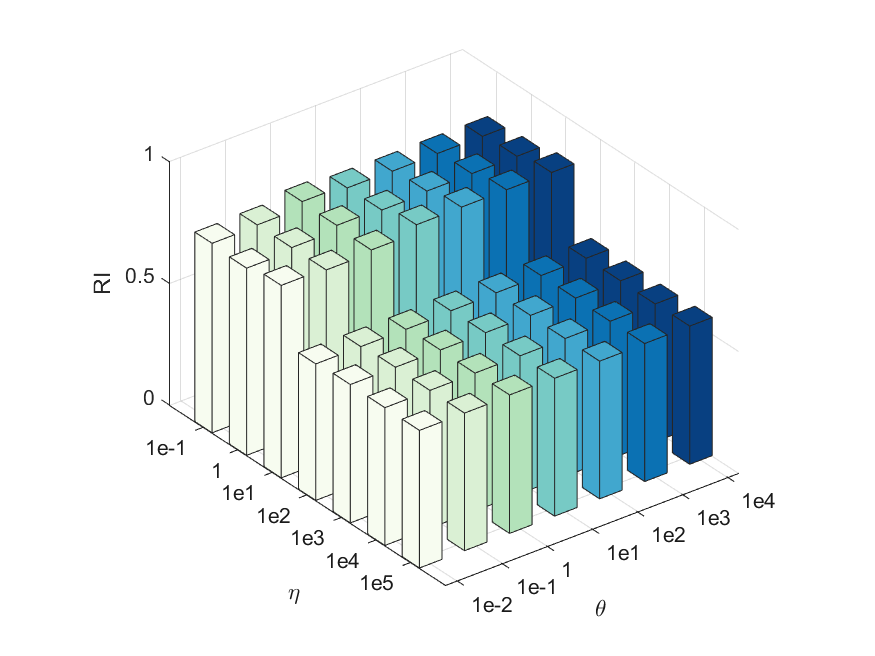}}
	\subfloat[\label{fig:p1b4}]{
		\includegraphics[width=0.25\linewidth]{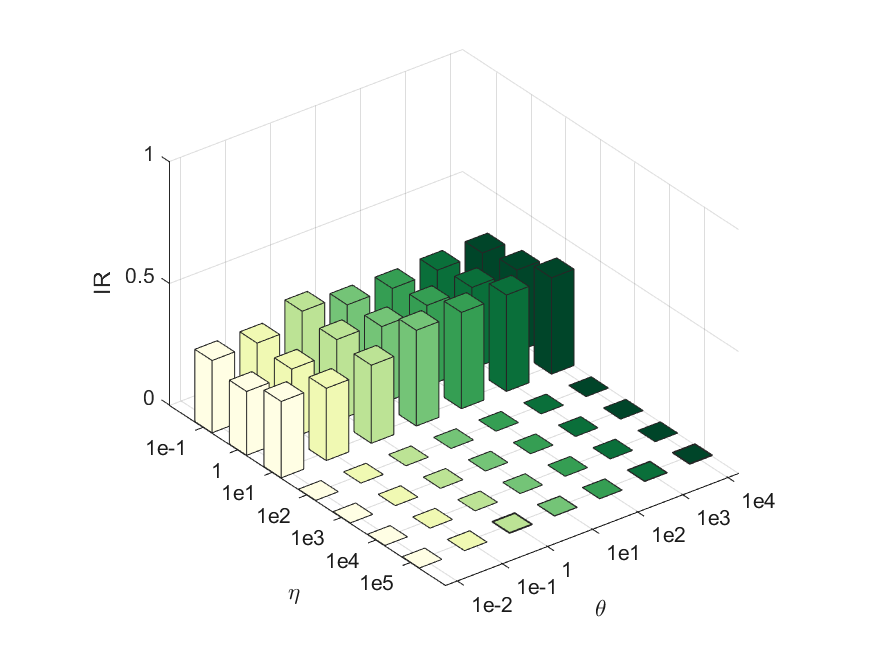}}
	\caption{How Parameter $\theta$ and $\eta$ effect the performance of MvLRECM on the Abalone dataset. (a)ACC, (b)NMI, (c)Purity, (d)Precision, (e)Recall, (f)F-score, (g)RI and (h)IR, where $x$ and $y$ axis are $\theta$ and $\eta$, respectively.}
	\label{fig:Spara1} 
\end{figure*}

\begin{figure*}
	\centering
	\subfloat[\label{fig:p2a1}]{
		\includegraphics[width=0.25\linewidth]{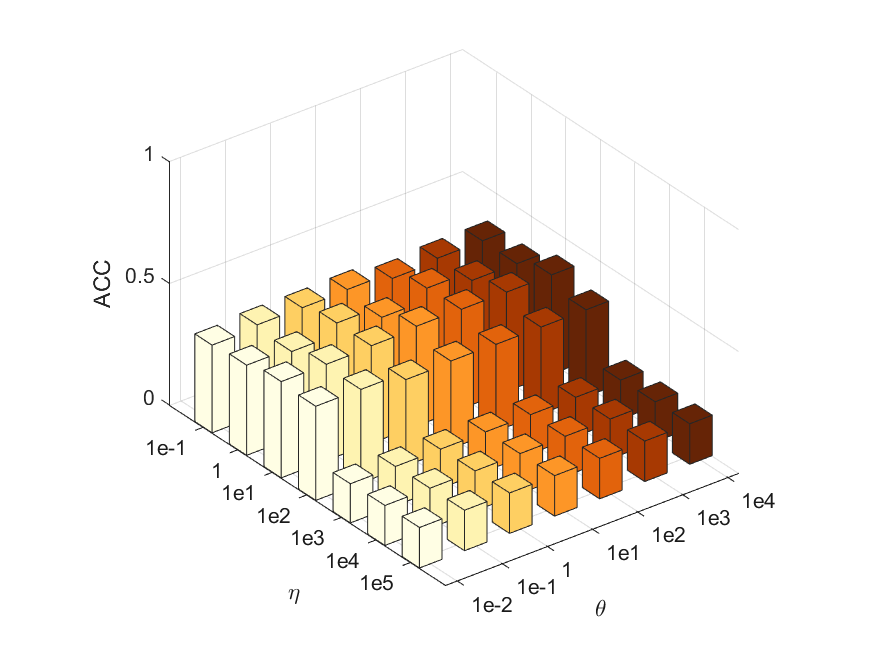}}
	\subfloat[\label{fig:p2a2}]{
		\includegraphics[width=0.25\linewidth]{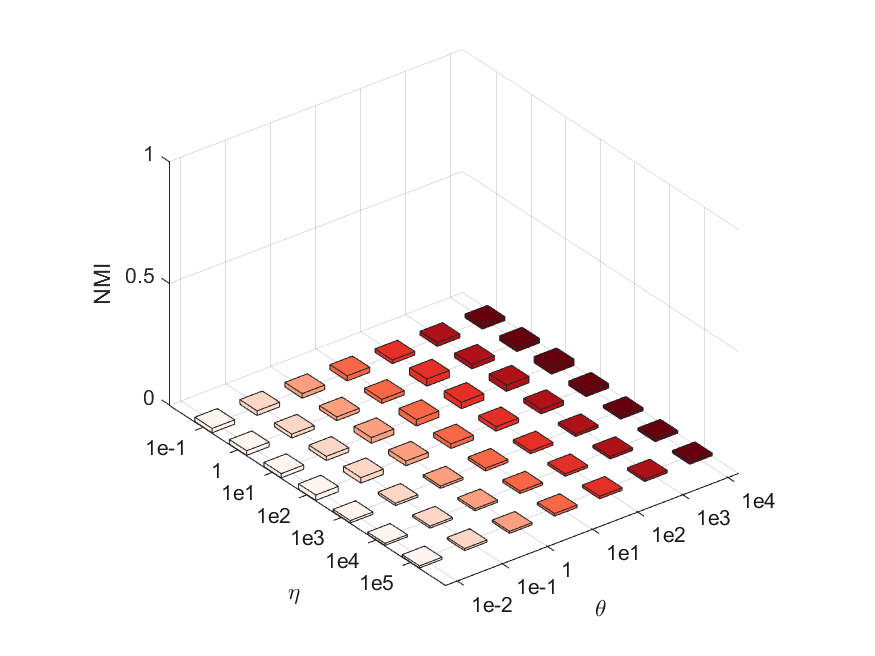}}
	\subfloat[\label{fig:p2a3}]{
		\includegraphics[width=0.25\linewidth]{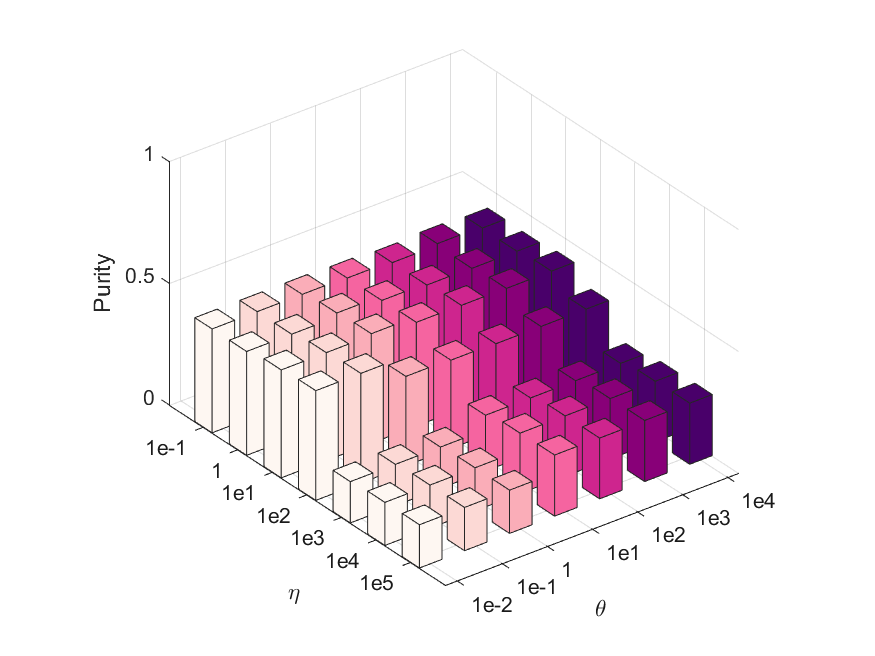}}
	\subfloat[\label{fig:p2a4}]{
		\includegraphics[width=0.25\linewidth]{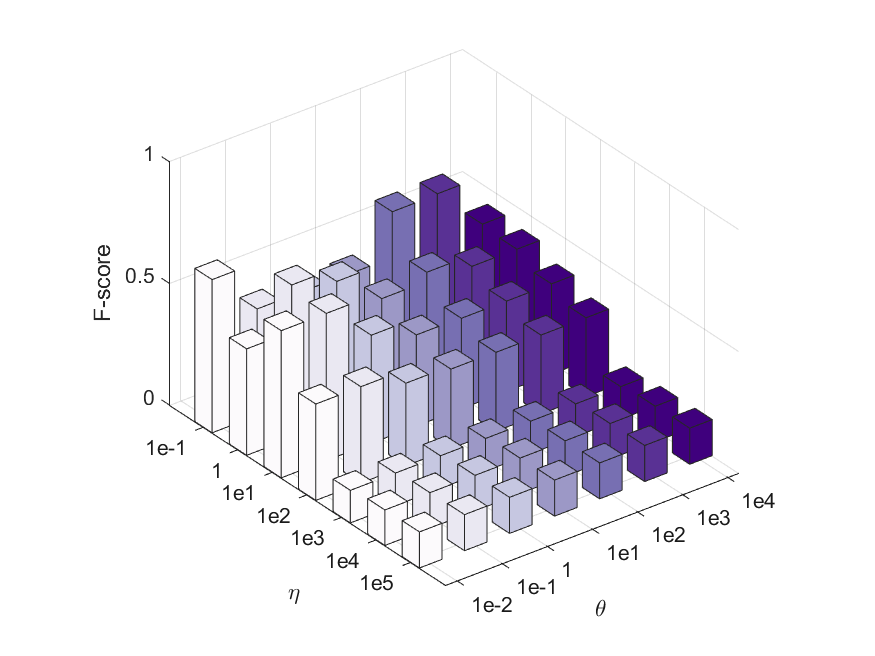}}
	\\
	\subfloat[\label{fig:p2b1}]{
		\includegraphics[width=0.25\linewidth]{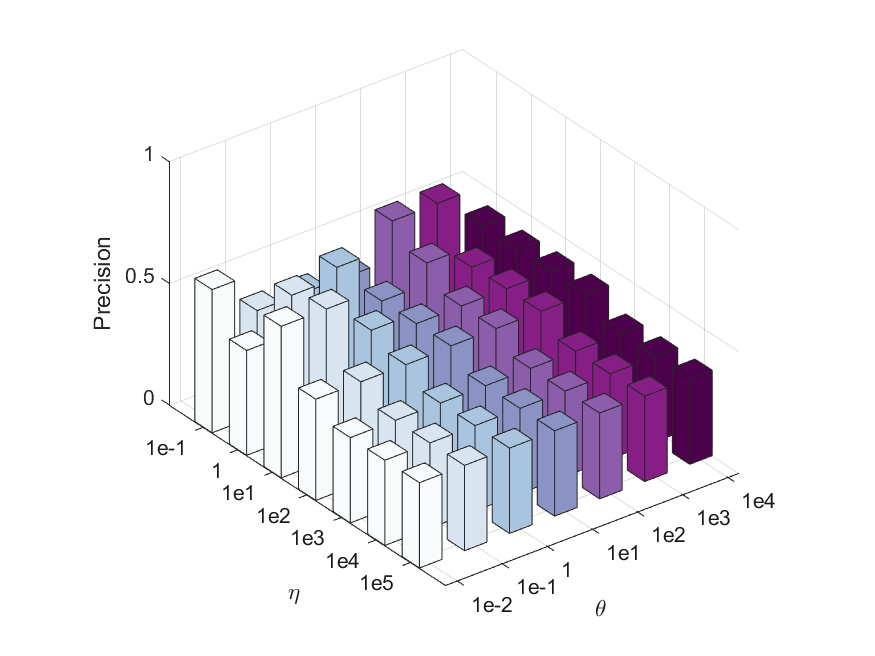}}
	\subfloat[\label{fig:p2b2}]{
		\includegraphics[width=0.25\linewidth]{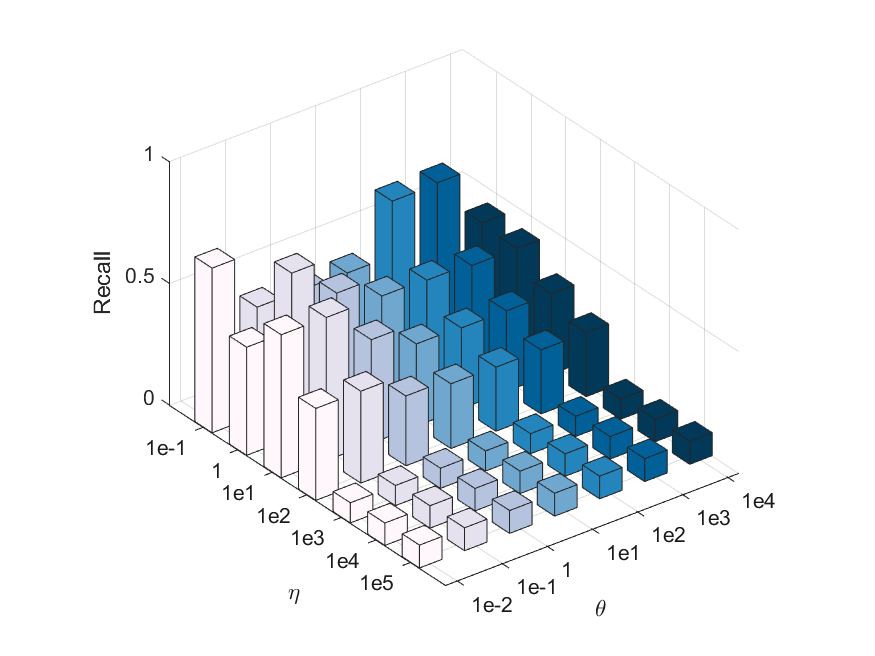}}
	\subfloat[\label{fig:p2b3}]{
		\includegraphics[width=0.25\linewidth]{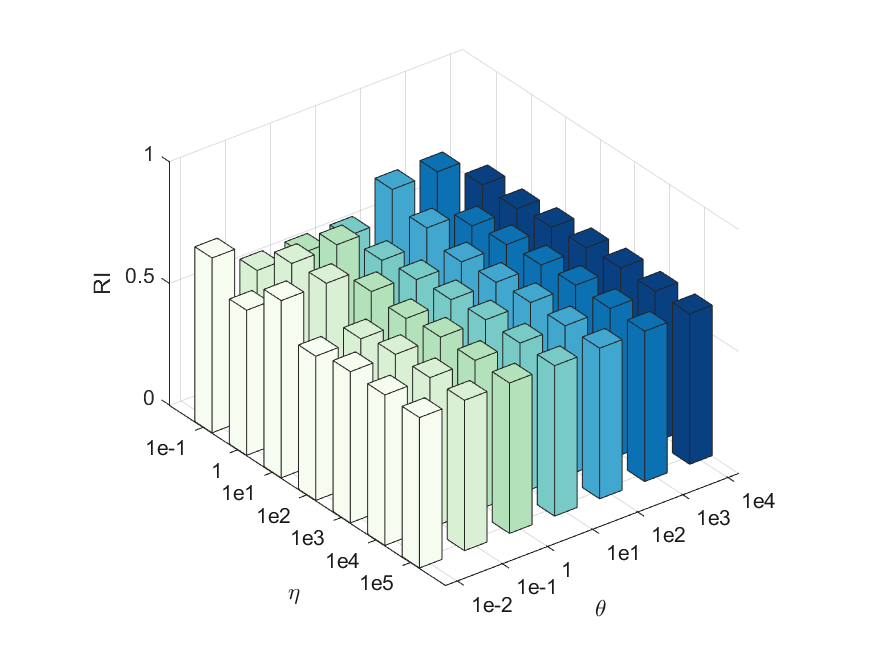}}
	\subfloat[\label{fig:p2b4}]{
		\includegraphics[width=0.25\linewidth]{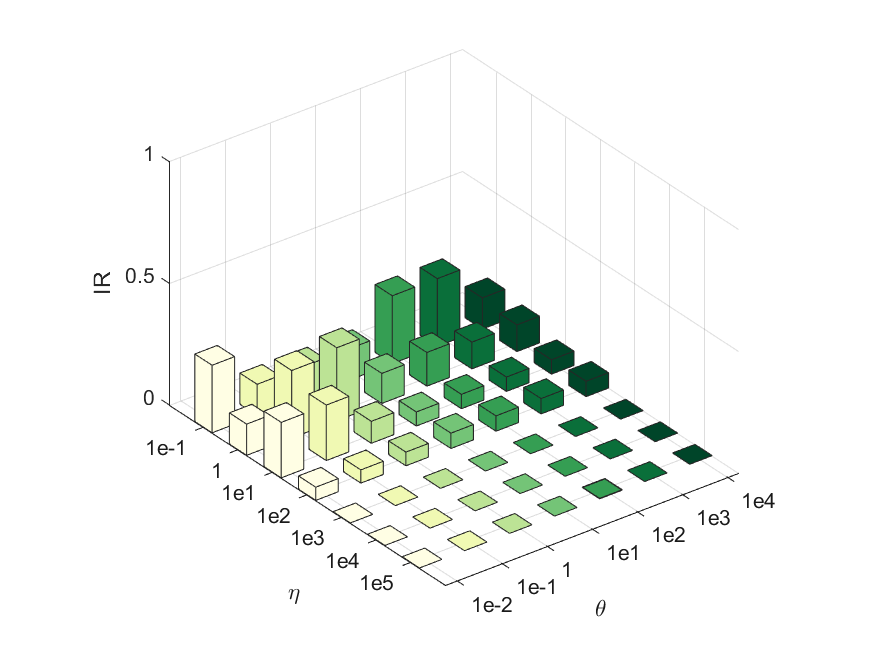}}
	\caption{How Parameter $\theta$ and $\eta$ effect the performance of MvLRECM on the Contraceptive dataset. (a)ACC, (b)NMI, (c)Purity, (d)Precision, (e)Recall, (f)F-score, (g)RI and (h)IR, where $x$ and $y$ axis are $\theta$ and $\eta$, respectively.}
	\label{fig:Spara2} 
\end{figure*}

\begin{figure*}
	\centering
	\subfloat[\label{fig:p3a1}]{
		\includegraphics[width=0.25\linewidth]{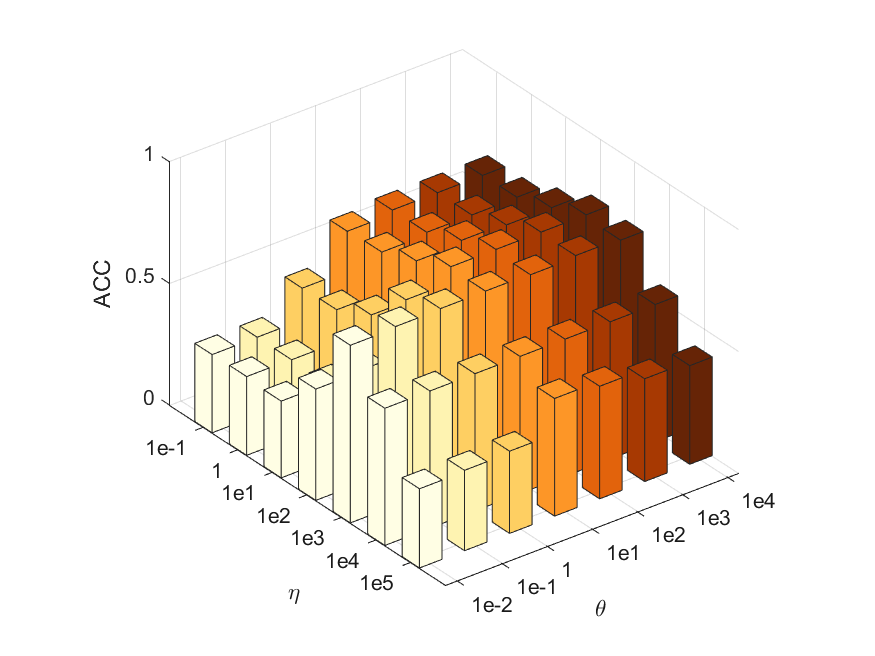}}
	\subfloat[\label{fig:p3a2}]{
		\includegraphics[width=0.25\linewidth]{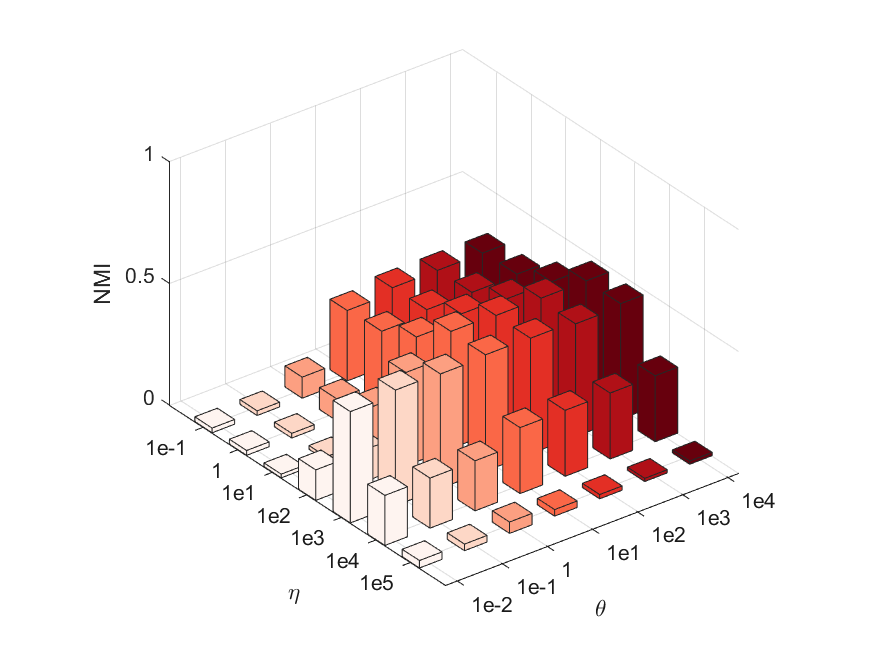}}
	\subfloat[\label{fig:p3a3}]{
		\includegraphics[width=0.25\linewidth]{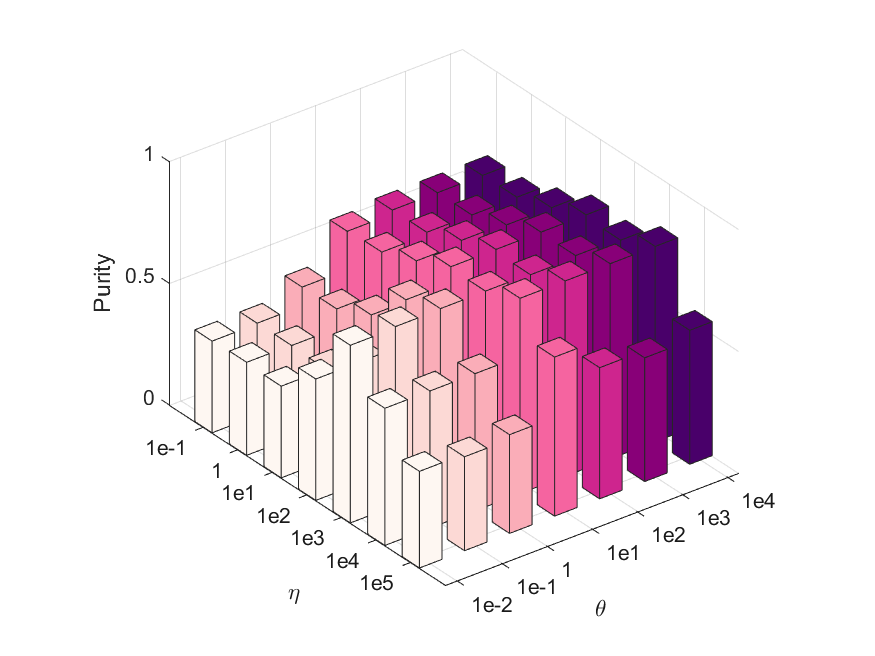}}
	\subfloat[\label{fig:p3a4}]{
		\includegraphics[width=0.25\linewidth]{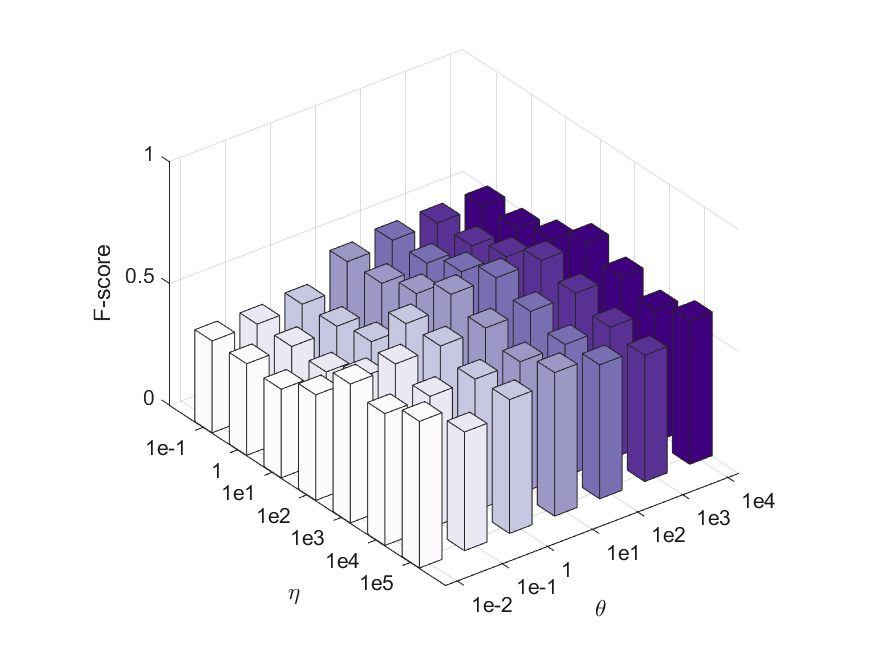}}
	\\
	\subfloat[\label{fig:p3b1}]{
		\includegraphics[width=0.25\linewidth]{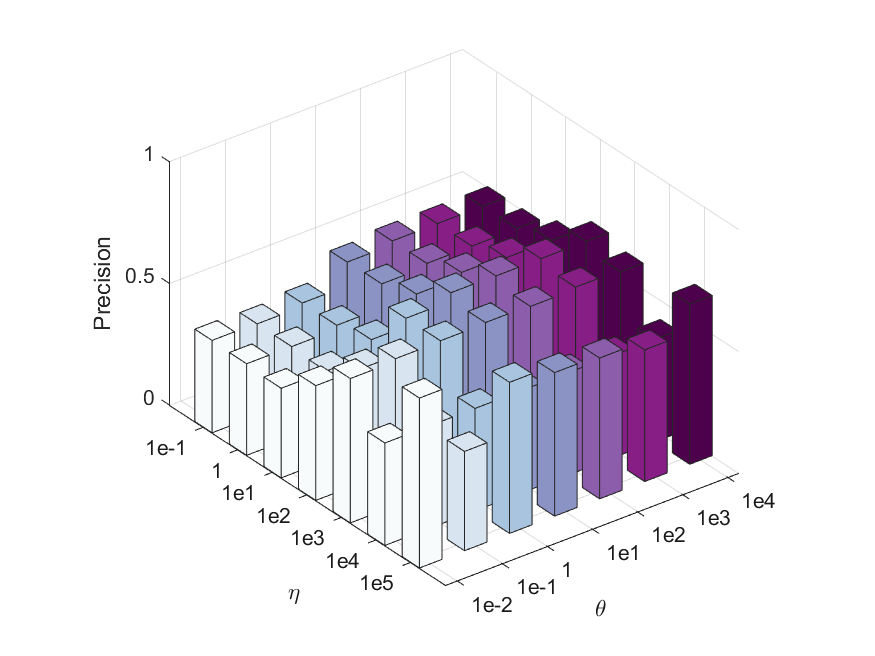}}
	\subfloat[\label{fig:p3b2}]{
		\includegraphics[width=0.25\linewidth]{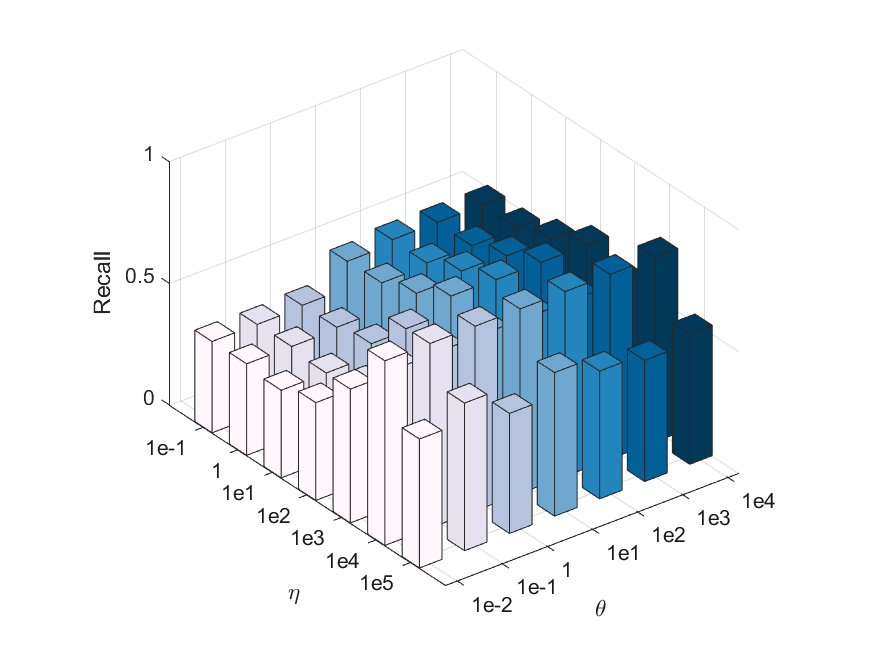}}
	\subfloat[\label{fig:p3b3}]{
		\includegraphics[width=0.25\linewidth]{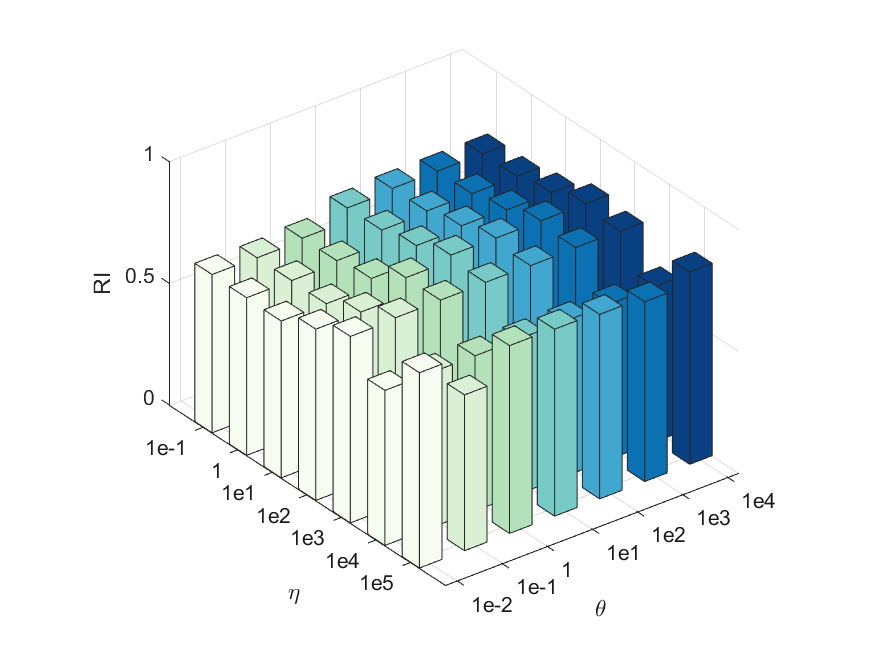}}
	\subfloat[\label{fig:p3b4}]{
		\includegraphics[width=0.25\linewidth]{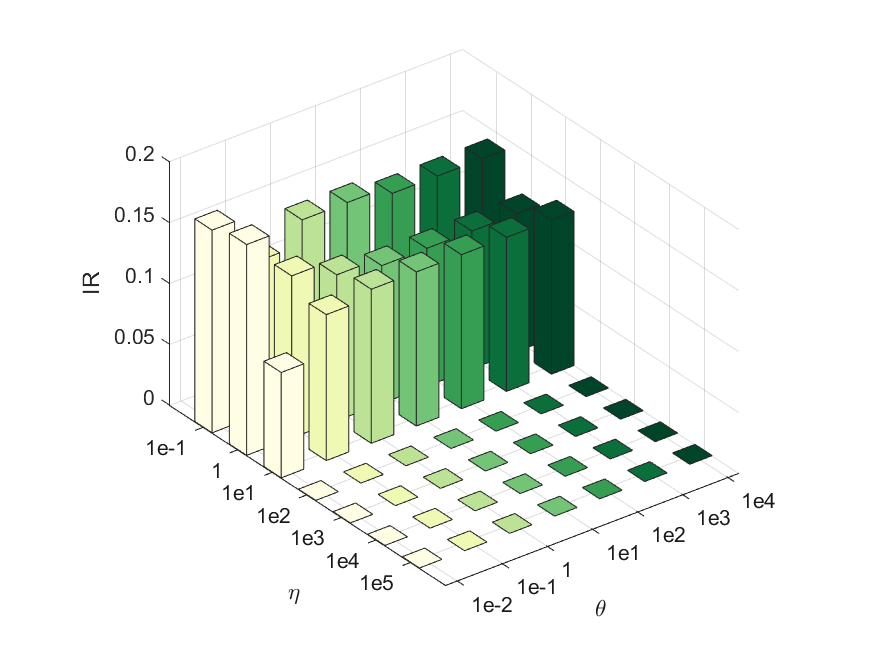}}
	\caption{How Parameter $\theta$ and $\eta$ effect the performance of MvLRECM on the Foresttype dataset. (a)ACC, (b)NMI, (c)Purity, (d)Precision, (e)Recall, (f)F-score, (g)RI and (h)IR, where $x$ and $y$ axis are $\theta$ and $\eta$, respectively.}
	\label{fig:Spara3} 
\end{figure*}

\begin{figure*}
	\centering
	\subfloat[\label{fig:p4a1}]{
		\includegraphics[width=0.25\linewidth]{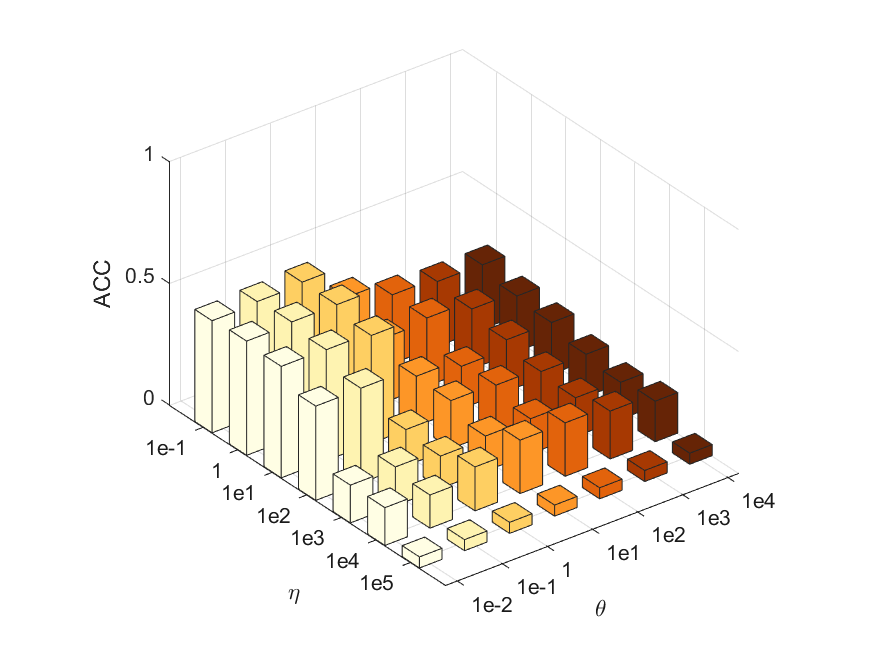}}
	\subfloat[\label{fig:p4a2}]{
		\includegraphics[width=0.25\linewidth]{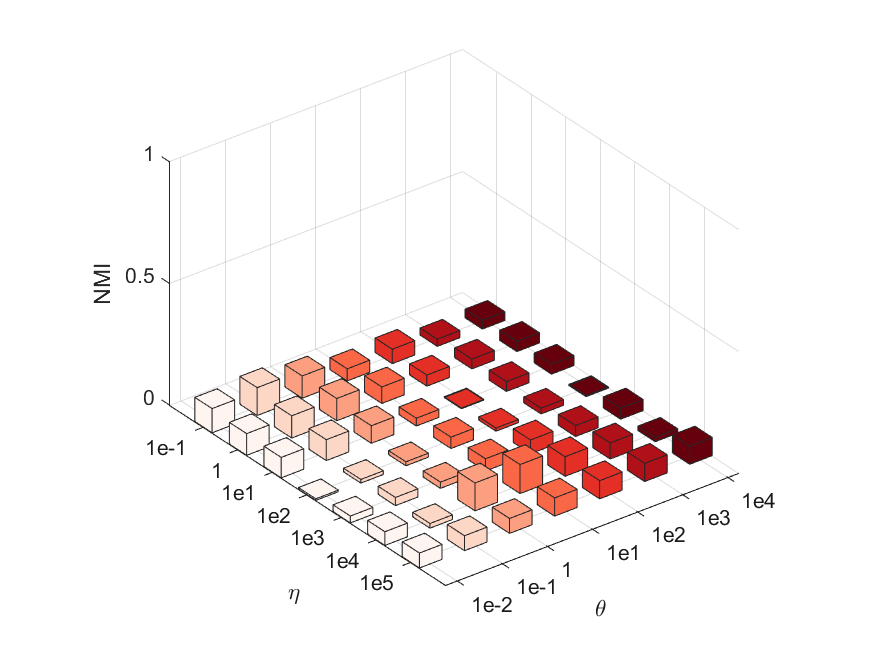}}
	\subfloat[\label{fig:p4a3}]{
		\includegraphics[width=0.25\linewidth]{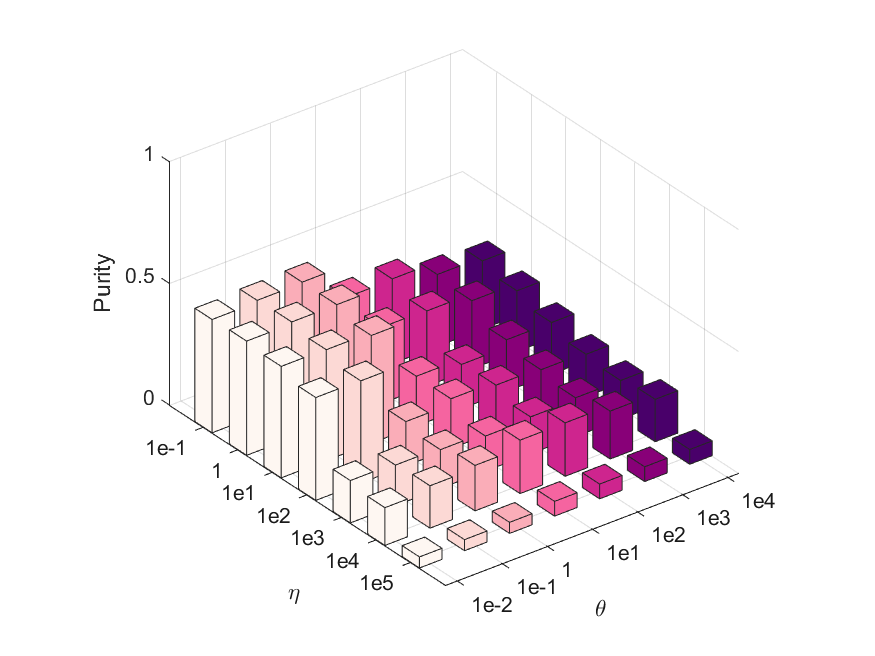}}
	\subfloat[\label{fig:p4a4}]{
		\includegraphics[width=0.25\linewidth]{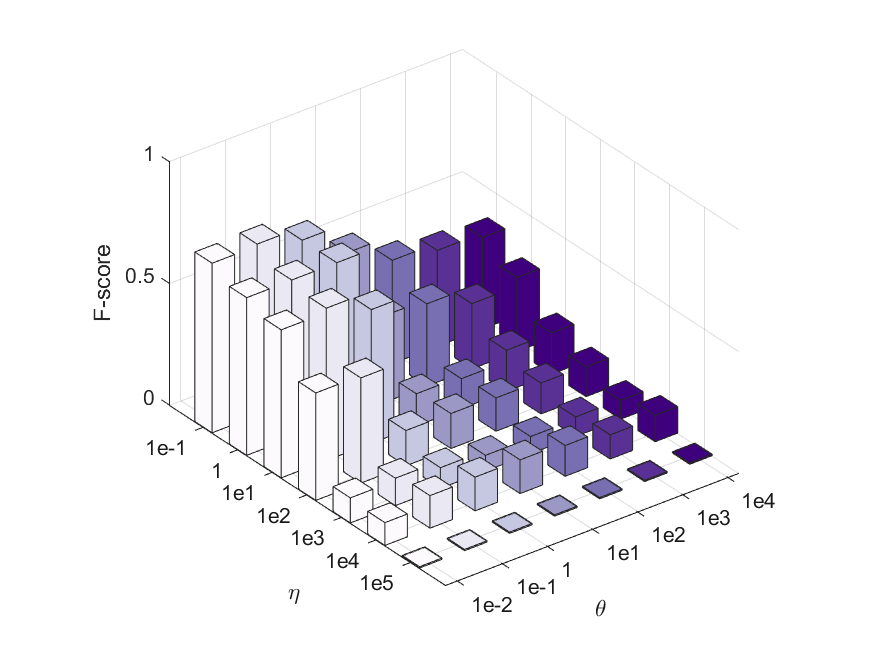}}
	\\
	\subfloat[\label{fig:p4b1}]{
		\includegraphics[width=0.25\linewidth]{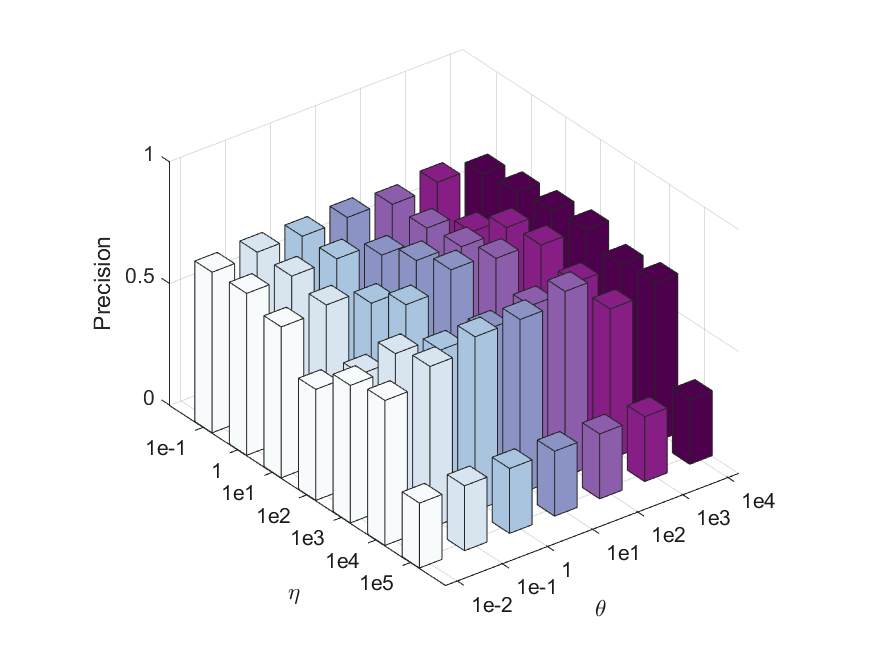}}
	\subfloat[\label{fig:p4b2}]{
		\includegraphics[width=0.25\linewidth]{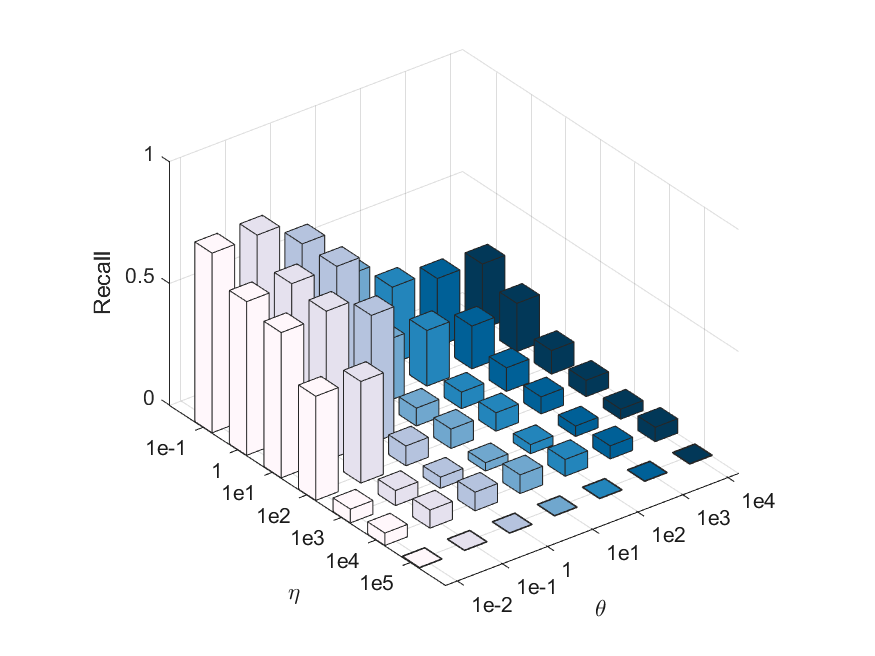}}
	\subfloat[\label{fig:p4b3}]{
		\includegraphics[width=0.25\linewidth]{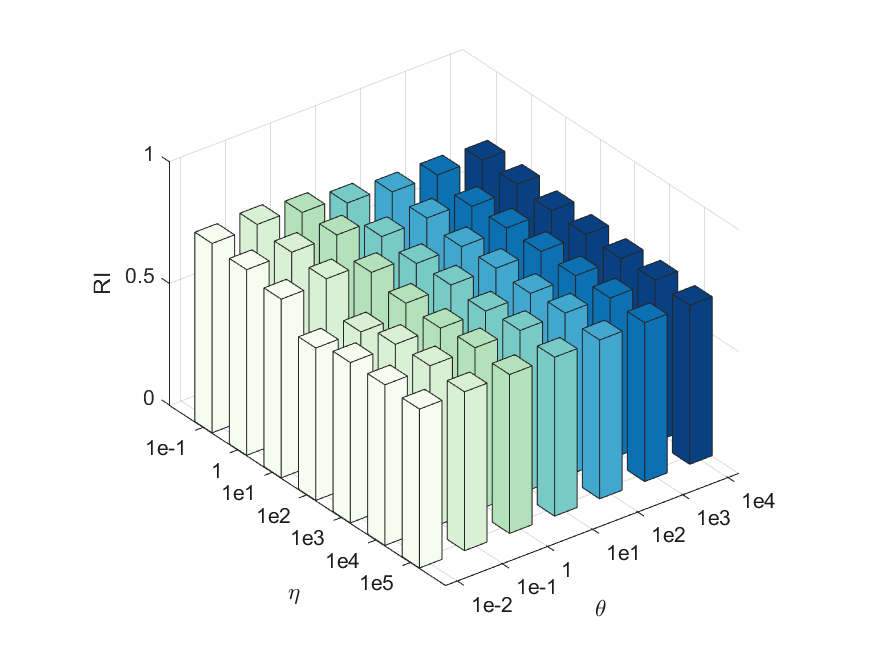}}
	\subfloat[\label{fig:p4b4}]{
		\includegraphics[width=0.25\linewidth]{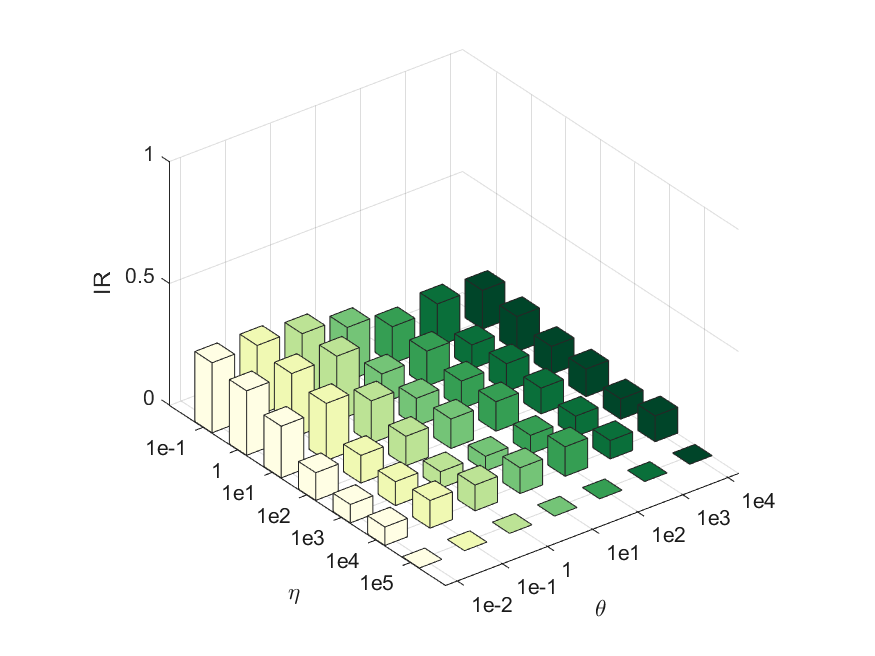}}
	\caption{How Parameter $\theta$ and $\eta$ effect the performance of MvLRECM on the Hayes dataset. (a)ACC, (b)NMI, (c)Purity, (d)Precision, (e)Recall, (f)F-score, (g)RI and (h)IR, where $x$ and $y$ axis are $\theta$ and $\eta$, respectively.}
	\label{fig:Spara4} 
\end{figure*}

\begin{figure*}
	\centering
	\subfloat[\label{fig:p5a1}]{
		\includegraphics[width=0.25\linewidth]{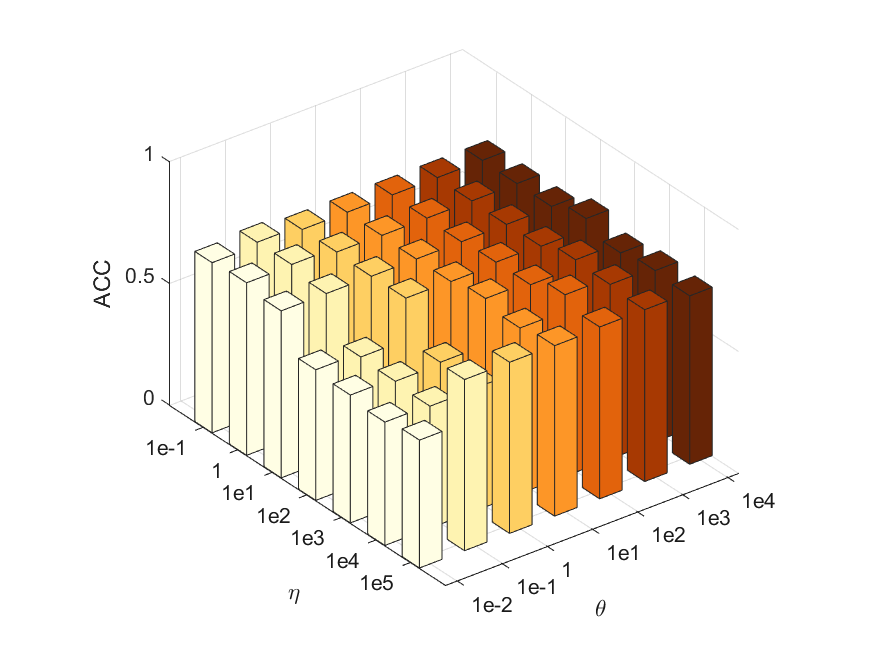}}
	\subfloat[\label{fig:p5a2}]{
		\includegraphics[width=0.25\linewidth]{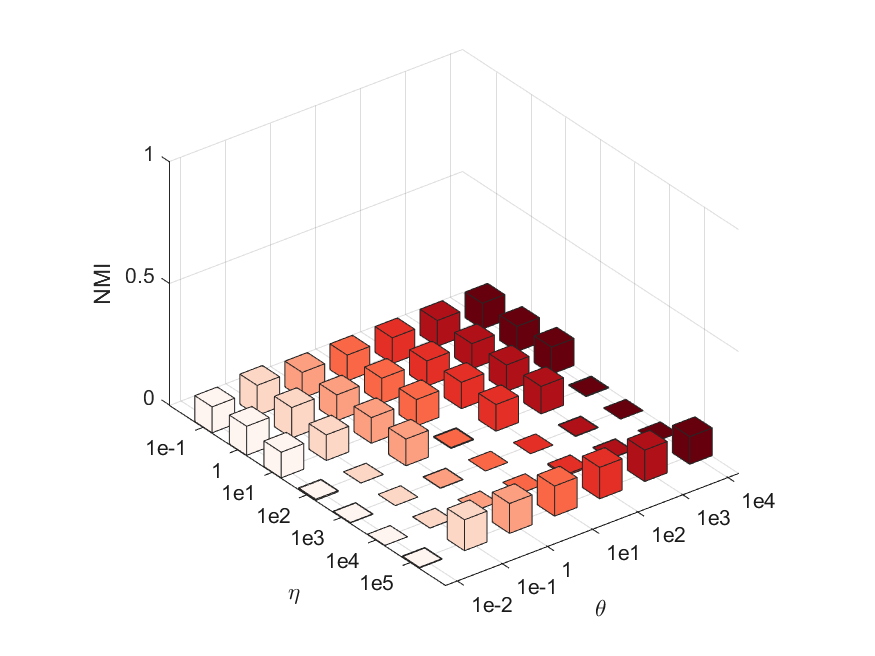}}
	\subfloat[\label{fig:p5a3}]{
		\includegraphics[width=0.25\linewidth]{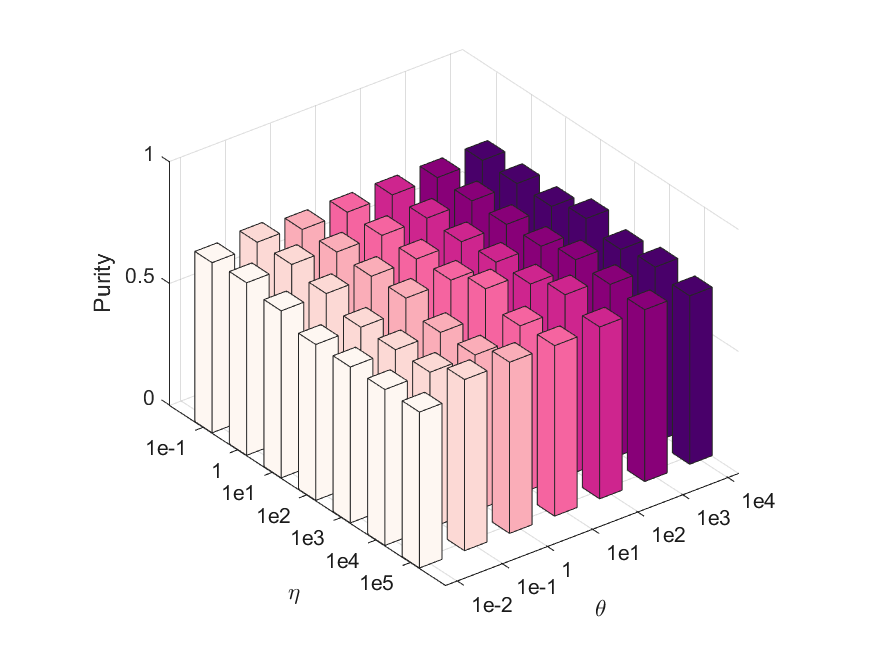}}
	\subfloat[\label{fig:p5a4}]{
		\includegraphics[width=0.25\linewidth]{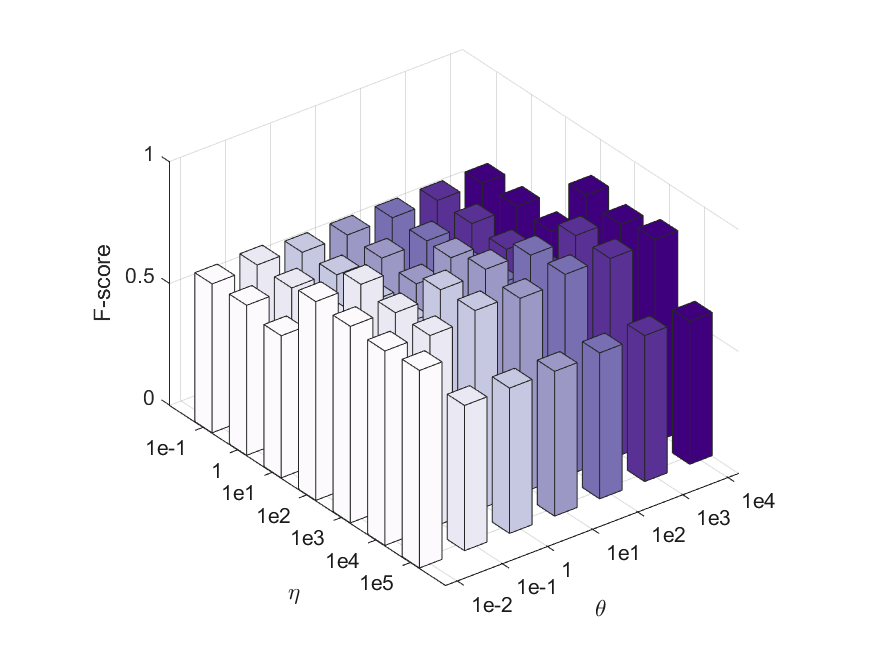}}
	\\
	\subfloat[\label{fig:p5b1}]{
		\includegraphics[width=0.25\linewidth]{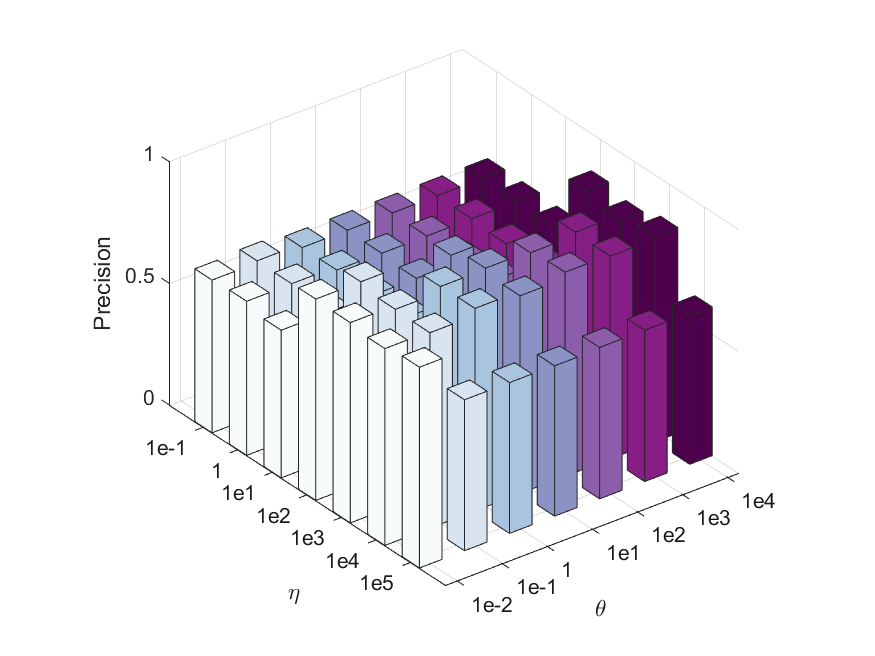}}
	\subfloat[\label{fig:p5b2}]{
		\includegraphics[width=0.25\linewidth]{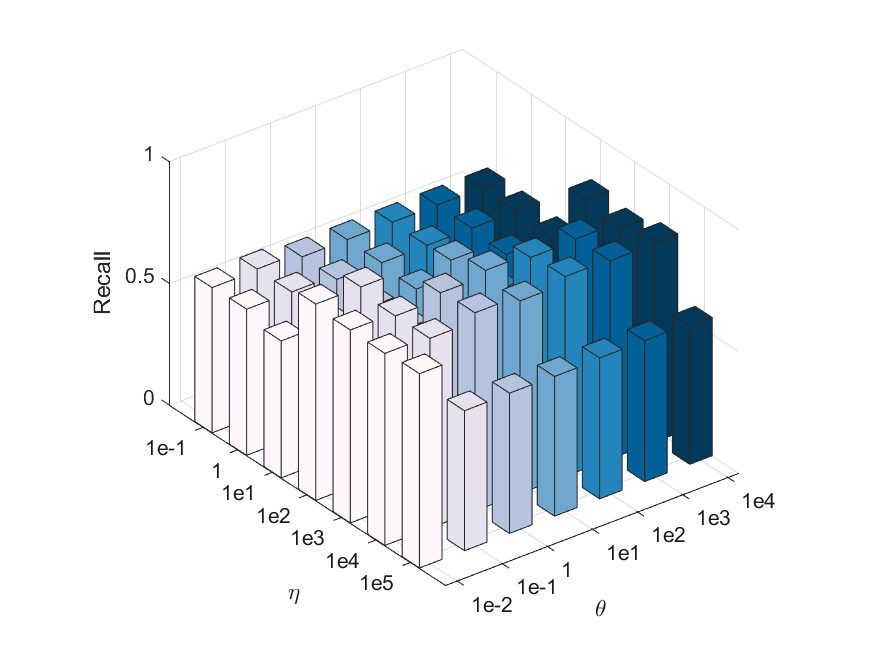}}
	\subfloat[\label{fig:p5b3}]{
		\includegraphics[width=0.25\linewidth]{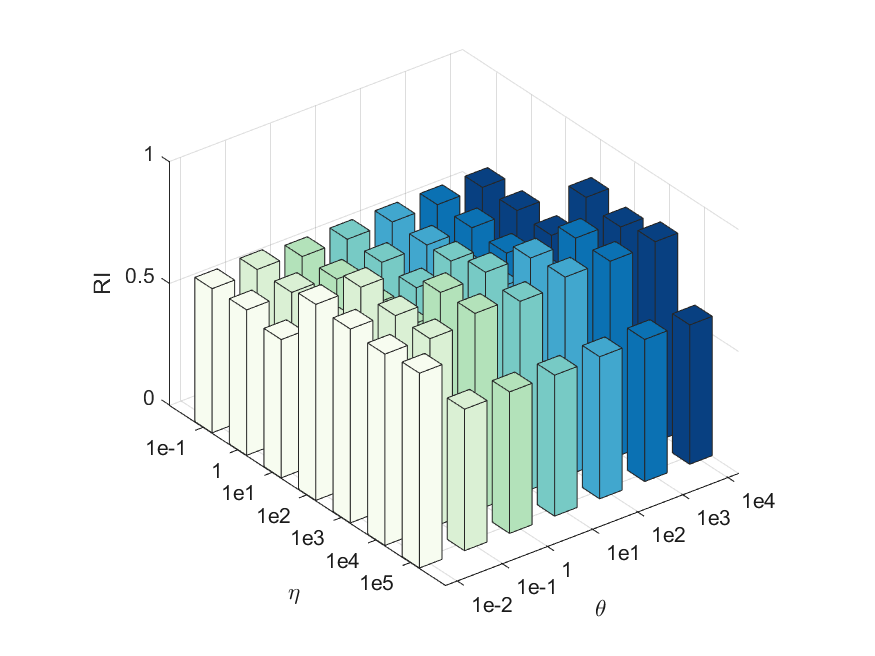}}
	\subfloat[\label{fig:p5b4}]{
		\includegraphics[width=0.25\linewidth]{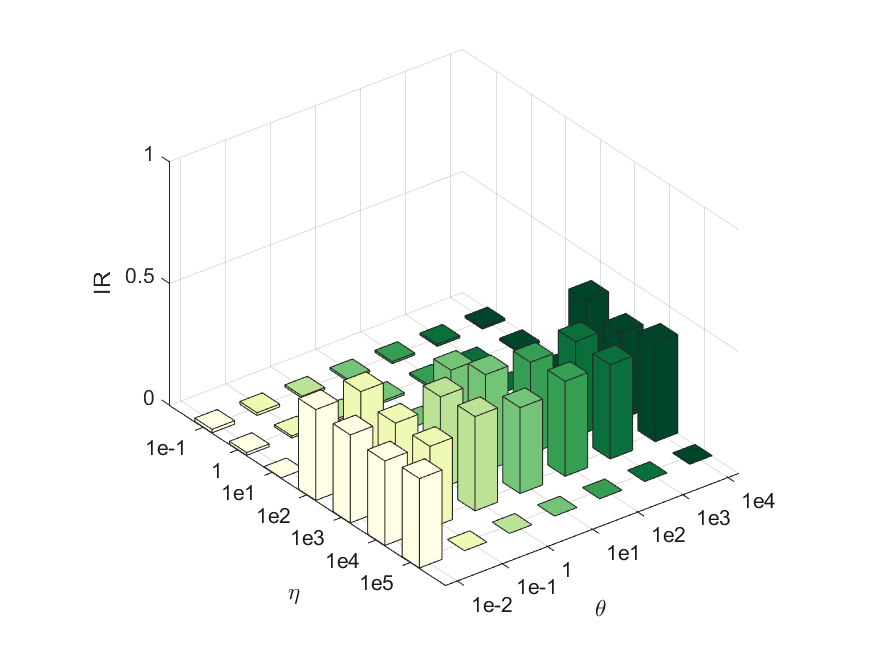}}
	\caption{How Parameter $\theta$ and $\eta$ effect the performance of MvLRECM on the Ionosphere dataset. (a)ACC, (b)NMI, (c)Purity, (d)Precision, (e)Recall, (f)F-score, (g)RI and (h)IR, where $x$ and $y$ axis are $\theta$ and $\eta$, respectively.}
	\label{fig:Spara5} 
\end{figure*}

In Sec.~\ref{sec:para} in the manuscript, we only present the performance on the Iris dataset for parameter study. 
In addition, we provide the performance of MvLRECM on the other five real-world datasets in Figs.~\ref{fig:Spara1}-\ref{fig:Spara5}.


\section{Conclusion}
\label{sec:Scon}
MvLRECM can characterize uncertainty and imprecision in multi-view clustering. 
In practical scenarios, the results of the MvLRECM can be considered as the first decision in applications. It narrows down the number of objects identified and the potential solutions of the objects which have imprecise cluster information. Subsequently, we can acquire new partial knowledge to identify these objects precisely. Although this would be costly, it is essential in some cases. In addition, MvLRECM filters out the objects identified, which helps extract more representative features of the singleton cluster, called class in the classification methods. Thus MvLRECM can also be used as the pre-processing step of classification methods. It can optimize the training set to obtain a more accurate classifier.

	\bibliographystyle{IEEEtran}
	\bibliography{TETCI}

	%
	%
	%
	%
	%
	%
	%
	%
	
	\vfill